\documentclass[10pt,journal,compsoc]{IEEEtran}

\newif\ifshowtodos
\showtodosfalse

\newif\ifFORM
\FORMfalse

\newif\iftr
\newif\ifcnf

\trtrue
\cnffalse

\newif\ifnonb
\nonbtrue

\newif\ifnohl
\nohlfalse

\newif\ifsq     

\newif\ifsqCAP
\newif\ifsqVS
\newif\ifsqEN
\newif\ifsqTIT

\newcommand{\ignore}[1]{}

\sqtrue
\sqCAPfalse
\sqENtrue
\sqVStrue
\sqTITtrue

\sqfalse
\sqCAPfalse
\sqENfalse
\sqVSfalse
\sqTITfalse

\usepackage{etex}
\usepackage{balance}
\usepackage{epstopdf}
\usepackage{placeins}
\usepackage{comment}

\usepackage{graphicx}
\usepackage{float}
\usepackage{dblfloatfix}
\usepackage{multirow}
\usepackage{rotating}
\usepackage{makecell}
\usepackage{tabulary}
\usepackage{parcolumns}
\usepackage{tikz}
\usetikzlibrary{tikzmark}

\usepackage{xpatch}
\expandafter\xpatchcmd
\csname pgfk@/tikz/every picture/.@cmd\endcsname
{\thepage}{\arabic{page}}{}{}

\tikzstyle{comment} = [draw, fill=blue!70, text=white, text width=3cm, minimum height=1cm, rounded corners, align=left, font=\scriptsize]
\tikzstyle{background_alg} = [draw, fill=blue!20, opacity=0.4, inner sep=4pt, rounded corners=2pt]

\usetikzlibrary{shapes}
\usetikzlibrary{plotmarks}
\usetikzlibrary{calc, fit}

\usepackage{enumitem}

\usepackage{amsthm}
\usepackage{amsmath,amssymb,amsfonts}
\usepackage{mathtools,mathrsfs}

\usepackage{soul}
\usepackage{fontawesome}
\usepackage{pifont}
\usepackage{textcomp}
\usepackage{booktabs}
\usepackage{url}
\usepackage{pbox}
\usepackage[normalem]{ulem}
\usepackage[10pt]{moresize}

\newcommand{\noAnswer}{\textcolor{black}{\faQuestionCircle}}

\ifsqCAP
\usepackage[font={normalfont, scriptsize}]{caption}
\usepackage[font={normalfont, scriptsize}]{subcaption}
\else
\usepackage[font={normalfont, scriptsize}]{caption}
\usepackage[font={normalfont, scriptsize}]{subcaption}
\fi

\newcommand{\vspaceSQ}[1]{\ifsqVS\vspace{#1}\fi}
\newcommand{\enlargeSQ}[1]{\ifsqEN\enlargethispage{\baselineskip}\fi}

\ifsqTIT
\usepackage[compact]{titlesec}
\titlespacing*{\section}{0pt}{3pt}{-1pt}
\titlespacing*{\subsection}{0pt}{0pt}{-3pt}
\titlespacing*{\subsubsection}{0pt}{2pt}{1pt}
\fi

\usepackage{xcolor}
\definecolor{darkgrey}{RGB}{70,70,70}
\definecolor{lightgrey}{RGB}{200,200,200}
\definecolor{lyellow}{RGB}{255,255,100}
\definecolor{llyellow}{RGB}{250,250,180}
\definecolor{lgreen}{RGB}{144,238,144}
\definecolor{raphael_comments}{RGB}{13, 145, 24}

\usepackage[customcolors]{hf-tikz}
\hfsetbordercolor{white}
\hfsetfillcolor{vlgray}

\definecolor{vlgray}{rgb}{0.77 0.77 0.77}
\definecolor{ablack}{rgb}{0.2 0.2 0.2}
\definecolor{vllgray}{rgb}{0.9 0.9 0.9}
\definecolor{bblue}{rgb}{0.7 0.7 0.99}

\usepackage{colortbl}

\usepackage{inconsolata}
\usepackage{listings}

\ifsq
\else
\fi

\newcommand{\maciej}[1]{\textcolor{blue}{[Maciej: #1]}}
\newcommand{\nils}[1]{\textcolor{orange}{[Nils: #1]}}

\newcommand{\florim}[1]{\textcolor{violet}{[Florim: #1]}}
\newcommand{\greg}[1]{\textcolor{cyan}{[Greg: #1]}}
\newcommand{\robert}[1]{\textcolor{teal}{[Robert: #1]}}
\newcommand{\piao}[1]{\textcolor{magenta}{[Piao: #1]}}
\newcommand{\zhenyu}[1]{\textcolor{olive}{[Zhenyu: #1]}}

\definecolor{hlL}{rgb}{0.8 0.8 0.99}

\newcounter{highlight}

\newcounter{hlLR}

\newcounter{hlLIR}

\newcounter{hlLIIR}

\newcounter{Ahighlight}

\usepackage{scalerel,stackengine}
\stackMath
\newcommand\rwh[1]{%
\savestack{\tmpbox}{\stretchto{%
  \scaleto{%
        \scalerel*[\widthof{\ensuremath{#1}}]{\kern-.6pt\bigwedge\kern-.6pt}%
                  {\rule[-\textheight/2]{1ex}{\textheight}}
                              }{\textheight}%
}{0.5ex}}%
\stackon[1pt]{#1}{\tmpbox}%
}

\usepackage[hang,flushmargin]{footmisc}

\renewcommand{\epsilon}{\ensuremath\varepsilon}

\renewcommand{\phi}{\ensuremath{\varphi}}

\newcommand{\fRB}[1]{\left(#1\right)}

\DeclareMathOperator*{\argmax}{arg\,max}

\usepackage{fontawesome}
\usepackage{pifont}
\usepackage{textcomp}

\usepackage{xcolor}
\usepackage{soul}
\usepackage{dblfloatfix}

\usepackage{colortbl}

\iftr
\usepackage[hidelinks]{hyperref}
\fi

\ifcnf

\else

\fi

\renewcommand{\comment}[1]{\ignorespaces}

\usepackage{physics}

\nohltrue

\ifnohl

\renewcommand{\rowcolor}[1]{}
\renewcommand{\marginpar}[1]{}

\fi

\newcolumntype{y}{>{}l}

\IEEEaftertitletext{\vspace{-2\baselineskip}}

\newif\ifHL
\HLfalse

\newcommand{\faY}[0]{\faBatteryFull}
\newcommand{\faH}[0]{\faBatteryHalf}
\newcommand{\faN}[0]{\faTimes}
\newcommand{\faU}[0]{\noAnswer}

\IEEEoverridecommandlockouts

\begin{document}

\title{Demystifying Chains, Trees, and Graphs of Thoughts}

\ifnonb
\author{Maciej Besta$^{1\dagger}$\thanks{$^\dagger$Corresponding author}, Florim Memedi$^1$, Zhenyu Zhang$^1$, Robert Gerstenberger$^1$, \\
Guangyuan Piao$^2$, Nils Blach$^1$, Piotr Nyczyk$^3$, Marcin Copik$^1$, \\
Grzegorz Kwa\'{s}niewski$^1$, Jürgen Müller$^4$, Lukas Gianinazzi$^1$, Ales Kubicek$^1$, \\
Hubert Niewiadomski$^3$, Aidan O'Mahony$^2$, Onur Mutlu$^1$, Torsten Hoefler$^1$\\
\vspace{0.5em}{\small $^1$ETH Zurich \quad$^2$Dell \quad $^3$Cledar \quad $^4$BASF SE}}
\fi

\IEEEtitleabstractindextext{%
\begin{abstract}
The field of natural language processing (NLP) has witnessed significant progress in recent years, with a notable focus on improving large language models' (LLM) performance through innovative prompting techniques. Among these, prompt engineering coupled with \emph{structures} has emerged as a promising paradigm, with designs such as Chain-of-Thought, Tree of Thoughts, or Graph of Thoughts, in which the overall LLM reasoning is guided by a {structure} such as a graph. As illustrated with numerous examples, this paradigm significantly enhances the LLM's capability to solve numerous tasks, ranging from logical or mathematical reasoning to planning or creative writing.
To facilitate the understanding of this growing field and pave the way for future developments, we devise a general blueprint for effective and efficient LLM reasoning schemes. For this, we conduct an in-depth analysis of the prompt execution pipeline, clarifying and clearly defining different concepts. We then build the first taxonomy of \textit{structure-enhanced} LLM reasoning schemes. We focus on identifying fundamental classes of harnessed structures, and we analyze the representations of these structures, algorithms executed with these structures, and many others. We refer to these structures as \emph{reasoning topologies}, because their representation becomes to a degree spatial, as they are contained within the LLM context. Our study compares existing prompting schemes using the proposed taxonomy, discussing how certain design choices lead to different patterns in performance and cost. We also outline theoretical underpinnings, relationships between prompting and other parts of the LLM ecosystem such as knowledge bases, and the associated research challenges. Our work will help to advance future prompt engineering techniques.
\vspace{-0.5em}
\end{abstract}

\iftr
\fi
}

\maketitle

\IEEEdisplaynontitleabstractindextext
\IEEEpeerreviewmaketitle

\iftr
\fi


\section{Introduction}




Large Language Models (LLMs) have become a dominant tool in modern machine learning (ML). Originating from simple Natural Language Processing (NLP) tasks~\cite{brown2020language, hartmann2022survey, radford2019language}, their far-extending potential has been quickly applied in other fields, such as logical reasoning~\cite{creswell2022selection}, planning~\cite{wang2023describe}, medicine~\cite{thirunavukarasu2023large}, and many others.
%
%
%
Since the primary communication medium with LLMs is natural language, prompt engineering has become a new area of study that gained widespread attention and importance~\cite{qiao2022reasoning, wu2022promptchainer}. First, it democratizes the access to LLMs and to the overall generative AI landscape, by being easy to use and try by anybody. Second, it is cost-effective and does not require fine-tuning or pre-training, which are expensive and time-consuming.

Crafting LLM queries to increase both the accuracy of outcomes as well as cost-effectiveness in tasks such as logical or algebraic queries is challenging. 
Despite continuous advances in the size and cognitive power of LLMs, solving elaborate tasks with a single straightforward prompt yields imprecise or plain incorrect results due to the left-to-right, one-token-at-a-time nature of generative Transformer models~\cite{wei2022chain}. Therefore, recent works focus on guiding LLMs towards the final solution through intermediate steps. Examples of such schemes include Chain-of-Thought (CoT)~\cite{wei2022chain}, Tree of Thoughts (ToT)~\cite{yao2023tree}, Graph of Thoughts (GoT)~\cite{besta2023graph}, AutoGPT~\cite{richards2023autogpt}, ReAct~\cite{yao2022react}, or LLMCompiler~\cite{kim2023llm}. This line of works increases the performance of the LLM reasoning.

Yet, despite all these advancements, state-of-the-art schemes still exhibit numerous limitations.
First, they are still limited to simple tasks such as Game of 24 -- it is critical to further enhance prompting to enable solving complex multifaceted tasks. Moreover, state-of-the-art prompting schemes often entail high inference costs~\cite{besta2023graph, yao2023tree}. Third, designing, developing, maintaining, and extending these schemes is hard. On one hand, this is due to the rapid development and enrichment of the ``LLM ecosystem'' that must be seamlessly integrated into the prompting pipeline. This includes retrieval-augmented generation (RAG), accessing the Internet, executing Python scripts, fine-tuning, and others. On other hand, {different concepts related to the LLM reasoning are not well-defined}, hindering effective design of new more powerful schemes. For example, while many schemes rely on the notion of the LLM \emph{thought}, it is not clear how it relates to concepts such as a prompt.


\begin{figure*}[t]
\centering
\vspaceSQ{-0.5em}
\includegraphics[width=1.0\textwidth]{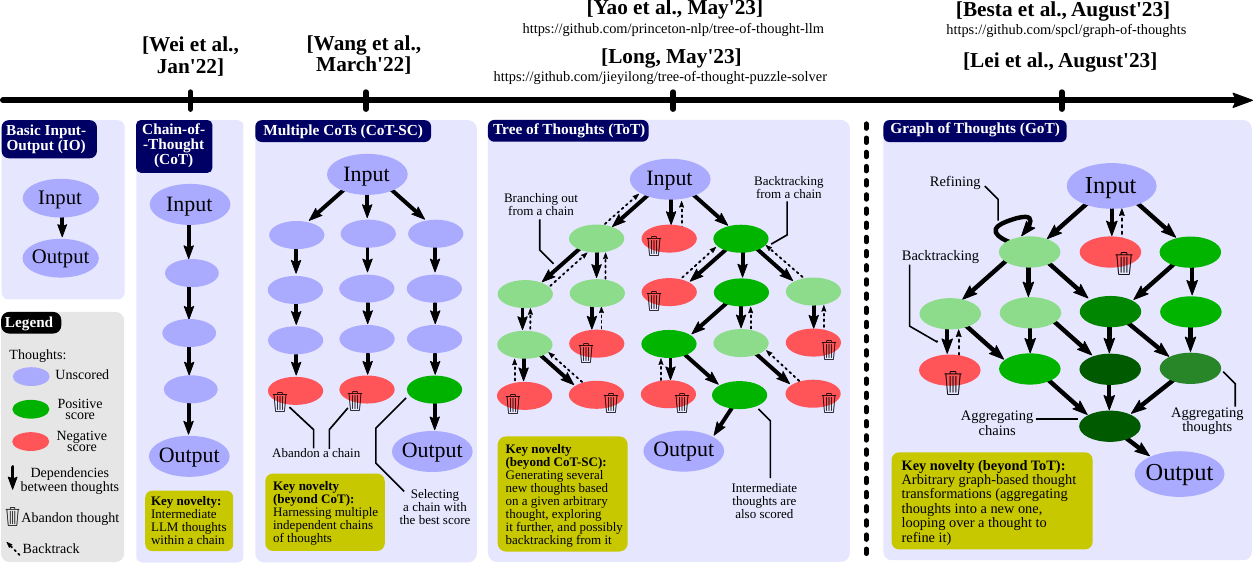}
\vspace{-1.5em}
\caption{\textbf{Evolution of reasoning topologies used in prompting schemes}.}
\iftr
\vspaceSQ{-0.5em}
\fi
\label{fig:history}
\end{figure*}

To address the above issues, we first identify and crystallize fundamental building blocks and concepts in the general prompt execution pipeline. Then, we analyze and clarify these blocks and concepts in the context of recent schemes such as CoT, ToT, and GoT (\textbf{contribution~\#1}). Our study is based on a broad analysis of recent works on LLM reasoning.
Then, we use the gained insights to develop a general blueprint and a taxonomy of the LLM reasoning schemes, focusing on how the underlying \emph{structure of reasoning} can be used to facilitate more efficient, effective, and productive prompting (\textbf{contribution~\#2}).
For this, we observe that the reasoning process in many recent prompting schemes can be modeled as a \emph{graph}. While the nature of interacting with the LLM is temporal, the representation of the graph structure behind the LLM reasoning is periodically merged with the LLM context, becoming -- to a degree -- spatial, thus forming \emph{different topologies}. These topologies can be a plain path graph (as in CoT~\cite{wei2022chain}), multiple parallel path graphs with a single root (as in CoT with Self-Consistency)~\cite{wang2022self}, a tree (as in ToT~\cite{yao2023tree}), or an arbitrary graph (as in GoT~\cite{besta2023graph}).
We then use our taxonomy to survey and analyze existing prompting schemes (\textbf{contribution~\#3}). We dissect these schemes into fundamental aspects such as the class of graphs (i.e., the topology) used to model the reasoning process, the representation of this reasoning, or the encoding of the reasoning schedule. We focus on investigating which classes of schemes offer more performance in terms of the accuracy of predictions, the latency of execution, or the cost effectiveness (\textbf{contribution~\#4}). 
We finally list open challenges and potential for novel research directions (\textbf{contribution~\#5}).

\section{Evolution of Reasoning Topologies}

We first summarize the evolution of reasoning topologies; see Figure~\ref{fig:history} for an overview.
For the sake of brevity, we do not yet precisely define the used terminology, instead relying on terms broadly used in the literature. In Sections~\ref{sec:essence-general}--\ref{sec:essence-topologies}, we introduce and discuss precise naming.

In the basic \textbf{Input-Output (IO)} prompting, the LLM provides a final reply immediately upon receiving the user initial prompt. There are no intermediate steps in the LLM reasoning. 
%
%
\textbf{Chain topologies}, introduced in Chain-of-Thought by Wei et al.~\cite{wei2022chain}, improve upon IO prompting by incorporating explicit intermediate ``steps of reasoning'' in addition to the input and output. 
%
%
%
%
\textbf{Chain-of-Thought with Self-Consistency} (CoT-SC)~\cite{wang2022self} improves upon CoT by introducing several independent reasoning chains, originating from the same initial input. Then, the best outcome from the final thoughts is chosen, according to a predefined function $S$. The driving idea is to harness the randomness within the LLM reasoning, as it can generate different thoughts from the same prompt. 

%


\textbf{Tree of Thoughts (ToT)}~\cite{long2023large, yao2023tree} elevates the CoT limitations by allowing prompt branching at any point of the chain of thoughts. Therefore, different exploration paths are not fundamentally independent, like in CoT-SC, but a chain of thoughts can branch during the reasoning process to explore different options.
%
%
A single tree node represents a partial solution. Based on a given node, the \emph{thought generator} constructs a given number $k$ of new nodes. Then, the \emph{state evaluator} generates scores for each such new node. Depending on the use case, the evaluation could be conducted using an LLM itself, or it can harness human scores. Finally, the schedule of extending the tree is dictated by the utilized search algorithm (e.g., BFS or DFS).

\begin{figure*}[t]
\centering
\vspaceSQ{-0.5em}
\iftr
\includegraphics[width=1.0\textwidth]{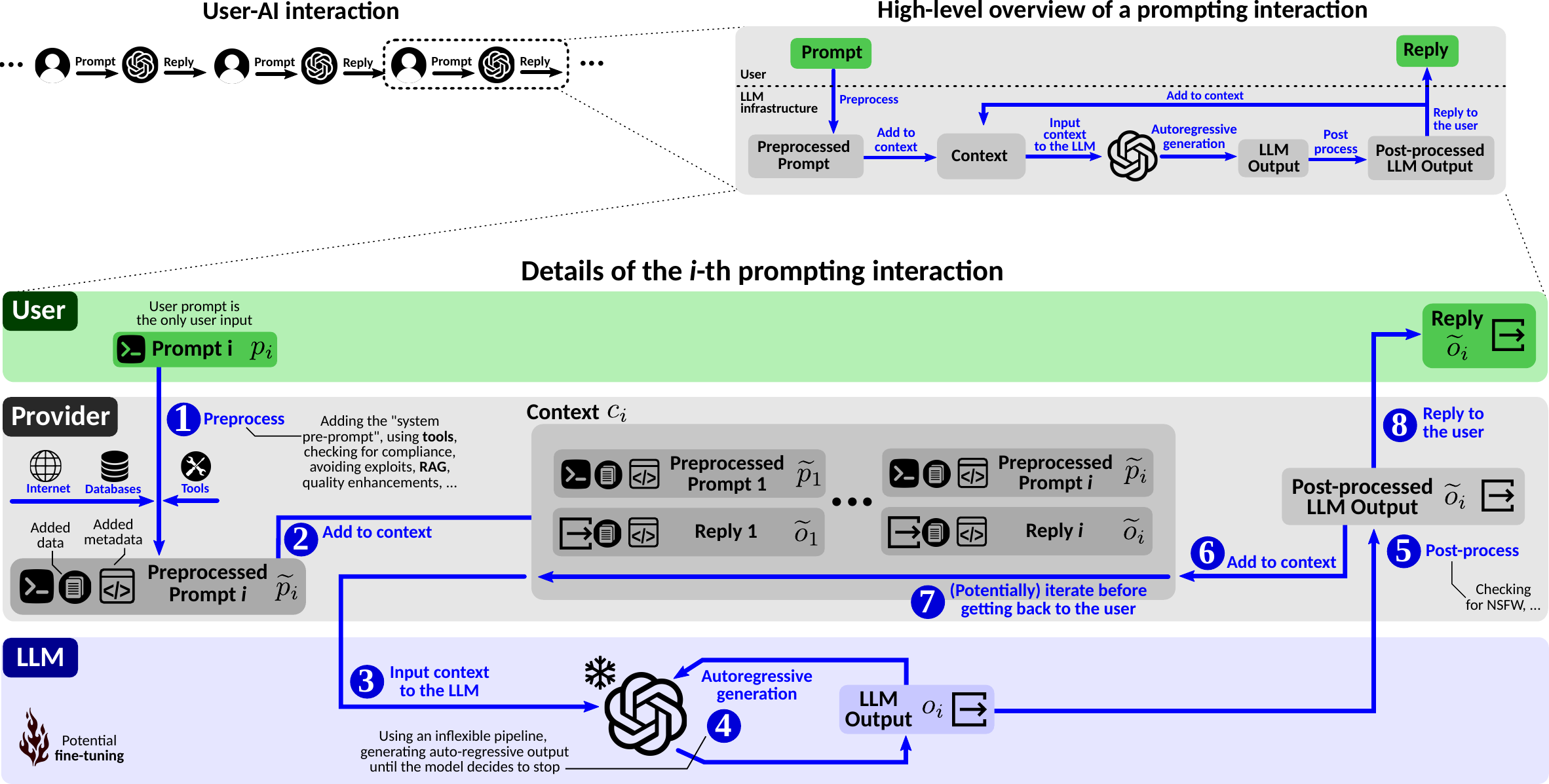}
\vspace{-2em}
\caption{\textbf{Overview of a general prompting pipeline.}}
\fi
\iftr
\vspaceSQ{-0.5em}
\fi
\label{fig:pipeline}
\end{figure*}

Finally, \textbf{Graph of Thoughts (GoT)}~\cite{besta2023graph} enables arbitrary reasoning dependencies between generated thoughts. Similarly to ToT, every thought can generate multiple child thoughts. However, each thought can also have \emph{multiple} parents, which can form an aggregation operation. GoT, allowing both branching (thoughts with out-degree $>1$) and aggregation (thoughts with in-degree $>1$) operations, can express -- for example -- reasoning patterns resembling dynamic programming, where GoT subgraphs are responsible for solving subproblems, which are then combined to form a final solution.

\section{Essence of General Prompt Execution}
\label{sec:essence-general}

We first summarize general prompt execution by giving a detailed overview of the prompting pipeline (Section~\ref{sec:pipeline}) and then establishing a functional formulation for \emph{any} prompting scheme (Section~\ref{sec:formulation}). This formulation will facilitate our subsequent analysis of reasoning topologies.

\subsection{Basic Prompting Pipeline}
\label{sec:pipeline}

\iftr
Figure~\ref{fig:pipeline} summarizes the prompting pipeline. On the left side of the figure, it shows a high-level user--AI interaction, which consists of a series of exchanges of information between the user \includegraphics[scale=0.2,trim=0 16 0 0]{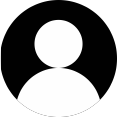} and the LLM infrastructure \includegraphics[scale=0.2,trim=0 16 0 0]{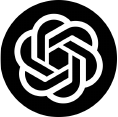}; the user sends \emph{prompts} \includegraphics[scale=0.2,trim=0 16 0 0]{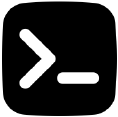} while the LLM sends back \emph{replies} \includegraphics[scale=0.15,trim=0 16 0 0]{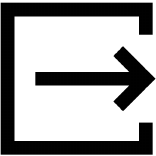}.

An individual $i$-th prompting interaction, pictured in the main central part of Figure~\ref{fig:pipeline}, starts with the user sending a prompt $p_i$.
\fi
Before being fed to the model, the prompt can be preprocessed by the LLM provider \includegraphics[scale=0.2,trim=0 16 0 0]{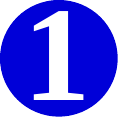}, becoming $\widetilde{p}_i$. This could include adding a ``system pre-prompt'' or some additional metadata \includegraphics[scale=0.15,trim=0 16 0 0]{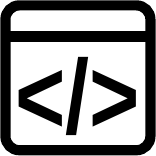}, checking for the compliance with some policies, enhancing the prompt quality, conducting retrieval augmentation \includegraphics[scale=0.2,trim=0 16 0 0]{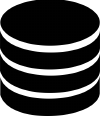}, or including outcomes from running external tools such as Python scripts \includegraphics[scale=0.2,trim=0 16 0 0]{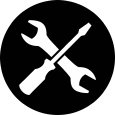} or accessing the Internet \includegraphics[scale=0.2,trim=0 16 0 0]{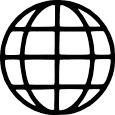}.
The preprocessed prompt is added to the LLM context \includegraphics[scale=0.2,trim=0 16 0 0]{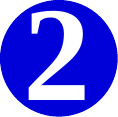}, which then is fed to the model \includegraphics[scale=0.2,trim=0 16 0 0]{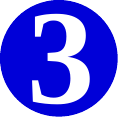} \includegraphics[scale=0.16,trim=0 16 0 0]{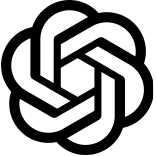}. This results in the autoregressive output generation \includegraphics[scale=0.2,trim=0 16 0 0]{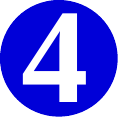} \includegraphics[scale=0.15,trim=0 16 0 0]{o.pdf}. 
The model is usually frozen \includegraphics[scale=0.2,trim=0 16 0 0]{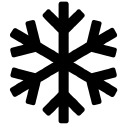} such that its weights do not change. In some considered prompting schemes, the model can also be fine-tuned \includegraphics[scale=0.2,trim=0 16 0 0]{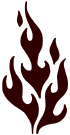}.
The output $o_i$ is potentially post-processed \includegraphics[scale=0.2,trim=0 16 0 0]{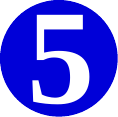}, which could involve running $o_i$ through additional neural layers (e.g., for sentiment analysis) or other forms of post-processing on the provider side, such as checking for NSFW, adding more metadata \includegraphics[scale=0.15,trim=0 16 0 0]{m.pdf}, and other operations. The post-processed output $\widetilde{o}_i$ is also added to the context \includegraphics[scale=0.2,trim=0 16 0 0]{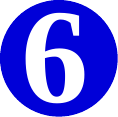} and it is sent back to the user as a reply \includegraphics[scale=0.2,trim=0 16 0 0]{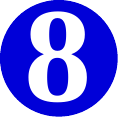} \includegraphics[scale=0.15,trim=0 16 0 0]{o.pdf}.
Note that $\widetilde{o}_i$ could also be potentially fed back to the model directly, for additional iterations, before getting back to the user \includegraphics[scale=0.2,trim=0 16 0 0]{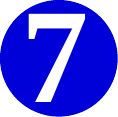}.

\subsection{Functional Formulation \& Building Blocks}
\label{sec:formulation}

We formalize the basic prompting pipeline from Section~\ref{sec:pipeline}. This allows us to crystallize its fundamental building blocks, facilitating future optimizations and propelling both efficient and effective designs. The fundamental \emph{functional} building blocks are $f_{\text{pre}}$ (for the prompt preprocessing \includegraphics[scale=0.2,trim=0 16 0 0]{1.pdf}), $f_\text{post}$ (for the post-processing of the LLM output \includegraphics[scale=0.2,trim=0 16 0 0]{5.pdf}), $\text{LLM}$ (for the auto-generative LLM execution \includegraphics[scale=0.2,trim=0 16 0 0]{4.pdf}), $f_c$ (for determining how the context is updated in stage \includegraphics[scale=0.2,trim=0 16 0 0]{2.pdf} and $f'_c$ (for determining how the context is updated in stage \includegraphics[scale=0.2,trim=0 16 0 0]{6.pdf}).
For this, we observe that the $i$-th prompting interaction (for $i = 1, ...$ and $c'_0 = \{\}$) can be formally described as
\ifcnf
\vspace{-0.5em}
\fi
\begin{align}
    \widetilde{p}_i &= f_{\text{pre}}(p_i) \\
    c_{i} &= f_c(\widetilde{p}_i, c'_{i-1}) \\
    o_i &= \text{LLM}^X (c_i) \\
    \widetilde{o}_i &= f_{\text{post}}(o_i) \\
    c'_{i} &= f'_c(c_{i}, \widetilde{o}_{i})
\end{align}
\ifcnf
\vspace{-1.5em}
\fi

\noindent
where

\begin{itemize}
    \item $p_i$ is the user prompt in the $i$-th prompting interaction,
    \item $f_{\text{pre}}(p_i)$ is a preprocessing transformation applied to $p_i$. It may involve \textbf{Retrieval Augmented Generation (RAG)}, executing a script, accessing the Internet, and using other \textbf{tools},
    \item $\widetilde{p}_i$ is the preprocessed version of the $i$-th prompt,
    \item $c_i$ is the context at the beginning of the $i$-th prompting interaction (after executing $f_c$),
    \item $c'_i$ is the context after executing $f'_c$ of the $i$-th prompting interaction (note that in the actual implementation, $c_i$ and $c'_i$ would be referring to the same data structure); note that $c'_0 = \{\}$,
    \item $o_i$ is the output of the auto-generative execution of a given LLM $X$ (e.g., LLaMA),
    \item $f_{\text{post}}(o_i)$ is a post-processing transformation applied to $o_i$; this may involve additional neural layers (e.g., for sentiment analysis), checking for compliance with guidelines, and others. While most existing schemes do not focus on this part, we expect that in the future, the post-processing transformation could also -- similarly to preprocessing -- involve executing a script, accessing the Internet, RAG, and others,
    \item $\widetilde{o}_i$ is the post-processed LLM output $o_i$; $\widetilde{o}_i = f_{\text{post}}(o_i)$,
    \item $f_c, f'_c$ are transformations that determine the exact form of updating the context.
\end{itemize}

\subsection{Implementing Building Blocks}

The provided building blocks can serve as the basis for productive implementations of prompting baselines on different architectures. For example, scheduling different parts of the prompting pipeline in the cloud setting could be done using the granularity of these blocks: a lightweight post-processing $f_{\text{post}}$ could execute a fast function, while a longer and stateful RAG operation within $f_{\text{pre}}$ could be automatically placed on EC2.

The details of $f_{\text{pre}}$, $\text{LLM}^X$, $f_{\text{post}}$, and $f_c$ depend on the specific LLM infrastructure. In general, they can be used to implement different parts of the generative AI ecosystem.
For example, most of RAG-based frameworks would implement \textbf{RAG} in $f_{\text{pre}}$.
Similarly, adding a \textbf{system pre-prompt} can be implemented as a part of $f_{\text{pre}}$.
The details of how the context is updated, or how some of its parts are removed when the input length reaches its limit, are specified in $f_c$.

In many cases, it is the user responsibility to specify the behavior of $f_{\text{pre}}, f_{\text{post}}, f_c$; this is the case -- for example -- with LLaMA or when using the OpenAI API. Contrarily, when interacting with commercial services such as ChatGPT, these transformations are defined and implemented on the LLM infrastructure side.

\iftr
\begin{figure*}[hbtp]
\centering
\vspaceSQ{-0.5em}
\includegraphics[width=0.8\textwidth]{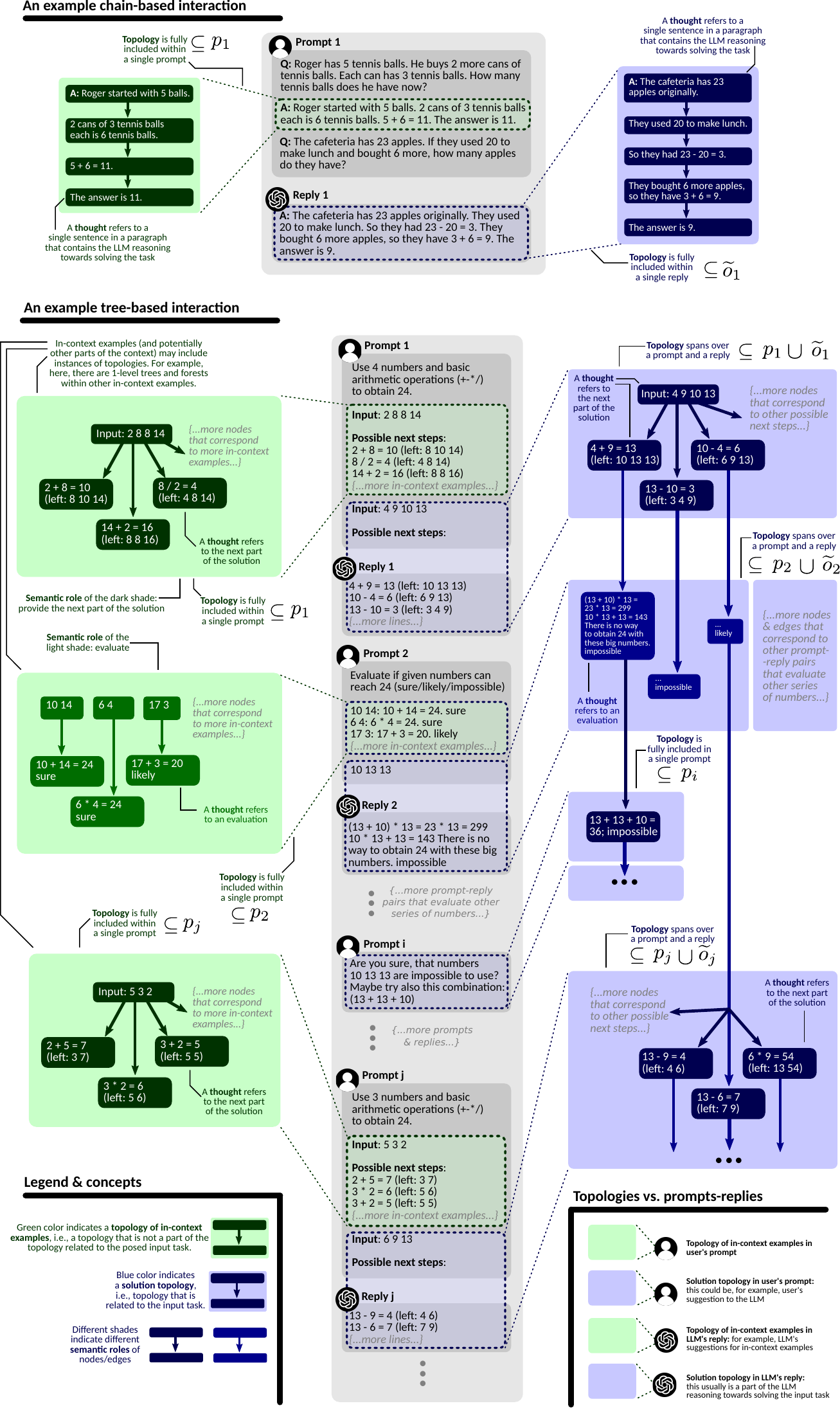}
\caption{\textbf{Overview and examples of how reasoning topologies correspond to the user--AI prompting interactions.}}
\vspaceSQ{-0.5em}
\label{fig:topos-form}
\end{figure*}
\fi

\section{Essence of Reasoning Topologies}
\label{sec:essence-topologies}


We now crystallize different concepts in the area of reasoning topologies, and relate it to the general prompting pipeline and the functional formulation from Section~\ref{sec:essence-general}.

\subsection{What Is a Thought and a Reasoning Topology?}
\label{sec:what-is-thought}

Many works use the term ``thought''. Yet, its precise meaning can differ, depending on the setting.
\iftr
For example, in CoT, a thought refers to a statement within a paragraph that contains a part of the reasoning process aimed at solving the input task.
We show this in the top part of Figure~\ref{fig:topos-form}.
In ToT, in some tasks, such as Game of 24, a thought means an intermediate or a final solution to the initial question. However, in creative writing, it could be a plan of solving the input task, or a passage of text.
In GoT, a thought contains a solution of the input task (or of its subtask). For example, it can be a subset of documents to be summarized, or a sequence of numbers to be sorted.

To encompass all these cases, we define a thought to be a \emph{semantic unit of task resolution, i.e., a step in the process of solving a given task.}
All the above examples fall into this definition: a step in task resolution can be a statement, a plan, a text passage, a set of documents, or a sequence of numbers.
\fi
\textbf{We model thoughts with nodes; edges between nodes correspond to dependencies between these thoughts.} The details of these dependencies are also use case specific.
\iftr
For example, when generating a paragraph of text, if a given paragraph $y$ is a refined version of a previous version $x$, then $x$ and $y$ become nodes in the topology, and there is an edge from $x$ to $y$ indicating that $y$ depends on $x$. If
\fi
the task is to sort a sequence of numbers, and the strategy is based on splitting the sequence into sub-sequences, sorting them independently, and merging, then the initial sequence could be modeled as a node $x$, and the sub-sequences would form further nodes $y, z, ...$, with edges $(x, y), (x, z), ...$ from $x$ to all the nodes modeling sub-sequences.
\emph{Now, a reasoning topology is a graph of these nodes and edges}.

Formally, a topology can be defined as $G = (V,E)$, where $V$ is a set of nodes modeling thoughts, and $E$ is a set of edges between these nodes, modeling reasoning dependencies between thoughts. Such a graph-theoretic approach to reason about chains, trees, and graphs of thoughts facilitates devising more efficient reasoning schemes. For example, when aiming for minimizing the latency of solving a given task, one would attempt to devise a topology with low distance between the input and output nodes.

Please note, that the concept of a thought is also not clearly defined in the field of psychology, but in general refers either to the mental representations of outside objects~\cite{shadrikov2016psychological} or self-generated mental visualizations in, e.g., daydreaming~\cite{smallwood2021neural}. This partially aligns with our definition as a semantic unit in a reasoning process (as mental representations or visualizations usually carry certain well-defined semantic meaning).

\subsection{Semantic Roles of Thoughts \& Topologies}

%
Graph nodes can model different aspects of reasoning. For example, in writing tasks, some nodes model plans of writing a paragraph, while other nodes model the actual paragraphs of text. We refer to such aspects as different \textbf{semantic roles}.
As already observed in the prompting literature~\cite{besta2023graph}, semantic roles can also be modeled with graph theory, namely with heterogeneous graphs. This enables harnessing a powerful machinery for novel LLM reasoning works. For example, one could consider using some of the heterogeneous graph learning methods~\cite{ma2023single, sun2012mining, zhang2019heterogeneous} in future prompting approaches.

\ifcnf
\begin{figure*}[hbtp]
\centering
\vspaceSQ{-0.5em}
\includegraphics[width=0.8\textwidth]{essence-topos_2-cnf.pdf}
\vspace{-0.5em}
\ifnonb
\caption{\textbf{Overview and example of how a reasoning tree topology corresponds to the user--AI prompting interactions.} An example for a chain topology can be found in the extended technical report.}
\else
\caption{\textbf{Overview and example of how a reasoning tree topology corresponds to the user--AI prompting interactions.}}
\fi
\vspace{-1.5em}
\vspaceSQ{-0.5em}
\label{fig:topos-form}
\end{figure*}
\fi

\subsection{Fundamental Use Cases of Thoughts \& Topologies}

We identify two fundamental use cases of thoughts and topologies: in-context examples and reasoning steps that bring us towards a solution.
In a topology of thoughts, a node $v$ is reachable from another node $u$, if there exists a path from $u$ to $v$.
If a node is reachable from the node modeling the input task statement, we call such a node a \textbf{solution node}, and the corresponding topology is a \textbf{solution topology}.
However, certain nodes may be not reachable from the input node. For example, a user may provide (in their prompt) in-context examples that form a small topology, which are not a step in the reasoning towards solving the input task, but merely examples. We will refer to such thoughts and topologies as thoughts and topologies of \textbf{in-context examples}.
\iftr
Examples of the two use cases can be found in Figure~\ref{fig:topos-form}.
\fi
Solution thoughts and topologies are \textcolor{blue}{\textbf{marked with the blue color}} while in-context examples are \textcolor{green}{\textbf{marked with the green color}}.
Topologies of in-context examples do not span beyond a single prompt. Solution topologies, on the other hand, can span across many prompts and replies.

Distinguishing between solution and in-context example thoughts and topologies can enable more effective and efficient LLM reasoning schemes.
For example, a graph topology has to be represented in a certain way. Now, knowing that topologies of in-context examples are usually limited to a single prompt, while solution topologies usually stretch beyond an individual prompt or LLM reply, one could use different representations for these two topology classes, in order to minimize token utilization in each of them.

Both topologies can collectively be modeled also as a {graph}, with multiple \emph{components} corresponding to topologies that are not connected with one another. To further facilitate future optimizations, one could potentially harness a \emph{hypervertex} model~\cite{bonifati2018data}, in which arbitrary subgraphs can be modeled as individual nodes called hypervertices. In such a view, one could model each separate topology of in-context examples as a hypervertex, connected with other nodes or hypervertices with hyperedges. This approach could be harnessed to provide a theoretical framework for optimizing the holistic performance of a reasoning scheme, including its ingredients such as in-context examples, in relation to all other scheme ingredients.

\subsection{Functional Formulation of Reasoning Topologies}


The LLM reasoning that harnesses topologies is formulated in the exactly same way as described in Section~\ref{sec:essence-general}.
However, one has to take into account the fact that prompts~$p_i$, replies~$\widetilde{o}_i$, and context $c_i$ in general, they all contain thoughts and their dependencies. Now, the exact way in which a topology is mapped to $p_i$, $\widetilde{o}_i$, and $c_i$, depends on a specific prompting scheme.
For example, in the $i$-th prompting interaction of CoT, a reasoning topology $T$ is a subset (potentially a proper subset) of either the LLM reply $\widetilde{o}_i$ (when $T$ is a solution topology), or of the user prompt $p_i$, as an in-context example (when $T$ is a topology of in-context examples).
In many tree and graph based schemes, however, this mapping is not so simple, and a topology can span across multiple prompts and replies.
We illustrate these examples in Figure~\ref{fig:topos-form}.

When developing an LLM reasoning scheme that harnesses reasoning topologies, one needs to specify the details of such mappings, but also build the representation of that topology, the schedule of traversing the topology, and many others. To facilitate devising future LLM reasoning schemes, we now provide a blueprint that clearly defines all these aspects and how they can be instantiated.

\begin{figure*}[t]
\centering
\vspaceSQ{-0.5em}
\iftr
\includegraphics[width=1.0\textwidth]{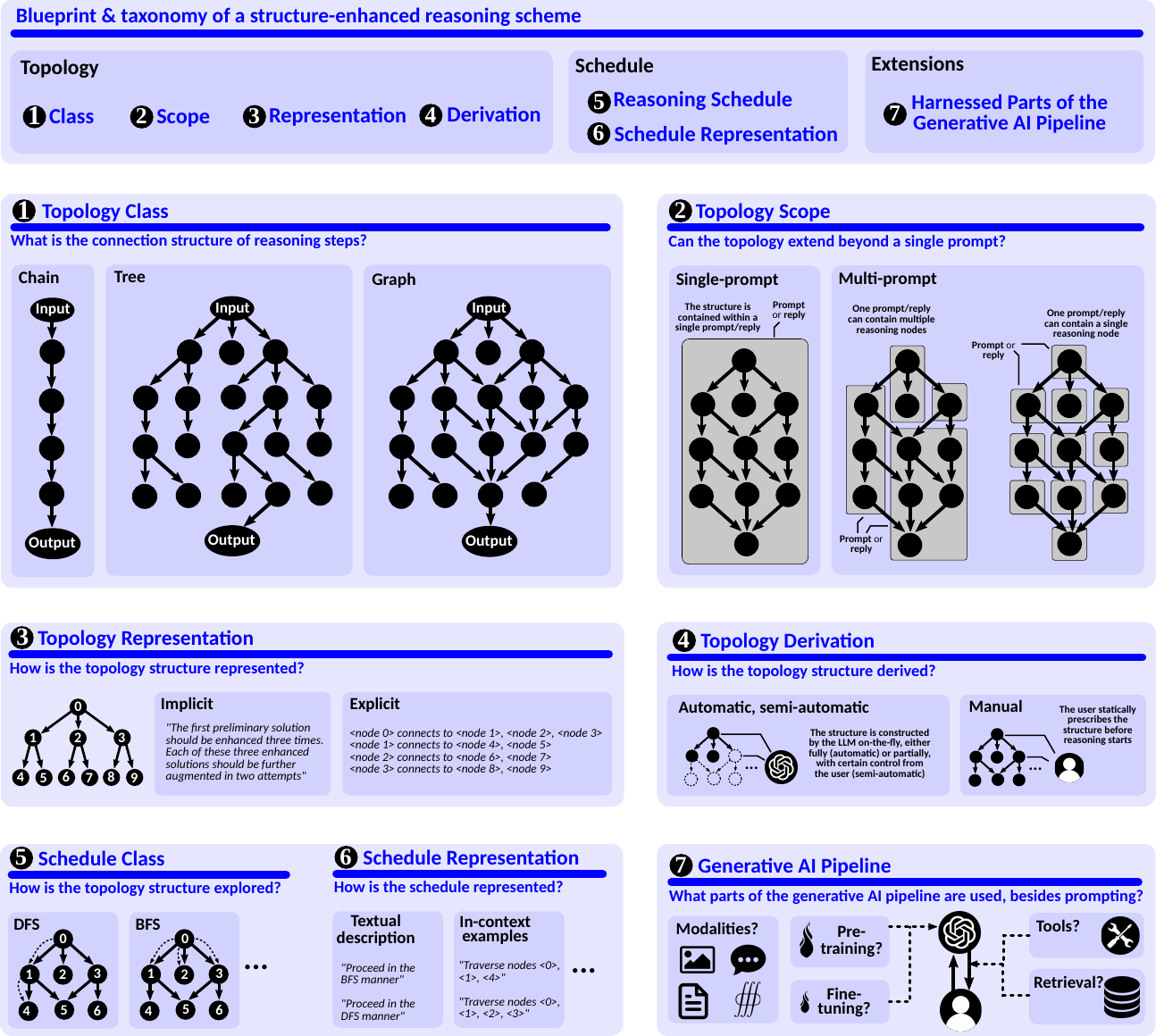}
\vspace{-1em}
\fi
\caption{\textbf{Taxonomy and blueprint for structure-enhanced reasoning.}}
\iftr
\vspaceSQ{-0.5em}
\fi
\label{fig:taxonomy}
\end{figure*}

\subsection{A Blueprint for LLM Reasoning}
\label{sec:taxo}

%

We identify the following fundamental aspects of an LLM reasoning scheme that harnesses a topology: \ding{182} \emph{topology class} (the structure of connections between intermediate steps of the LLM reasoning, Section~\ref{sec:taxo-topo}), 
\ding{183} \emph{topology scope} (the mapping between the topology and prompts/replies/context, Section~\ref{sec:taxo-topo}), \ding{184} \emph{topology representation} (how a given topology is represented within a prompt/reply/context, Section~\ref{sec:taxo-topo}), \ding{185} \emph{topology derivation} (how a given topology is obtained, Section~\ref{sec:taxo-topo}), \ding{186} \emph{reasoning schedule} (how is a given topology traversed to conduct the LLM reasoning, Section~\ref{sec:taxo-schedule}), \ding{187} \emph{schedule representation} (how is a given schedule represented within a prompt/thought, Section~\ref{sec:taxo-schedule}), and \ding{188} \emph{harnessed parts of the AI pipeline} (what parts of the generative AI pipeline, beyond prompting, are used, Section~\ref{sec:taxo-pipeline}).
We picture the blueprint in Figure~\ref{fig:taxonomy}, and we analyze existing schemes with respect to this blueprint\footnote{We encourage participation in this analysis. In case the reader is in possession of additional information relevant for the analyzed schemes, the authors would welcome the input.}, in Table~\ref{tab:schemes}.

The provided blueprint and taxonomy are applicable to both solution and in-context example topologies. For example, a topology modeling an individual in-context example can have its own representation, schedule, etc.. However, for clarity, we will focus on applying the blueprint and taxonomy mostly to solution topologies.

\subsubsection{Topology of Reasoning}
\label{sec:taxo-topo}


A reasoning scheme can harness different \textbf{topologies} for the LLM reasoning process. Here, we distinguish \textbf{chains}, \textbf{trees}, and \textbf{graphs}.
Note that both a chain and a tree are each a special case of a connected graph: a tree is an acyclic connected undirected graph, and a chain is a path graph (i.e., a tree where each node has at most one child); see Part~\ding{182} of Figure~\ref{fig:taxonomy}. Still, we treat them separately because they differ in their effectiveness for different prompting tasks~\cite{besta2023graph, long2023large, yao2023tree}.
Under this view, the plain IO prompt can be viewed as a single node graph.


Second, we observe that these topologies can be harnessed \emph{within an individual prompt or a reply} (\textbf{single-prompt topology}), but also \emph{across prompts or replies} (\textbf{multi-prompt topology}); see Part~\ding{183} of Figure~\ref{fig:taxonomy}. 

An important aspect is the \textbf{representation} of the topology, see Part~\ding{184} of Figure~\ref{fig:taxonomy}. The representation can be \textbf{implicit} (the nodes and edges are not specified explicitly) or \textbf{explicit} (nodes and edges are stated explicitly). Explicit representations vary and include a set of triples~\cite{besta2023demystifying} or a description of nodes and edges in natural text. Implicit representation depends on a scheme -- for example, it could be a textual recipe that prescribes generating the next reasoning steps.
%
%
%
Finally, we also identify \textbf{how the topology is derived} -- for example, it can be constructed by the user or by the LLM itself (Part~\ding{185} of Figure~\ref{fig:taxonomy}).
Specifically, derivation of a multi-prompt topology can be manual (fixed by user before the LLM reasoning), automatic (decided dynamically by the LLM), or semi-automatic (the overall reasoning structure is predefined before the LLM reasoning starts, but the user/LLM have some control over the structure as well during the actual reasoning).







\ifcnf

\begin{table*}[t]
\vspaceSQ{-0.5em}
\centering
\setlength{\tabcolsep}{1.5pt}
\ifsq\renewcommand{\arraystretch}{0.9}\fi
\footnotesize
\scriptsize
\ssmall
\sf
\begin{tabular}{llllllllllllllll}
\toprule
& \multicolumn{6}{c}{{\textbf{Topology}}} & \multicolumn{3}{c}{{\textbf{Reasoning}}} & \multicolumn{5}{c}{{\textbf{AI}}} & \\
\multirow{2}{*}{\textbf{Scheme}} & \multicolumn{3}{c}{{\textbf{single-prompt}}} & \multicolumn{3}{c}{{\textbf{multi-prompt}}} & \multicolumn{3}{c}{{\textbf{Schedule}}} & \multicolumn{5}{c}{{\textbf{Pipeline}}} & \multirow{2}{*}{\textbf{Remarks}} \\
\cmidrule(lr){2-4} \cmidrule(lr){5-7} \cmidrule(lr){8-10} \cmidrule(lr){11-15}
 & \textbf{Class} & \textbf{Rp.} & \textbf{Dv.} & \textbf{Class} & \textbf{Rp.} & \textbf{Dv.} & \textbf{Scheme} & \textbf{Rp.} & \textbf{Dv.} & \textbf{P} & \textbf{F} & \textbf{R} & \textbf{T} & \textbf{Modalities} & \\
\midrule
Chain-of-Thought (CoT)~\cite{wei2022chain} & chain & I (text) & SA & - & - & - & - & - & - & \faN & \faN & \faN & \faN & text & \\ 
Zero-shot-CoT~\cite{kojima2022large} & chain  & I (text) & SA & - & - & - & - & - & - & \faN & \faN & \faN & \faN & text & \\
SelfAsk~\cite{press2210measuring} & chain & I (text) & SA & - & - & - & - & - & - & \faN & \faN & \faH & \faN & text & \\
Plan-and-Solve Prompting~\cite{wang2023plan} & chain & I (text) & SA & - & - & - & - & - & - & \faN & \faN & \faN & \faN & text & \\
Program of Thoughts (PoT)~\cite{chen2022program}& chain & I (text,code) & SA & - & - & - & - & - & - & \faN & \faN & \faN & \faY & text,code,table & \\
Selection-Inference (SI)~\cite{creswell2022selection} & - & - & - & chain & E & M & linear & I & M & \faN & \faH & \faN & \faN & text & \\ 
Chain-of-Symbol (CoS)~\cite{hu2023chain} & chain & I (text) & SA & chain & E & SA & linear & I & M & \faN & \faN & \faN & \faN & text & \\
Least-to-Most Prompting~\cite{zhou2022least}  & - & - & - & chain & E & SA & linear & I & M & \faN & \faN & \faN & \faN & text & \\
Decomposed Prompting~\cite{khot2022decomposed}& - & - & - & chain & E & SA & linear & I & M & \faN & \faN & \faH & \faH & text & \\

LogiCoT~\cite{zhao2023enhancing}& chain & I (text) & SA & \textbf{tree} & E & SA & linear & I & M & \faN & \faN & \faN & \faN & text & \\
SELF-REFINE~\cite{madaan2023self}& - & - & - & chain & E & SA & linear & I & M & \faN & \faN & \faN & \faN & text & \\
Reflexion~\cite{shinn2023reflexion}& - & - & - & chain & E & SA & linear & I & M & \faN & \faN & \faN & \faN & text & \\
Reasoning Graph Verifier (RGV)~\cite{cao2023enhancing}& chain & I (text) & SA & \textbf{graph} & E & SA & linear & I & M & \faN & \faN & \faN & \faN & text & \\
Plan, Verify and Switch (PVS)~\cite{liu2023plan}& chain & I (text,code) & SA & chain & E & SA & linear & I & M & \faN & \faN & \faN & \faY & text,code & \\
Chameleon~\cite{lu2023chameleon}& - & - & - & chain & E & SA & linear & I & M & \faN & \faN & \faY & \faY & text,code & \\
ChatCoT~\cite{chen2023chatcot} & chain & I (text) & SA & chain & E & SA & linear & I & M & \faN & \faN & \faY & \faY & text & \\

\midrule
%
Tree-of-Thought (ToT)~\cite{long2023large} &
tree & I (text) & M & 
\textbf{tree} & E & SA &
\textbf{arbitrary} & E & M &
\faN & \faN & \faN & \faN & text & \\ 
Tree of Thoughts (ToT)~\cite{yao2023tree} & 
tree & I (text) & M & 
\textbf{tree} & E & SA &
\textbf{arbitrary} & E & M &
\faN & \faN & \faN & \faN & text & \\ 
Thought Decomposition~\cite{xie2023decomposition} & 
tree & I (text) & M & 
\textbf{tree} & E & SA &
\textbf{beam$^\dagger$} & E & SA &
\faN & \faN & \faN & \faH & text,code & $^\dagger$stochastic\\ 
%
Self-Consistency with CoT~\cite{wang2022self} &
chain & I (text) & M & 
\textbf{tree (\faH)$^\dagger$} & E & SA &
- & - & - & 
\faN & \faN & \faN & \faN & text & $^\dagger$depth one \\
Creswell and Shanahan~\cite{creswell2022faithful} & 
tree & I (text) & M &
\textbf{tree} & E & SA &
\textbf{beam} & E & A &
\faN & \faY & \faN & \faN & text & \\ 
Dynamic Least-to-Most Prompting~\cite{drozdov2022compositional} & 
tree & I (text) & M &
\textbf{tree} & E & A &
\textbf{bottom up} & E & A &
\faN & \faN & \faH & \faN & text,code & \\ 
Algorithm of Thoughts (AoT)~\cite{sel2023algorithm} & 
\textbf{tree} & I (text) & M & 
- & - & - & 
\textbf{DFS, (BFS)} & I & M & 
\faN & \faN & \faN & \faN & text & \\ 
Tree of Uncertain Thought (TouT)~\cite{mo2023tree} & 
tree & I (text) & M & 
\textbf{tree} & E & SA &
\textbf{BFS, DFS} & E & M &
\faN & \faN & \faN & \faN & text & \\ 
Tree-of-Mixed-Thought~\cite{hu2023treeofmixedthought} & 
tree & I (text) & M & 
\textbf{tree} & E & SA &
\textbf{DFS} & E & A &
\faN & \faN & \faN & \faY & scene graphs & \\ 
Tree of Clarifications (ToC)~\cite{kim2023tree} & 
tree (\faH)$^\dagger$ & I (text) & M & 
\textbf{tree} & E & SA &
\textbf{BFS} & E & A &
\faN & \faN & \faY & \faN & text & $^\dagger$depth one \\ 
%
%
Tree Prompting~\cite{morris2023tree} & 
- & - & - & 
\textbf{tree} & E & A & 
\textbf{top-down} & E & A & 
\faN & \faN & \faN & \faN & text & \\ 
Skeleton-of-Thought (SoT)~\cite{ning2023skeleton} &
tree (\faH)$^\dagger$ & I (text) & M & 
\textbf{tree (\faH)$^\dagger$} & E & A &
\textbf{parallel} & E & A &
\faN & \faN & \faN & \faN & text & $^\dagger$depth one \\ 
\midrule
Branch-Solve-Merge (BSM)~\cite{saha2023branch} & 
tree (depth one) & I (text) & M &
\textbf{graph (\faH)$^{\dagger}$} & E &
SA & \textbf{BFS} & E & M &
\faN & \faN & \faN & \faN & text & $^\dagger$double tree (\faH) \\ 
Thought Propagation (TP)~\cite{yu2023thought} & 
graph (\faY) & arbitrary & M &
\textbf{graph (\faH)$^{\dagger}$} & E & SA &
\textbf{BFS} & E & M & \faN & \faN & \faN & \faN & text & $^\dagger$double tree (\faY) \\ 
Socratic Questioning~\cite{qi2023art} & 
tree (depth one) & I (text) & M &
\textbf{graph (\faH)$^{\dagger}$} & E & SA &
\textbf{DFS} & E & M & 
\faN & \faN & \faN & \faN & multi & $^\dagger$double tree\\ 
Graph of Thoughts (GoT)~\cite{besta2023graph} & 
graph (\faY) & arbitrary & M & 
\textbf{graph} & E & M &
\textbf{arbitrary} & E & M &
\faN & \faN & \faN & \faN & text &  \\
Graph of Thought (GoT)~\cite{lei2023boosting} & 
\faU & \faU & \faU & 
\textbf{graph} & E & (S)A &
\textbf{DFS} & E & \faU &
\faN & \faN & \faN & \faN & text & \\ 
Graph-of-Thought (GoT)~\cite{yao2023beyond} & 
\textbf{graph} & I (text) & M &
chain & E & M & 
linear & E & M &
\faN & \faY & \faN & \faN & text,image \\ 
ControlLLM~\cite{liu2023controlllm} & 
graph & E (json) & M &
\textbf{graph} & E & M &
\textbf{DFS} & E & M &
\faN & \faN$^\dagger$ & \faN & \faY & text,image,video,audio & $^\dagger$can be used\\ 
Cumulative Reasoning~\cite{zhang2023cumulative} & 
graph (DAG) & I (text) & M &
\textbf{graph (DAG)} & E & SA &
\textbf{arbitrary} & E & M &
\faN & \faN & \faN & \faN & text & \\ 
%
%
%
Everything of Thoughts (XoT)~\cite{ding2023thoughts} & 
\textbf{graph} & \faU & L &
chain & E & M & 
linear & E & M &
\faY & \faN & \faN & \faN & text & \\ 
ResPrompt~\cite{jiang2023resprompt} & 
\textbf{graph} & I (text) & M &
- & - & - & 
- & - & - & 
\faN & \faN & \faN & \faN & text & \\ 
Hypergraph-of-Thought (HoT)~\cite{yao2023thinking} & 
\textbf{hypergraph} & \faU & M &
- & - & - & 
- & - & - & 
\faN & \faY & \faN & \faN & text,image \\ 
\midrule
BatchPrompt~\cite{lin2023batchprompt} & batch & E (text) & M & chain & E & M & linear & E & M & \faN & \faN & \faN & \faN & text & \\ 
Memory Injections~\cite{sakarvadia2023memory} & - & - & - & - & - & - & - & - & - & \faN & \faN & \faN & \faN & text \\ 
\bottomrule
\end{tabular}
\ifcnf
\vspace{-1em}
\fi
\caption{\textbf{Comparison of LLM reasoning schemes with respect to the provided taxonomy (Section~\ref{sec:taxo} and Figure~\ref{fig:taxonomy}).}
``\textbf{Topology}'': Details of the harnessed structure.
``\textbf{Single- / Multi-prompt}'': Does a given scheme support \textbf{\emph{single-} / \emph{multi}-prompt} topology? If yes, what is the supported \textbf{Class}, \textbf{Representation}, and \textbf{Derivation}?
%
%
%
%
``\textbf{Reasoning Schedule}'': Details of the harnessed \textbf{reasoning schedule}, including its specific \textbf{Scheme}, \textbf{Representation}, and \textbf{Derivation}.
``\textbf{AI pipeline}'': Does a given scheme harness parts of the AI pipeline beyond prompting? If yes, which ones? (``\textbf{P}'': pre-training, ``\textbf{F}'': fine-tuning, ``\textbf{R}'': retrieval, `\textbf{T}'': tools, ``\textbf{Modalities}'': modalities).
When describing representations, we use the following abbreviations:
``\textbf{E}'': explicit,
``\textbf{I}'': implicit.
%
%
When describing derivation, we use the following abbreviations:
``\textbf{A}'': automatic,
``\textbf{L}'': learned,
``\textbf{M}'': manual,
``\textbf{SA}'': semi-automatic.
``\faY'': full support (i.e., YES),
``\faH'': partially [supported],
``\faN'': no support (i.e., NO).
%
}
\label{tab:schemes}
\iftr
\vspaceSQ{-0.5em}
\else
\vspace{-1.5em}
\fi
\end{table*}

\fi

\subsubsection{Reasoning Schedule}
\label{sec:taxo-schedule}


The reasoning topology forms the ``skeleton'' for the LLM reasoning, effectively prescribing the algorithm for solving a given task. However, for a given fixed topology, many prompting schemes offer different approaches for the execution of the intermediate reasoning steps. For example, ToT harnesses Breadth-First Search (BFS) or Depth-First Search (DFS). This motivates us to introduce another dimension of structure-enhanced reasoning, namely the \emph{reasoning schedule}. This schedule prescribes how the topology of reasoning is going to be processed (Part~\ding{186} of Figure~\ref{fig:taxonomy}).
%

%
Whenever the schedule is specified, it can be \textbf{represented} in different ways. 
%
%
It can be a description in a natural language, a code specification, in-context examples, or others (Part~\ding{187} of Figure~\ref{fig:taxonomy}).
Finally, as with the topology, the schedule itself could also be determined using different methods, for example by the LLM on-the-fly, or pre-determined (e.g., as the fixed BFS schedule).




\subsubsection{Beyond Prompting}
\label{sec:taxo-pipeline}

Many schemes go beyond pure prompting LLMs.
%
%
This may include \textbf{pre-training}, \textbf{fine-tuning}, \textbf{retrieval}, \textbf{tools}, or different \textbf{modalities} (Part~\ding{188} of Figure~\ref{fig:taxonomy}).
%
%
We also consider this aspect, as it provides insights into the integration of reasoning topologies with other mechanisms in the AI pipeline beyond plain prompting interactions.


\iftr

\fi

\section{Reasoning with Chains}

We now proceed to investigate in more detail individual schemes that use chain topologies. We analyze these works with respect to our blueprint and taxonomy in the top part of
\iftr
Table~\ref{tab:schemes} (detailed descriptions of each individual scheme are provided in the appendix)
\fi
We also illustrate fundamental concepts introduced in these works, namely \emph{multi-step reasoning, zero-shot reasoning, planning \& task decomposition, task preprocessing, iterative refinement, and tool utilization}. We finish this section with a comparative analysis and illustrations of example topology representations.

\subsection{Multi-Step Reasoning}

The concept of multi-step reasoning was first introduced through the seminal \textbf{Chain-of-Thought (CoT)}~\cite{wei2022chain}, a single-prompt scheme, which uses topologies of in-context examples, also known as few-shot examples, to guide the LLM into reasoning step-by-step before providing the final answer.
Different following works augment or adapt in-context examples to elicit different forms of reasoning steps, while still relying on the single-prompt chain topology. For instance, instead of only providing a step-by-step reasoning chain in the examples, \textbf{SelfAsk}~\cite{press2210measuring} expands each step in the chain to also pose a follow-up question that is then answered in the subsequent step.
Similarly, \textbf{Program of Thoughts (PoT)}~\cite{chen2022program} uses code examples, rather than the natural-language-based examples as in CoT, to obtain a step-by-step generated, functional Python program that can be executed to get the final result.

\subsection{Zero-Shot Reasoning Instructions}

Zero-shot reasoning instructions aim to elicit the same multi-step reasoning chains, but without the use of hand-tuned, problem-specific in-context examples, i.e., they eliminate chain topologies forming in-context examples.
\textbf{Zero-shot-CoT}~\cite{kojima2022large}, an extension to CoT, achieves this by simply prompting the LLM with one sentence, \emph{``Let's think step by step''}, or using other similar statements. Along the same lines, \textbf{PoT} can also leverage zero-shot reasoning instruction, e.g., \emph{``Let's write a Python program step by step and return the result. Firstly we need to define the variables."}. 

\subsection{Planning \& Task Decomposition}


Both planning and task decomposition aim to break down a task into a number of manageable sub-tasks that help reaching the final solution.
\textbf{Plan-and-Solve (PS) Prompting}~\cite{wang2023plan}, one of the key single-prompt schemes building on this concept, first divides the complex task into a chain of sub-tasks and then executes these step-by-step for the final solution. PS operates in a zero-shot, multi-step manner, thus also relies on the previous two concepts as well.
%

Planning and decomposition is also frequently used in multi-prompt chains. The introduction of a node for specifying the details of the decomposition at the start of a multi-prompt reasoning chain, usually not only determines the chain's depth, but also facilitates more effective reasoning approaches in the subsequent sub-steps. This allows for a finer-grained resolution of sub-tasks, enriching the overall reasoning process. 
Here, \textbf{Least-to-Most Prompting}~\cite{zhou2022least} grows a reasoning chain where decomposition of complex tasks or questions is conducted in the first node and the sub-tasks/sub-questions are solved in the subsequent nodes.
Specifically, the multi-prompt chain operates by first decomposing the original question into a list of sub-questions, which are each solved in individual sub-steps with the questions and answers of previous sub-steps included in the context.
The chain terminates when the final answer is returned after answering all sub-questions in the list.
Then, \textbf{Decomposed Prompting}~\cite{khot2022decomposed} is a modular framework for a detailed decomposition of complex tasks. To generate a reasoning chain, the LLM is prompted by demonstrations comprised of sequential question-operation triplets, which form ``sub-questions''. In contrast to Least-to-Most prompting, this allows for the recursive breakdown of questions into simpler sub-questions until they can be solved directly, as further decomposition is a valid operation in the framework.
Apart from these two above schemes, decomposition is applied in many similar works~\cite{Dua2022SuccessivePF}, \cite{jung2022maieutic},  \cite{ye2023large}.

\subsection{Task Preprocessing}


The concept of task preprocessing comprises any technique that preprocesses the context of a task by updating it or rephrasing the task description itself, before taking any reasoning steps. 
For instance, the multi-prompt scheme \textbf{Selection-Inference (SI)}~\cite{creswell2022selection} is designed to tackle multi-step logical reasoning problems where all essential information is already present within the input context. The key functionality of SI lies in its recurring process of context pruning before each reasoning step.  This means that it selectively filters the context to retain only the relevant information necessary for each specific subsequent step of reasoning, ensuring that the most pertinent data is always used for each decision-making stage.
On the other hand, instead of pruning the context, the multi-prompt scheme \textbf{Chain-of-Symbol (CoS)}~\cite{hu2023chain}, specifically designed for spatial planning tasks, augments the context with condensed symbolic representations, before subsequently using those as prompts for the LLM to conduct CoT-based reasoning.

\subsection{Iterative Refinement}


The introduction of verification enables the reasoning frameworks to iteratively refine the generated context and intermediate results. With this strategy, the execution of chain-based reasoning is effectively extended with loops, with conditions on how many times one can loop over a node (based on the number of iterations or some terminal conditions). The concept is applied in different works~\cite{li2023makingstep, ling2023deductive, madaan2023self, shinn2023reflexion, weng2022large, zhao2023enhancing}.

\subsection{Tool Utilization} 

To better integrate multiple execution methods, more effective schemes opt to devise a plan that specifies tools for handling each sub-task, before executing the reasoning chain. Examples include \textbf{AutoGPT}~\cite{richards2023autogpt}, \textbf{Toolformer}~\cite{schick2023toolformer}, \textbf{Chameleon}~\cite{lu2023chameleon}, \textbf{ChatCot}~\cite{chen2023chatcot}, \textbf{PVS}~\cite{liu2023plan} and others~\cite{zhao2023survey}.



\ifcnf
\begin{figure}[t]
\centering
\includegraphics[width=1.0\columnwidth]{chain_math_reasoning_prompt_v2_cnf.pdf}
\vspace{-1.5em}
\ifnonb
\caption{\textbf{\textsc{Math Reasoning.} An \ul{implicit} \ul{single-prompt} chain topology example, encoded with \ul{text}}. It shows how zero-shot CoT solves a math question from GSM8K~\cite{cobbe2021gsm8k}. This results in an implicit chain where individual reasoning steps can be regarded as nodes. A few-shot CoT example can be found in the extended technical report.}
\else
\caption{\textbf{\textsc{Math Reasoning.} An \ul{implicit} \ul{single-prompt} chain topology example, encoded with \ul{text}}. It shows how zero-shot CoT solves a math question from GSM8K~\cite{cobbe2021gsm8k}. This results in an implicit chain where individual reasoning steps can be regarded as nodes.}
\fi
\vspace{-1.5em}
\label{fig:CoT-mathPrompt-zeroShot}
\end{figure}
\fi

\subsection{Analysis \& Comparison of Designs}

We now broadly discuss and analyze chain designs with respect to different aspects of our blueprint.

\iftr
\subsubsection{Topology \& Its Construction}
\fi
%
In single-prompt schemes, the entire reasoning process is executed within a single prompting round. This approach is less common for complex tasks, as it often demands sophisticated prompt engineering to encompass the entire reasoning pathway in one go. On the other hand, most chain designs employ multi-prompt schemes, where the reasoning process is segmented into multiple rounds of prompting. This allows for a more nuanced and step-wise approach to problem-solving. {Key novel architectural features of chain designs} include the ability to appropriately {decompose tasks}, {verify and refine} intermediate solutions, preprocess initial prompts, and utilize {external tools}, such as Python scripts. This multifaceted approach enables LLMs to tackle more complex problems by breaking them down into smaller, more manageable components and iteratively refining the solutions.



\begin{figure}[t]
\centering
\includegraphics[width=1.0\columnwidth]{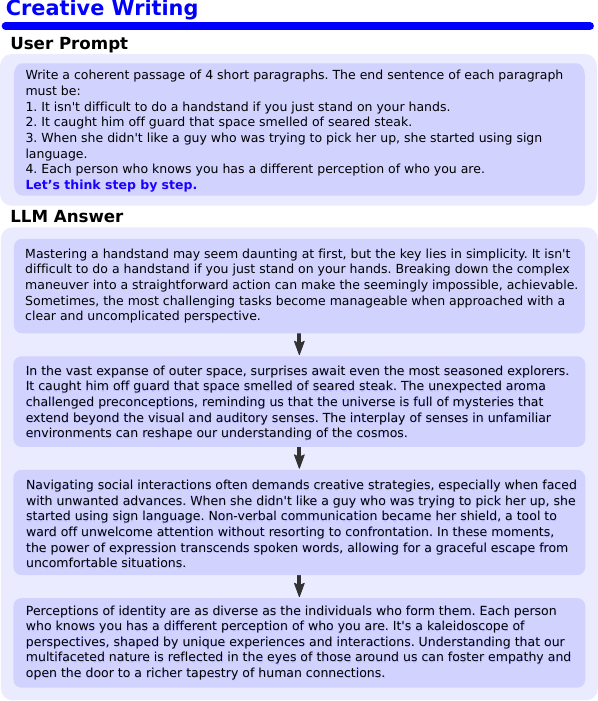}
\ifcnf
\vspace{-1.5em}
\fi
\caption{\textbf{\textsc{Creative Writing.} An example \ul{implicit} \ul{single-prompt} chain topology, encoded with \ul{text}}. It shows the Creative Writing example from ToT~\cite{yao2023tree} using zero-shot CoT prompting. This results in an implicit chain with a first node for the question and one node for each generated paragraph.}
\ifcnf
\vspace{-1.5em}
\fi
\label{fig:CoT-creativeWriting-zeroShot}
\end{figure}

\iftr
\subsubsection{Representations of Topology \& Schedule}
\fi
We now illustrate representative prompts that show differences between single- and multi-prompt chain-based topologies (the tasks are described in detail in the prompts of the figures).
\iftr
We use examples based on the well-known tasks of Game~of~24, Creative Writing, and Mathematical or Logical Reasoning.
For this, we illustrate an example simplest IO scheme in Figure~\ref{fig:CoT-gameof24-fewshot} (effectively a single-node CoT) and compare it to an implicit single-prompt few-shot CoT in Figure~\ref{fig:CoT-mathPrompt-fewShot}, an implicit single-prompt \emph{zero-shot} CoT in Figure~\ref{fig:CoT-creativeWriting-zeroShot} and to an example few-shot \emph{explicit} \emph{multi-prompt} CoT (Selection-Inference) in Figure~\ref{fig:SI-prompt}.
\fi

\iftr
\begin{figure}[t]
\centering
\includegraphics[width=1.0\columnwidth]{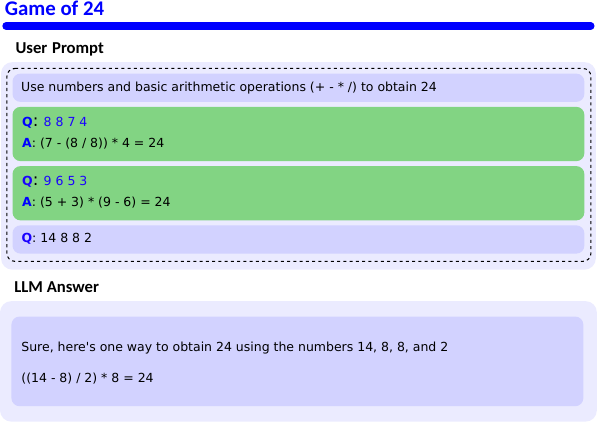}
\caption{\textbf{\textsc{Game of $24$.} An example showing an IO scheme (effectively an \ul{implicit} \ul{single-prompt} single-node chain topology).} A few-shot IO prompt is used, leading to a single output directly providing the result with no intermediate reasoning steps.}
\label{fig:CoT-gameof24-fewshot}
\end{figure}
\fi

\iftr
\begin{figure*}[t]
\centering
\includegraphics[width=1.0\textwidth]{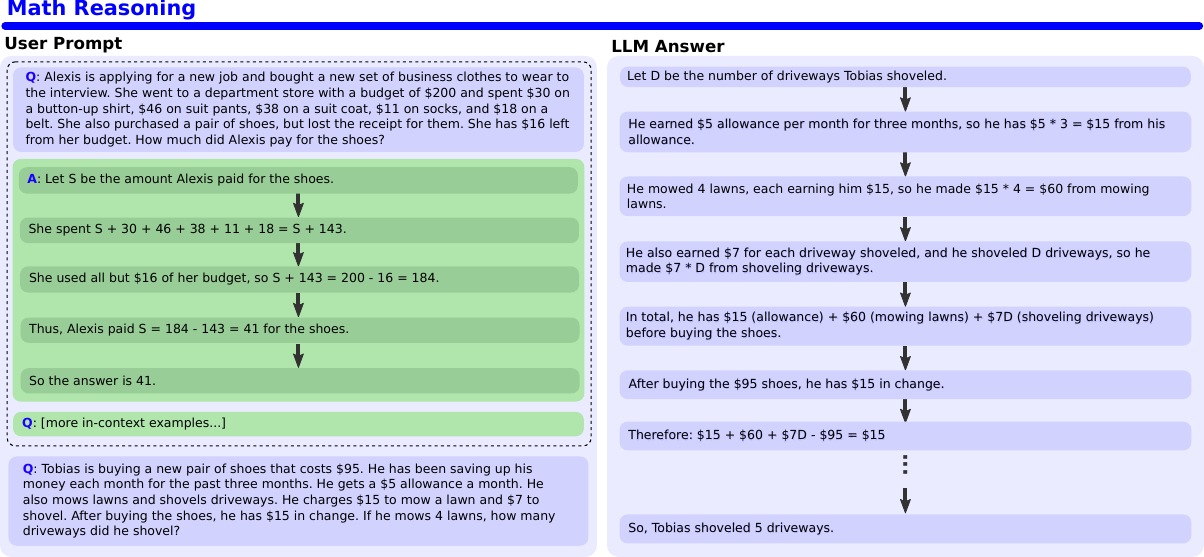}
\caption{\textbf{\textsc{Math Reasoning.} Another \ul{implicit} \ul{single-prompt} chain topology example, encoded with \ul{text}}. It shows how few-shot CoT solves a math question from GSM8K~\cite{cobbe2021gsm8k}. This results in an implicit chain where individual reasoning steps can be regarded as nodes.}
\label{fig:CoT-mathPrompt-fewShot}
\end{figure*}
\fi

\lstalias{plantuml}{}
\lstset{escapeinside={<@}{@>}}

\iftr
\begin{figure}[t]
\centering
\includegraphics[width=1.0\columnwidth]{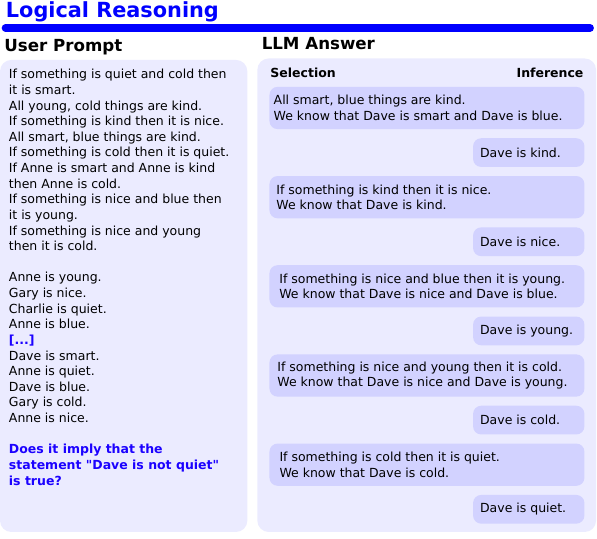}
\caption{\textbf{\textsc{Logical Reasoning.} An example \ul{explicit} \ul{multi-prompt} chain topology, encoded with \ul{text}}. It demonstrates the \ul{manually} derived chain topology of Selection-Inference~\cite{creswell2022selection}. "Selection" and "Inference" indicate \ul{independent} consecutive prompts/thoughts.}
\label{fig:SI-prompt}
\end{figure}
\fi

\iftr
\begin{figure*}[t]
\centering
\vspaceSQ{-0.5em}
\includegraphics[width=1.0\textwidth]{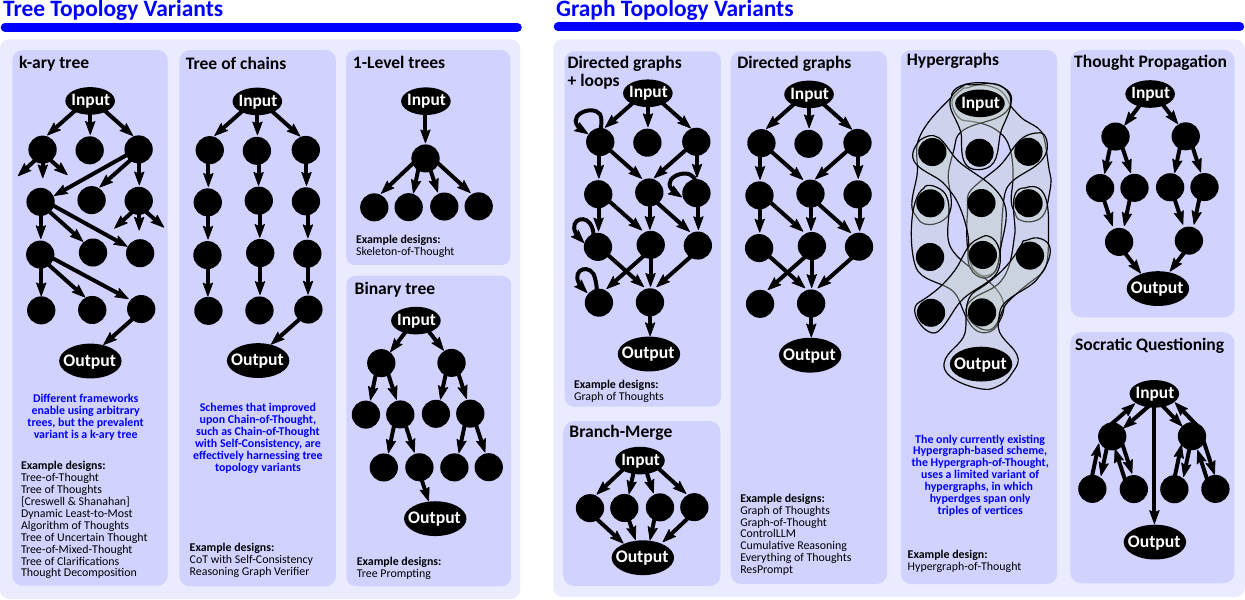}
\vspace{-1em}
\caption{\textbf{Variants of tree and graph prompting topologies}.}
\vspaceSQ{-0.5em}
\label{fig:topologies-dets}
\end{figure*}
\fi

\subsubsection{Performance}


\iftr
We now summarize performance patterns found in chain topologies. For a more detailed performance comparison, see Appendix \ref{sec: chain_extensive}.
\fi

Overall, in Arithmetic Reasoning, CoT significantly outperforms Input-Output (IO) prompting, with notable improvements in terms of accuracy on several datasets from benchmarks such as 
\iftr
GSM8K (Grade School Math)~\cite{cobbe2021gsm8k}, SVAMP (Simple Variations on Arithmetic Math word Problem)~\cite{patel2021nlp}, and MAWPS (MAth Word ProblemS)~\cite{koncel2016mawps},
\fi
which focus on mathematical tasks. The effectiveness of CoT increases with the scale of the LLM. Variants like Zero-shot-CoT, PoT, and schemes with decomposition nodes (like Least-to-Most Prompting, PS+) show further improvements in specific datasets. Chameleon, with its table reader tools, enhances performance in tabular math questions.

In Commonsense Reasoning, CoT excels over IO prompting with a distinct advantage in datasets like StrategyQA~\cite{geva2021strategyqa}. Specialized methods like SelfAsk and the Selection-Inference framework show further improvements in multi-hop questions. Decomposition-based schemes like Least-to-Most and Decomposed Prompting achieve high accuracy in tasks requiring sequential actions or multi-faceted reasoning. ChatCoT, with its retrieval and external tool utilization, shows a 20\% gain on specific datasets.

In Symbolic Reasoning, CoT demonstrates near-perfect accuracy for in-context examples and substantial accuracy for out-of-domain cases for tasks like last letter concatenation and coin flip predictions. Schemes with decomposition nodes outperform CoT in more complex tasks and longer word scenarios.

Overall, across different domains, CoT and its variants show a consistent trend of outperforming basic IO prompting. The integration of additional tools, tailored prompting strategies (like few-shot or zero-shot), and the incorporation of decomposition and refinement nodes enhance the LLMs' performance significantly.

\ifcnf
\begin{figure}[t]
\centering
\includegraphics[width=1.0\columnwidth]{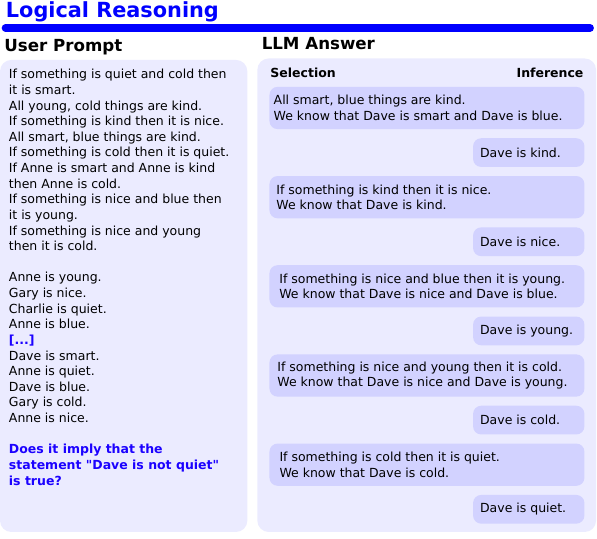}
\vspace{-1.5em}
\caption{\textbf{\textsc{Logical Reasoning.} An example \ul{explicit} \ul{multi-prompt} chain topology, encoded with \ul{text}}. It demonstrates the \ul{manually} derived chain topology of Selection-Inference~\cite{creswell2022selection}. "Selection" and "Inference" indicate \ul{independent} consecutive prompts/thoughts.}
\vspace{-1.5em}
\label{fig:SI-prompt}
\end{figure}

\begin{figure*}[t]
\centering
\vspaceSQ{-0.5em}
\includegraphics[width=1.0\textwidth]{schemes-topology_5.pdf}
\vspace{-2em}
\caption{\textbf{Variants of tree and graph prompting topologies}.}
\vspace{-1.5em}
\label{fig:topologies-dets}
\end{figure*}
\fi

\section{Reasoning with Trees}

We next investigate in more detail individual schemes that use tree topologies. We analyze these works with respect to our blueprint and taxonomy in the middle part of
\iftr
Table~\ref{tab:schemes} (detailed descriptions of each individual scheme are provided in the appendix).
\fi
We structure the discussion based on the harnessed topology variants, namely trees of chains, 1--level trees, and $k$--ary trees, see Figure~\ref{fig:topologies-dets} for details.
As in chain schemes, we also discuss fundamental concepts introduced or harnessed in these works. Tree schemes, most importantly, introduce \emph{exploration} (i.e., generating multiple thoughts from a given one).
The purpose behind exploration is usually either \emph{task decomposition} (which is similar to CoT, but it differs as decompositions are not limited to a single linear plan) or \emph{sampling} (i.e., having a higher chance of obtaining a high-quality solution). Moreover, tree schemes also introduce \emph{voting} (i.e., automatic selection of the best outcome of all the generated outputs) and they harnesses various architectural concepts also used in chain schemes, such as iterative refinement or task preprocessing.
We finish this section with a comparative analysis and illustrations of example topology representations.

\subsection{Trees of Chains}

While trees as reasoning topologies have been explicitly established in the works by Long~\cite{long2023large} and Yao~\cite{yao2023tree}, this idea has been present earlier.
%
%
%
%
\textbf{Chain-of-Thought with Self-Consistency (CoT-SC)}~\cite{wang2022self} is an early scheme that harnesses the tree structure to a certain degree. Here, multiple CoTs originate from the same initial (root) prompt, forming a ``tree of chains''. The chain providing the best outcome to the initial question, is selected as the final answer.

\subsection{Single-Level Trees}

A tree-based approach has also been harnessed in \textbf{Skeleton-of-Thought (SoT)}~\cite{ning2023skeleton}, which effectively harnesses a tree with a single level of depth.
This scheme aims to reduce the end-to-end generation latency of LLMs, caused by their inherent sequential decoding. Instead of generating one long continuous answer, this scheme uses a divide-and-conquer approach. In a first prompt, the LLM is instructed to generate a skeleton of the answer, i.e., a list of points that are independently answerable. Then, for each of these points, a new prompt is issued in parallel to answer just this specific part of the question. As these points are processed in parallel, the overall latency is reduced.

\ifcnf
\begin{figure*}[t]
\centering
\includegraphics[width=1.0\textwidth]{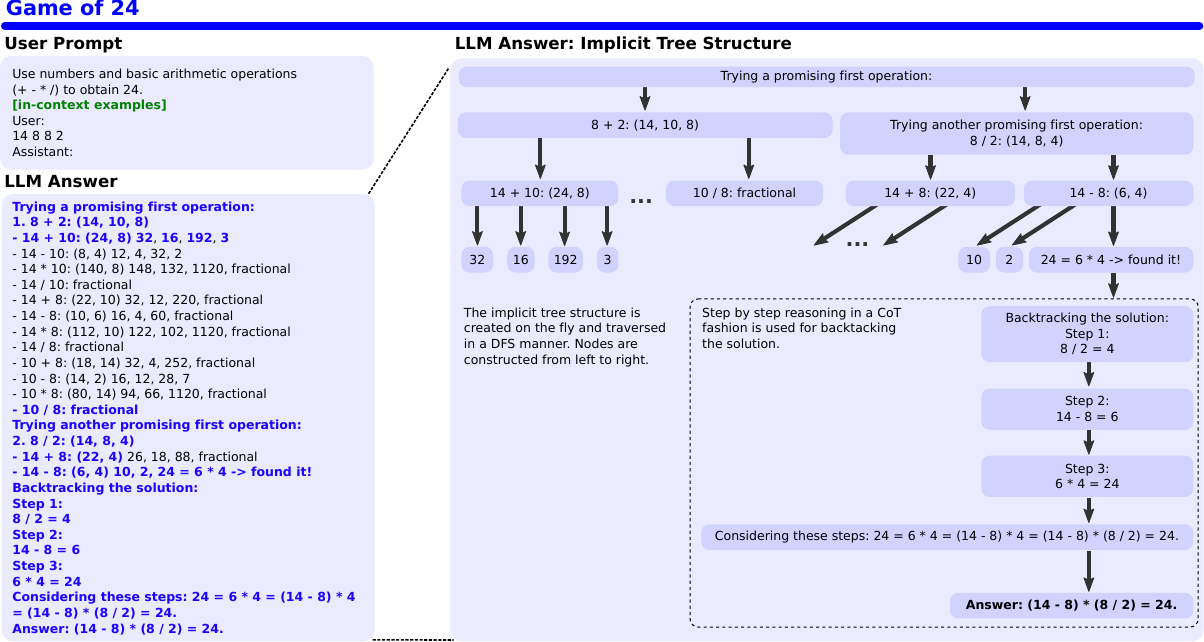}
\vspace{-1.5em}
\caption{\textbf{\textsc{Game of 24.} An example \ul{implicit} \ul{single-prompt} tree topology, encoded with \ul{text}.} It demonstrates a Game of $24$ DFS in-context example from AoT~\cite{sel2023algorithm}. The left view shows the user prompt and the single textual answer from the LLM. The right view shows the implicit tree structure that is explored during the generation of the LLM answer. We mark text corresponding to implicit nodes as \textcolor{blue}{\textbf{bold}}.}
\vspace{-1.5em}
\label{fig:AoT Prompt}
\end{figure*}
\fi

\subsection{$k$--Ary Trees}

Numerous schemes have harnessed more general $k$--ary trees.
First, the \textbf{Tree-of-Thought (ToT) design by Long}~\cite{long2023large} utilizes a tree structure to decompose a problem into sub-problems and solve them using separate LLM prompts. After the LLM suggests possible next steps and corresponding partial solutions, a \emph{checker module} decides if any of these solutions is valid, whether it can be selected as the final one, or whether it should backtrack to the previous step. All issued prompts and answers are explicitly stored as a tree structure and navigated through using a \emph{controller module}. The LLM prompting is only used to generate the next individual steps (i.e., hops) in this tree, whereas the overall problem solving process is coordinated by the controller. 

\textbf{Tree of Thoughts (ToT) by Yao et al.}~\cite{yao2023tree} differs from the above ToT approach in using the LLM itself as a solution \emph{evaluator} with access to all generated solutions, instead of using a programmed or learned evaluator module. This allows to rate states individually or vote across intermediate solutions to select the most promising one to continue with the search. Both mentioned ToT approaches are a generalization of the IO, CoT, and CoT-SC prompting schemes.


Other examples in this class of topologies include \textbf{Thought Decomposition}~\cite{xie2023decomposition} (a multi-prompt scheme based on stochastic beam search and self-evaluation), a scheme by \textbf{Creswell and Shanahan}~\cite{creswell2022faithful} (an extension of the chain-based Selection-Inference~\cite{creswell2022selection}), \textbf{Dynamic Least-to-Most Prompting}~\cite{drozdov2022compositional} (an extension of least-to-most prompting with a tree-based problem decomposition and a dynamic external tree-based few-shot example selection), \textbf{Algorithm of Thoughts (AoT)}~\cite{sel2023algorithm} (a single-prompt approach that utilizes in-context examples formulated in an algorithmic tree-based fashion), \textbf{Tree of Uncertain Thought (TouT)}~\cite{mo2023tree} (an extension of ToT with local ``uncertainty scores'' by incorporating the variance of multiple LLM responses into the state evaluation function), \textbf{Tree-of-Mixed-Thought (TomT)}~\cite{hu2023treeofmixedthought} (a ToT-based reasoning scheme to answer questions on visual scene graphs), or \textbf{Tree of Clarifications (ToC)}~\cite{kim2023tree} (recursive prompting of an LLM to construct a tree of disambiguations for the initial question).

\subsection{Analysis \& Comparison of Designs}

We now broadly discuss and analyze tree designs with respect to different aspects of our blueprint.



\iftr
\subsubsection{Topology \& Its Construction}
\fi
The key novel architectural feature of tree schemes is the {exploration} of a thought, i.e., the ability to generate multiple new steps based on a given single one.
The vast majority of tree schemes are multi-prompt.
Most multi-prompt schemes use a dynamic approach to building the tree topology. The details of how the topology is exactly shaped depend on the specific question.
For most multi-prompt approaches, the user can adapt the tree topology to a certain degree, i.e., by varying the \emph{branching factor} (i.e., the number of thoughts generated from a given vertex) and limiting the depth of the tree.

\iftr
\subsubsection{Representations of Topology \& Schedule}
\fi
We showcase the differences between implicit vs. explicit and single- vs. multi-prompt topologies using representative examples (the tasks are described in detail in the prompts of the figures). We continue with the driving tasks of Game~of~24, Creating Writing, and Mathematical/Logical Reasoning.
For this, we illustrate an implicit single-prompt tree topology elicited by AoT in Figure~\ref{fig:AoT Prompt} as well as two example explicit multi-prompt tree topologies from ToT and CoT-SC in, respectively, Figure~\ref{fig:ToT-CreativeWriting-Prompt} and \ref{fig:CoT-SC_MathReasoning-Prompt}. Lastly, we show an example of a parallel execution schedule in Figure~\ref{fig:SoT Prompt} for SoT~\cite{ning2023skeleton}.

\iftr
\begin{figure*}[t]
\centering
\includegraphics[width=1.0\textwidth]{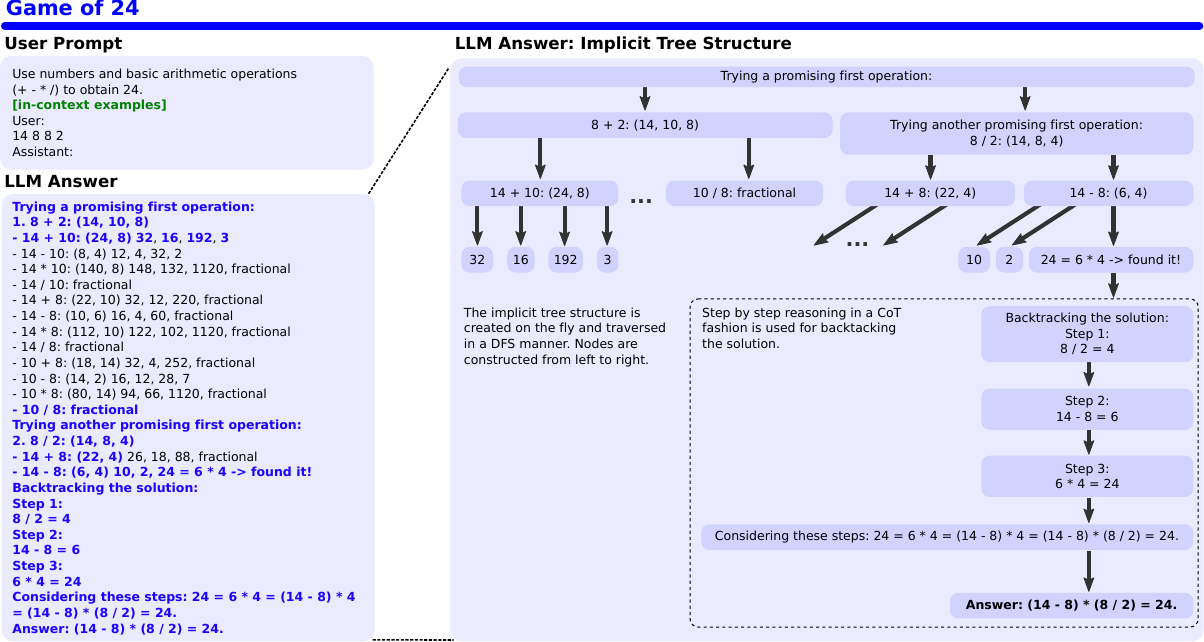}
\caption{\textbf{\textsc{Game of 24.} An example \ul{implicit} \ul{single-prompt} tree topology, encoded with \ul{text}.} It demonstrates a Game of $24$ DFS in-context example from AoT~\cite{sel2023algorithm}. The left view shows the user prompt and the single textual answer from the LLM. The right view shows the implicit tree structure that is explored during the generation of the LLM answer. We mark text corresponding to implicit nodes as \textcolor{blue}{\textbf{bold}}.}
\label{fig:AoT Prompt}
\end{figure*}
\fi

\ifcnf
\begin{figure}[t]
\centering
\includegraphics[width=1.0\columnwidth]{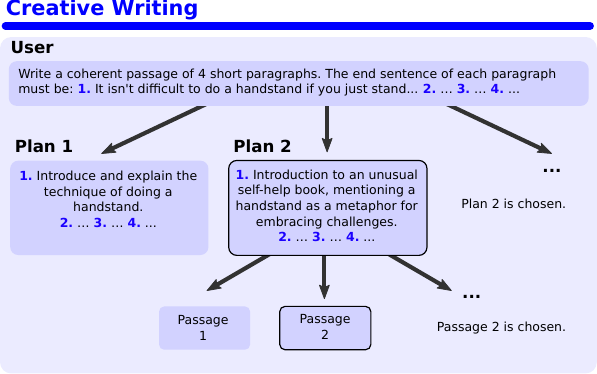}
\vspace{-1.5em}
\caption{\textbf{\textsc{Creative Writing.} An example \ul{explicit} \ul{multi-prompt} tree topology, encoded with \ul{text}}, from the Tree of Thoughts (ToT) scheme \cite{yao2023tree} for creative writing. Given the task of writing a coherent passage of four paragraphs ending in given sentences, first multiple plans (nodes) are generated and then ranked. In a next step, the best plan is used to generate multiple possible passages as outputs. Finally, the best ranked passage is the output of the ToT reasoning.}
\vspace{-1.5em}
\label{fig:ToT-CreativeWriting-Prompt}
\end{figure}
\fi

\iftr
\begin{figure}[t]
\centering
\includegraphics[width=1.0\columnwidth]{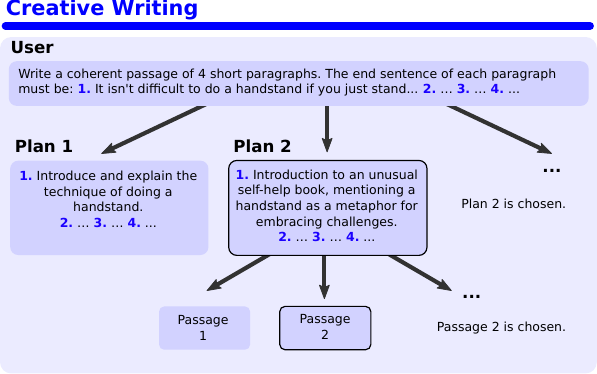}
\caption{\textbf{\textsc{Creative Writing.} An example \ul{explicit} \ul{multi-prompt} tree topology, encoded with \ul{text}}, from the Tree of Thoughts (ToT) scheme \cite{yao2023tree} for creative writing. Given the task of writing a coherent passage of four paragraphs ending in given sentences, first multiple plans (nodes) are generated and then ranked. In a next step, the best plan is used to generate multiple possible passages as outputs. Finally, the best ranked passage is the output of the ToT reasoning.}
\label{fig:ToT-CreativeWriting-Prompt}
\end{figure}
\fi

\iftr
\begin{figure}[t]
\centering
\includegraphics[width=1.0\columnwidth]{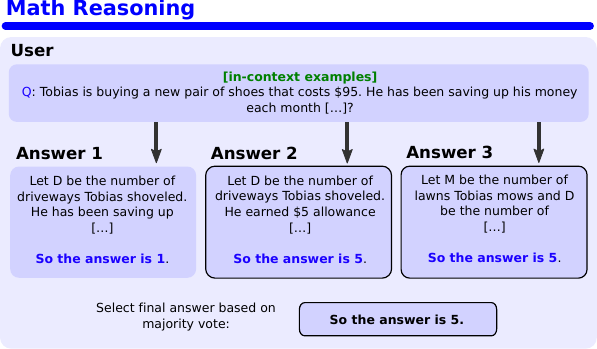}
\caption{\textbf{\textsc{Math Reasoning.} An example using \ul{explicit} \ul{multi-prompt} tree topology, encoded with \ul{text}}. Given a math reasoning task, CoT-SC~\cite{wang2022self} is used to generate multiple answers and pick a final one based on majority vote. Each of the generated answers contains multiple CoT reasoning steps, depicted here in a single node.}
\label{fig:CoT-SC_MathReasoning-Prompt}
\end{figure}
\fi

\iftr
\begin{figure*}[t]
\centering
\includegraphics[width=1.0\textwidth]{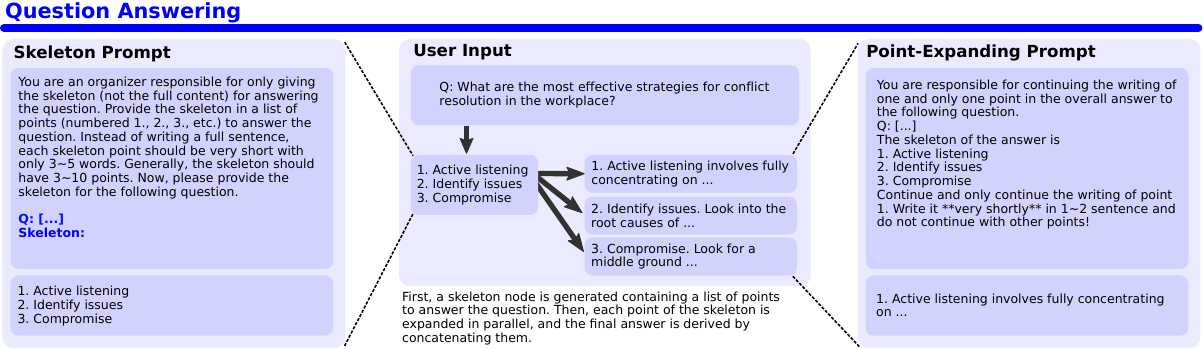}
\caption{\textbf{An example \ul{explicit} \ul{multi-prompt} tree topology, encoded with \ul{text}}. It demonstrates the \ul{automatically} derived tree topology of Skeleton-of-Thought (SoT)~\cite{ning2023skeleton} where the individual points are expanded in \textul{parallel}.}
\label{fig:SoT Prompt}
\end{figure*}
\fi


%

\ifcnf
\addtocounter{figure}{1}
\begin{figure*}[b]
\centering
\vspace{-1.5em}
\includegraphics[width=1.0\textwidth]{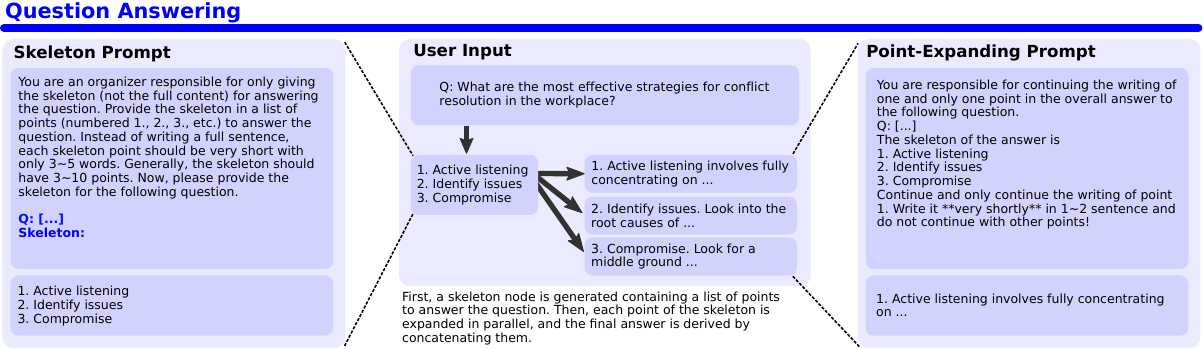}
\vspace{-1.5em}
\caption{\textbf{An example \ul{explicit} \ul{multi-prompt} tree topology, encoded with \ul{text}}. It demonstrates the \ul{automatically} derived tree topology of Skeleton-of-Thought (SoT)~\cite{ning2023skeleton} where the individual points are expanded in \textul{parallel}.}
\label{fig:SoT Prompt}
\end{figure*}

\addtocounter{figure}{-2}
\begin{figure}[t]
\centering
\includegraphics[width=1.0\columnwidth]{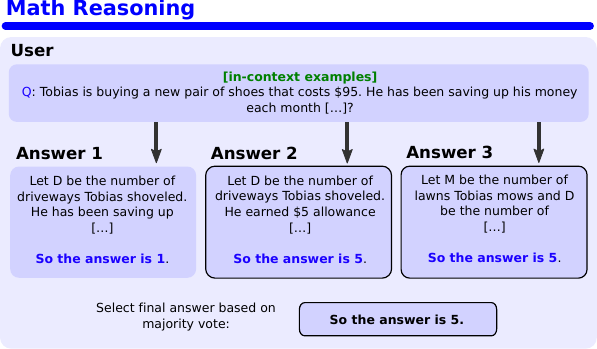}
\vspace{-1.5em}
\caption{\textbf{\textsc{Math Reasoning.} An example using \ul{explicit} \ul{multi-prompt} tree topology, encoded with \ul{text}}. Given a math reasoning task, CoT-SC~\cite{wang2022self} is used to generate multiple answers and pick a final one based on majority vote. Each of the generated answers contains multiple CoT reasoning steps, depicted here in a single node.}
\vspace{-1.5em}
\label{fig:CoT-SC_MathReasoning-Prompt}
\end{figure}
\fi

\iftr
\subsubsection{Performance}

We now summarize performance patterns found within tree topologies. A detailed analysis can be found in Appendix~\ref{sec: tree_extensive}.

\fi
Overall, increasing the {branching factor} (i.e., the number of thoughts generated from a given vertex) often leads to a higher {diversity} of outcomes, which can be beneficial for accuracy, but it also increases \#prompts, i.e., computational cost.
The most advantageous branching factor is hard to find and it often depends on the specific problem to solve. 
Easily decomposable problems may benefit less from more branching than complex problems. Specifically, more complicated problems profit more from decomposing them into many/diverse sub-problems (e.g., this ensures enough diversity for self-consistency to work better). In contrast, a question that has clearly only two sub-parts does not benefit from many more subdivisions, as the additional branches then can be either redundant or wrong.
Single-prompt approaches can perform better on some problems than multi-prompt approaches, while using only a single prompt compared to possibly hundreds~\cite{sel2023algorithm}. 

\section{Reasoning with Graphs}

We also analyze schemes that harness graph topologies, see the bottom part of
\iftr
Table~\ref{tab:schemes} (detailed descriptions of each individual scheme are provided, as for chains and trees, in the appendix).
\fi
Similarly to the tree analysis, we structure the discussion based on the harnessed topology variants, see Figure~\ref{fig:topologies-dets} for details.
We also discuss fundamental concepts introduced or harnessed in these works. Graph schemes, most importantly, introduce \emph{aggregation} (i.e., being able to combine multiple thoughts into a single one).
The purpose behind aggregation is usually \emph{synergy} (i.e., being able to produce an outcome better than the individual ingredients) or an effective \emph{composition of outcomes of tasks}. Graph schemes also use architectural concepts employed in chain or tree schemes, such as exploration or iterative refinement.
We conclude this section with a comparative analysis and illustrations of example representations of graph topologies.

\subsection{Special Classes of Graphs}
\label{sec:l-one-d-trees}

Different schemes harness certain special classes of graphs.
%
%
\textbf{Branch-Solve-Merge (BSM)}~\cite{saha2023branch} employs a 1--level double tree structure to first divide a problem into independently solvable sub-problems, and then combines them into a final solution. The first prompt instructs the LLM to propose sub-problems, which are then solved independently. The final prompt instructs the LLM to merge the results of the sub-problems into a single output.
\textbf{Socratic Questioning}~\cite{qi2023art} is a scheme that models recursive exploration of the thought space using a tree structure. Hereby, the original question is recursively decomposed into sub-tasks until all tasks can be solved with high confidence. These results are then aggregated and propagated back up the tree to answer the original question. This results in an overall double tree reasoning topology.

\subsection{Directed Graphs}
\label{sec:arbitrary-graphs}

Some schemes embrace a general directed graph model.
\textbf{Graph of Thoughts (GoT)}~\cite{besta2023graph} uses a multi-prompt approach to improve the LLM problem solving performance by decomposing a given task into sub-tasks that form a graph. This decomposition is specified as a \emph{Graph of Operations}, which coordinates how the LLM is prompted and how the results are further used in the reasoning process.
%
%
%
The \textbf{Graph of Thought}~\cite{lei2023boosting} presents a multi-prompt approach where a graph of thoughts is constructed recursively in a DFS manner by starting at the
\iftr
\emph{question node} that represents the question to be answered by the LLM.
\fi
From this node, possible reasoning paths are generated by the LLM. For each path, new nodes, i.e., intermediate reasoning steps, are generated by the LLM and are then used to grow the graph.
%
%
\textbf{Graph-of-Thought}~\cite{yao2023beyond} describes a two-stage framework to answer multi-modal questions, i.e., textual questions accompanied by images.
In the first stage, the model generates natural language rationales based on the input text, which provide additional context and knowledge to support answering the given question. This rationale generation is learned as part of the overall model pipeline. In the second stage, these rationales are then appended to the initial question and passed again through the model to predict an answer.
Other schemes in this class include \textbf{Cumulative Reasoning}~\cite{zhang2023cumulative}, \textbf{Everything of Thoughts (XoT)}~\cite{ding2023thoughts}, \textbf{ControlLLM}~\cite{liu2023controlllm}, and \textbf{ResPrompt}~\cite{jiang2023resprompt}.

\subsection{Hypergraphs}

Finally, we also consider a \emph{hypergraph}, which generalizes a graph by enabling edges to connect arbitrary subsets of nodes instead of being links between just two nodes. We include hypergraphs in the taxonomy, because preliminary works already harness them for multi-modal prompting~\cite{yao2023thinking}.
%
%
%
%
Here, \textbf{Hypergraph-of-Thought (HoT)}~\cite{yao2023thinking} is a multi-modal reasoning paradigm modeling the thought process as a hypergraph. First, a graph-of-thoughts as in~\cite{yao2023beyond} is constructed. Then a textual hypergraph is constructed, sharing the same nodes. The hyperedges are then defined as node triples, e.g., "(Lionel Messi, place of birth, Rosario)".
%
%
Additionally, a visual hypergraph-of-thought is constructed by performing $k$-means clustering on image patches, where a cluster corresponds to a hyperedge. Both hypergraphs are then encoded and combined to perform graph learning.

\ifcnf
\begin{figure*}[t]
\centering
\includegraphics[width=1.0\textwidth]{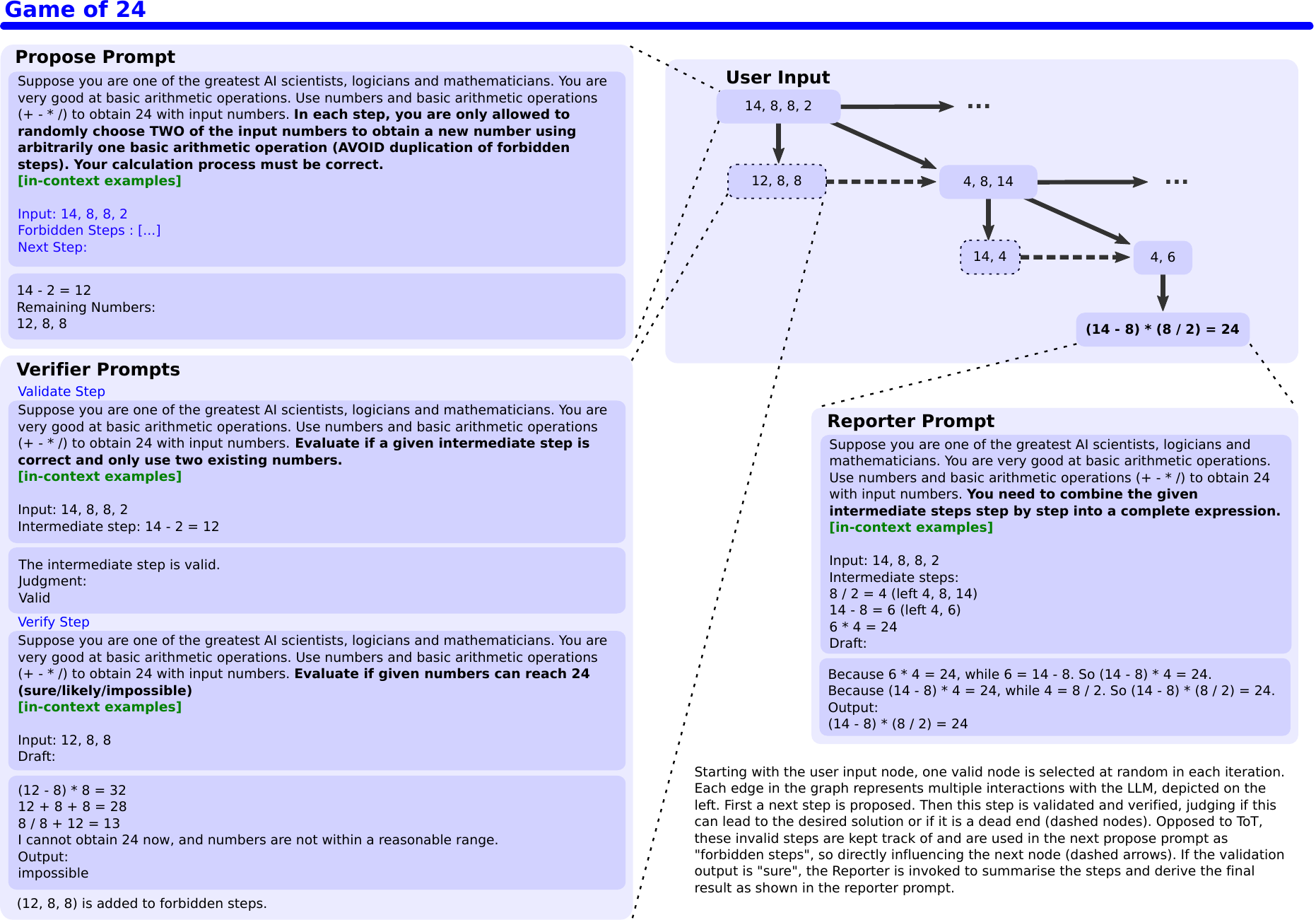}
\vspace{-1.5em}
\caption{\textbf{\textsc{Game of 24.} An example \ul{explicit} \ul{multi-prompt} graph topology, encoded with \ul{text}} using Cumulative Reasoning~\cite{zhang2023cumulative}.}
\label{fig:GameOf24-prompt_CR}
\vspace{-1.5em}
\end{figure*}
\fi

\subsection{Analysis \& Comparison of Designs}

We now broadly discuss and analyze graph designs with respect to different aspects of our blueprint.

\iftr
\subsubsection{Topology \& Its Construction}
\fi
Firstly, the considered schemes exhibit a blend of single- and multi-prompt aspects, allowing for a high degree of flexibility and control over the prompting process. This is evident in the diverse approaches taken by different schemes such as GoT, ControlLLM, and Cumulative Reasoning, each offering unique ways of constructing and utilizing graphs for problem-solving. Secondly, the user's control over the topology of the graph is significant in most schemes, enabling customization of the reasoning process based on specific needs, such as setting branching factors or defining the depth of the graph. Thirdly, the role of the LLM in these graph-based schemes is multifaceted, involving the generation, evaluation, and modification of nodes within the graph, as well as determining the conclusion of the reasoning process. Lastly, there is a notable variation in the degree of user and LLM influence on the topology across different schemes, with some allowing direct user control, while others rely on predefined heuristics or the LLM's decision-making capabilities.

\iftr
\subsubsection{Representations of Topology \& Schedule}
\fi
We now illustrate a representative set of prompts that show different aspects of graph-based prompting topologies, focusing on how the respective graph topologies are encoded inside the prompts (the tasks are described in detail in the prompts of the figures).
\iftr
We illustrate ResPrompt and Cumulative Reasoning (as example implicit single-prompt representations), ControlLLM  (as an example explicit single-prompt representation), and Branch-Solve-Merge (as an example multi-prompt representation).


\fi
Figure~\ref{fig:GameOf24-prompt_CR} shows a prompting example using Cumulative Reasoning~\cite{zhang2023cumulative} for the Game of $24$ with an \emph{explicit} \emph{multi-prompt} graph topology.
Figure~\ref{fig:ResPrompt_MathReasoning-Prompt} shows an in-context example of ResPrompt~\cite{jiang2023resprompt} with a multi-step math question, where the topology is \emph{single-prompt} and \emph{implicit}. Here, an (implicit) edge can be formed by repeating the same token, for example \emph{``earned from his allowance''}, which implies a connection between step 2 and 4 where each step indicates a node.
\iftr
Figure~\ref{fig:CumulativeReasoning_MathReasoning-Prompt} illustrates an in-context example where Cumulative Reasoning~\cite{zhang2023cumulative} is applied to the same math problem with user-specified number of intermediate nodes before reaching the final solution node. Here, \emph{implicit} vertices are formed using numbered positions 1 and 2 on the list, and edges connect points 1 and 2 to point~3.
\fi
In contrast to the implicit representation of topology, Figure~\ref{fig:ControlLLM_JSON-Prompt} shows an example from ControlLLM~\cite{liu2023controlllm} to represent the topology \emph{explicitly} with a JSON format, but also in the \emph{single-prompt} setting. Finally, Figure~\ref{fig:BSM-prompt} depicts a \emph{multi-prompt} example from Branch-Solve-Merge (BSM)~\cite{saha2023branch} for story generation. 
\ifcnf
\ifnonb
Additional prompting examples can be found in the extended technical report.
\fi
\fi

\iftr
\begin{figure*}[t]
\centering
\includegraphics[width=1.0\textwidth]{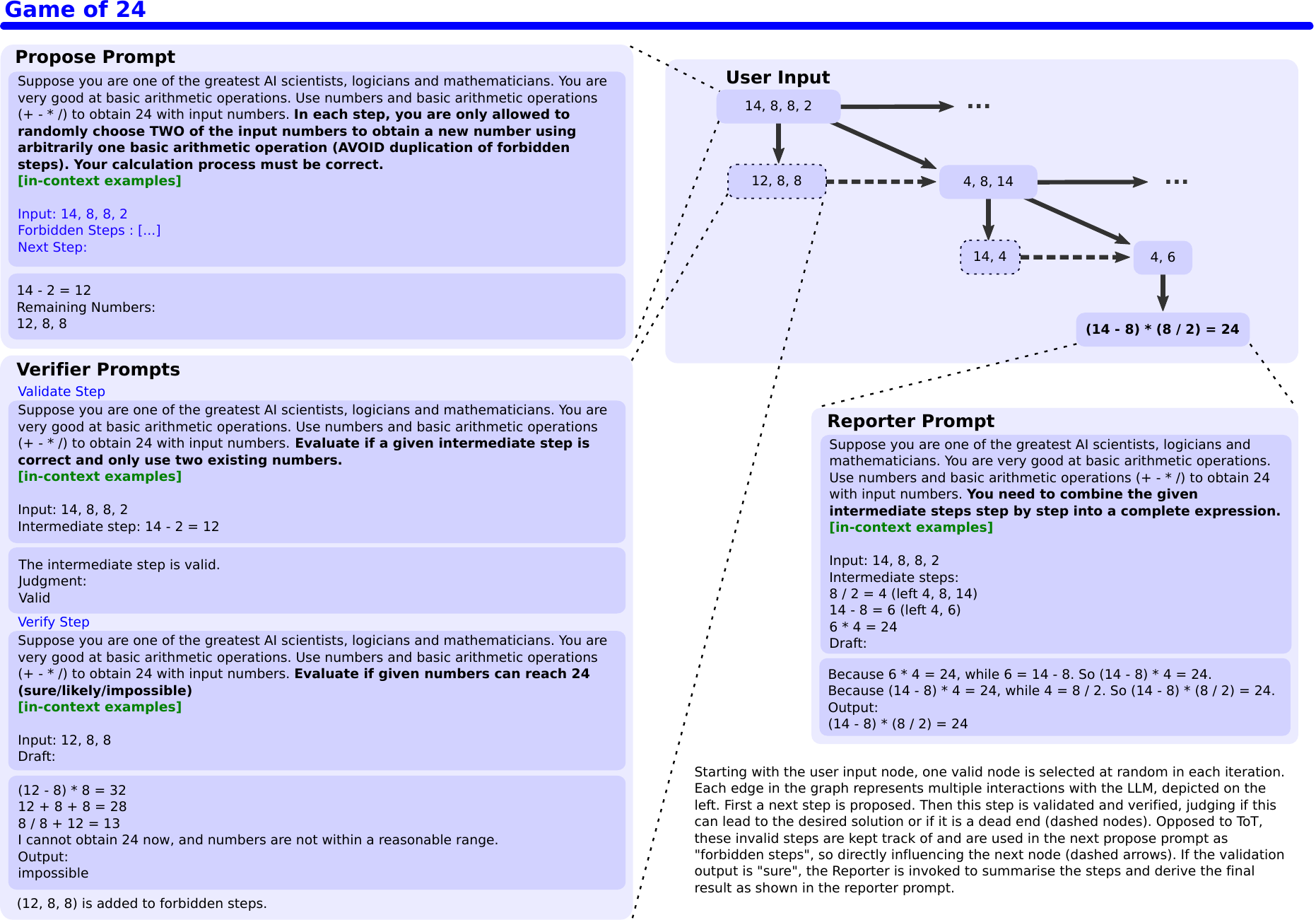}
\caption{\textbf{\textsc{Game of 24.} An example \ul{explicit} \ul{multi-prompt} graph topology, encoded with \ul{text}} using Cumulative Reasoning~\cite{zhang2023cumulative}.}
\label{fig:GameOf24-prompt_CR}
\vspace{3em}
\end{figure*}
\fi

\iftr
\begin{figure*}[!h]
\centering
\includegraphics[width=\textwidth]{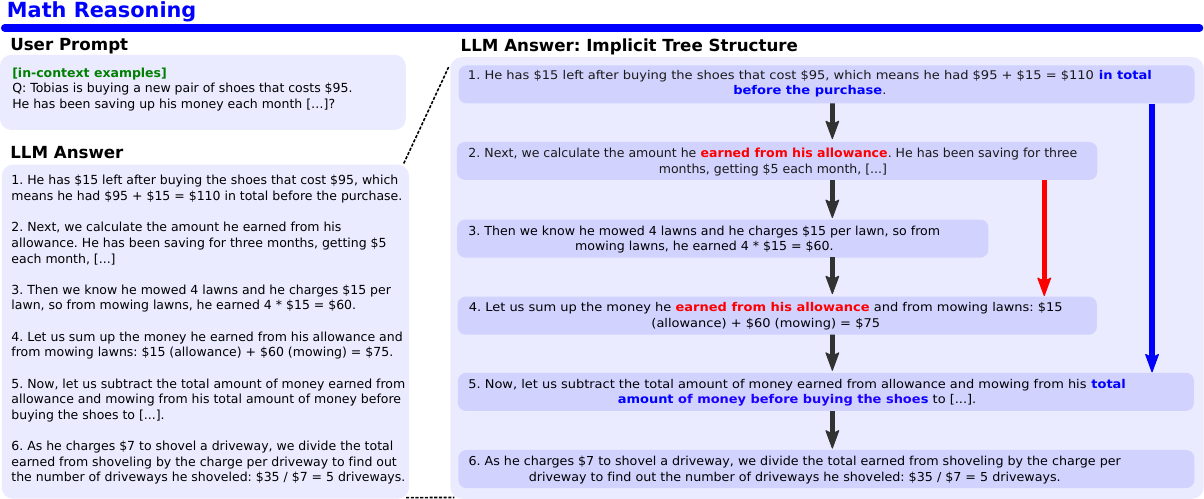}
\caption{\textbf{\textsc{Math Reasoning.} An example \ul{implicit} \ul{single-prompt} graph topology, encoded with \ul{text}}. It shows an in-context example of a math question from ResPrompt~\cite{jiang2023resprompt}, representing a linear sequence of six connected nodes with two implicit edges of the graph topology, marked with two different colors (\textcolor{red}{\textbf{red}} and \textcolor{blue}{\textbf{blue}}), together with their corresponding nodes.}
\label{fig:ResPrompt_MathReasoning-Prompt}
\end{figure*}
\fi

\iftr
\begin{figure*}[!h]
\centering
\includegraphics[width=\textwidth]{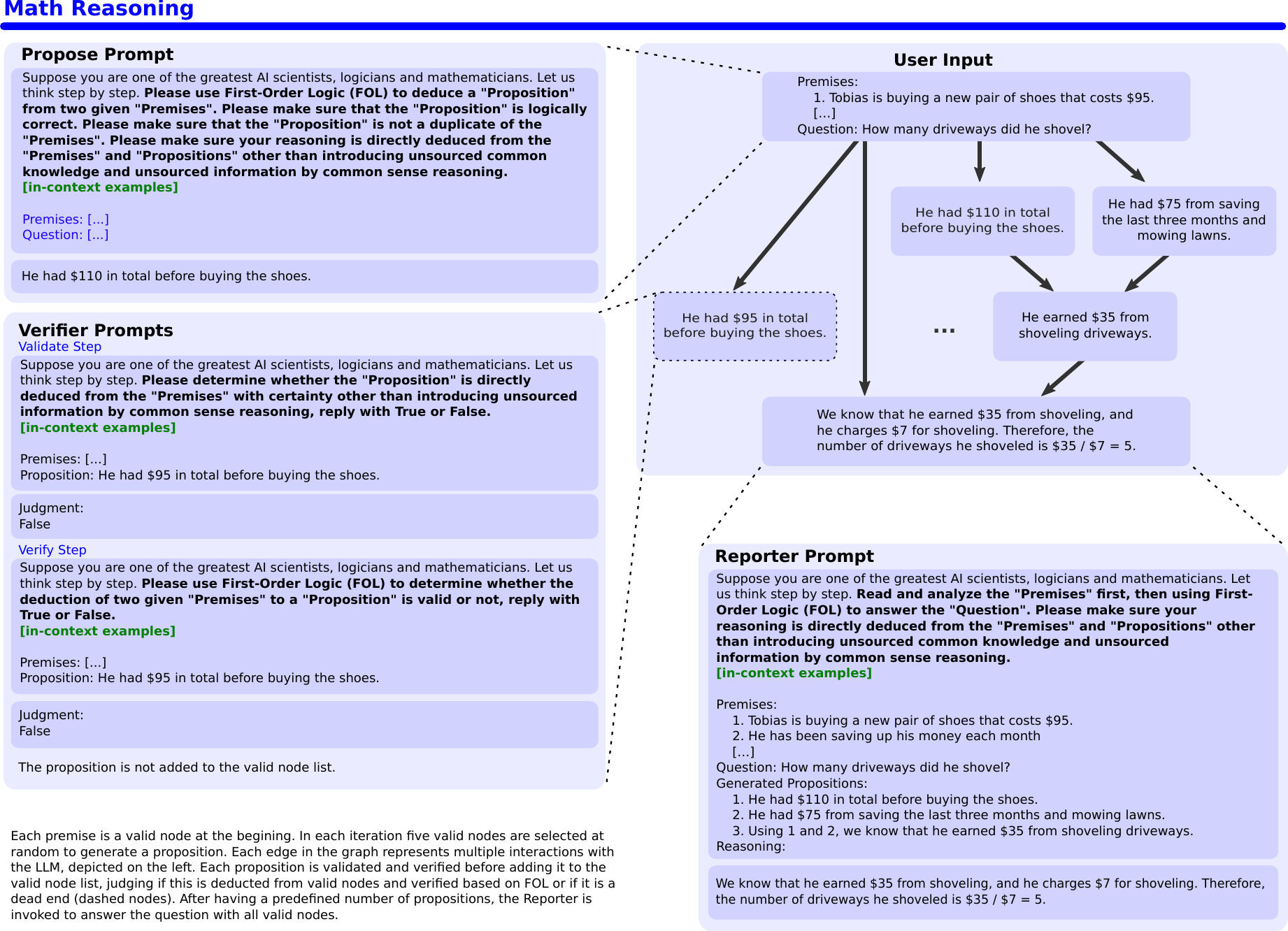}
\caption{\textbf{\textsc{Math Reasoning.} An example \ul{implicit} \ul{single-prompt} graph topology, encoded with \ul{text}}, for the same question as in Figure~\ref{fig:ResPrompt_MathReasoning-Prompt}, but using Cumulative Reasoning~\cite{zhang2023cumulative}. It illustrates three intermediate proposition nodes with direct connections from the first two nodes to the third one. The reasoning part indicates the final solution node, with direct connections from the third proposition and input nodes.}
\label{fig:CumulativeReasoning_MathReasoning-Prompt}
\end{figure*}

\fi


\iftr
\begin{figure}[!h]
\centering
\includegraphics[width=\columnwidth]{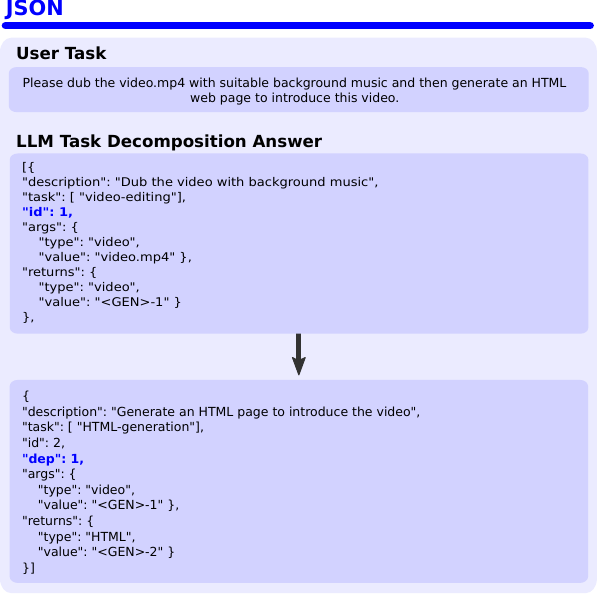}
\caption{\textbf{\textsc{JSON.} An example \ul{explicit} \ul{single-prompt} graph topology, encoded with \ul{JSON}}, based on the ControlLLM scheme~\cite{liu2023controlllm} for task decomposition. It shows two nodes describing decomposed subtasks for solving a given task. The "dep" field refers to dependent tasks, showing there is a direct edge from the first node (task 1) to the second.}
\label{fig:ControlLLM_JSON-Prompt}
\end{figure}
\fi

\iftr
\begin{figure}[!h]
\centering
\includegraphics[width=\columnwidth]{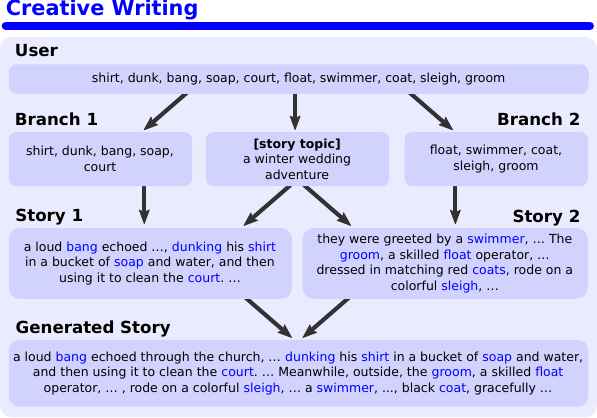}
\caption{\textbf{\textsc{Creative Writing.} An example \ul{multi-prompt} graph topology, encoded with \ul{text}}, from the Branch-Solve-Merge scheme \cite{saha2023branch} for story generation with branch, solve, and merge prompts. Given a list of concepts as input, the branch module generates three child nodes: two groups of concepts and one topic node. The solve module then creates two story nodes based on each group of concepts and the topic. Finally, these two story nodes are merged into the final solution node.}
\label{fig:BSM-prompt}
\end{figure}
\fi

\iftr
\subsubsection{Performance}
\fi
The considered works universally show improvements in effectiveness of graph-based prompting schemes over chains and trees across various tasks, suggesting a promising direction for future research and application in the field of AI and machine learning.


\ifcnf
\begin{figure*}[t]
\centering
\includegraphics[width=\textwidth]{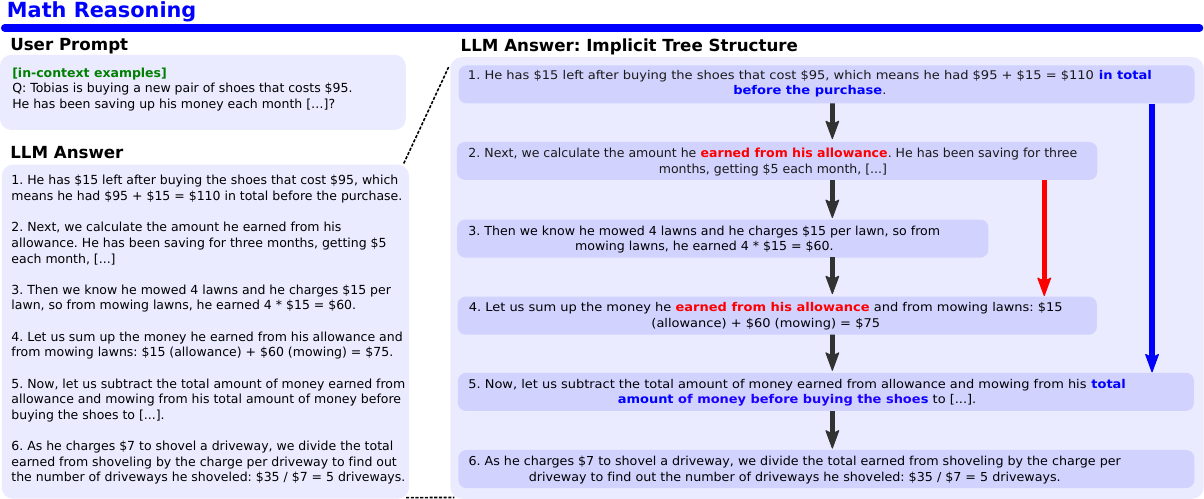}
\vspace{-1.5em}
\caption{\textbf{\textsc{Math Reasoning.} An example \ul{implicit} \ul{single-prompt} graph topology, encoded with \ul{text}}. It shows an in-context example of a math question from ResPrompt~\cite{jiang2023resprompt}, representing a linear sequence of six connected nodes with two implicit edges of the graph topology, marked with two different colors (\textcolor{red}{\textbf{red}} and \textcolor{blue}{\textbf{blue}}), together with their corresponding nodes.}
\vspace{-1.5em}
\label{fig:ResPrompt_MathReasoning-Prompt}
\end{figure*}

\begin{figure}[t]
\centering
\includegraphics[width=\columnwidth]{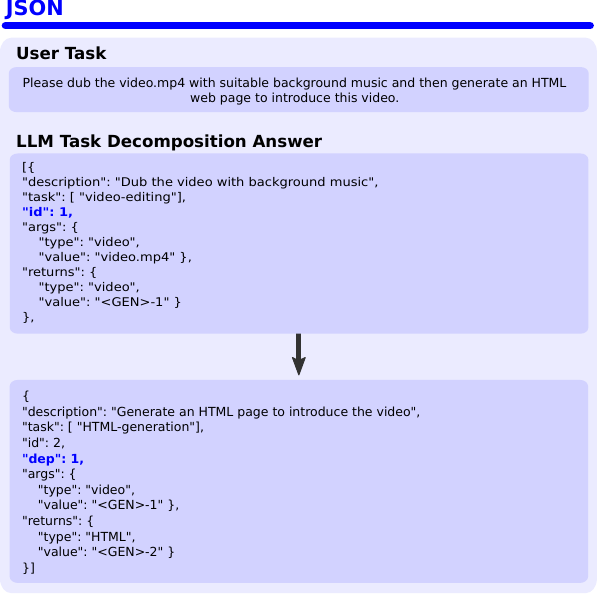}
\vspace{-1.5em}
\caption{\textbf{\textsc{JSON.} An example \ul{explicit} \ul{single-prompt} graph topology, encoded with \ul{JSON}}, based on the ControlLLM scheme~\cite{liu2023controlllm} for task decomposition. It shows two nodes describing decomposed subtasks for solving a given task. The "dep" field refers to dependent tasks, showing there is a direct edge from the first node (task 1) to the second.}
\vspace{-2em}
\label{fig:ControlLLM_JSON-Prompt}
\end{figure}

\begin{figure}[t]
\centering
\includegraphics[width=\columnwidth]{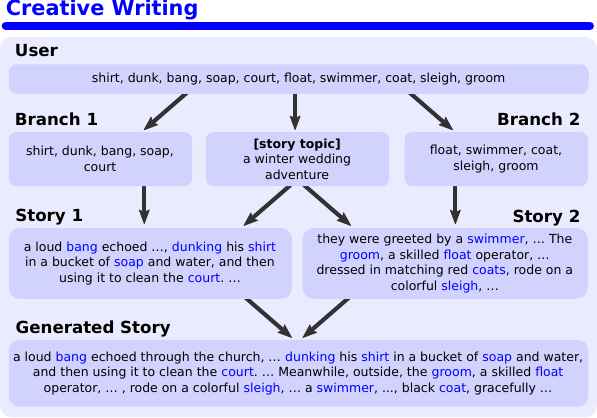}
\vspace{-1.5em}
\caption{\textbf{\textsc{Creative Writing.} An example \ul{multi-prompt} graph topology, encoded with \ul{text}}, from the Branch-Solve-Merge scheme \cite{saha2023branch} for story generation with branch, solve, and merge prompts. Given a list of concepts as input, the branch module generates three child nodes: two groups of concepts and one topic node. The solve module then creates two story nodes based on each group of concepts and the topic. Finally, these two story nodes are merged into the final solution node.}
\vspace{-1.5em}
\label{fig:BSM-prompt}
\end{figure}
\fi

\section{Chains vs.~Trees vs.~Graphs of Thoughts}

We also broadly discuss tradeoffs, commonalities, and differences between the three fundamental classes of topologies: chains, trees, and graphs.

The novelty in chain-based prompting lies in introducing explicit intermediate LLM thoughts between the input and the output. This linear sequence of thoughts guides the LLM in a step-by-step manner towards the solution, enhancing the clarity and traceability of the reasoning process. Beyond this, tree-based schemes bring the possibility to explore several next-step variants at each juncture, allowing the LLM to evaluate multiple pathways and select the most promising one. This branching structure facilitates a broader exploration of potential solutions. Graph-based schemes, however, represent the most complex structure, offering an arbitrary reasoning framework. They enable the aggregation of various reasoning steps into a synergistic solution, allowing for non-linear and multifaceted problem-solving approaches. This diversity in structures reflects the varying needs and complexities of tasks that LLMs are expected to handle.

In terms of cost-effectiveness, chain-based prompting schemes generally outperform their tree-based and graph-based counterparts. This efficiency largely stems from their single-prompt nature, which requires fewer computational resources and less processing time. Tree and graph-based schemes, on the other hand, often involve multiple rounds of prompting to explore the various branches or connections in their respective structures, naturally incurring higher costs. However, this is not an inherent limitation of these schemes. Future developments in tree and graph-based prompting should aim to better encode the tree and graph structures within a single prompt. Such advancements could potentially combine the cost-effectiveness of single-prompt schemes with the enhanced quality and complexity management offered by tree and graph-based structures.

When it comes to the quality of outcomes, tree-based and graph-based prompting schemes typically surpass chain-based ones. This superior performance is attributed to their ability to explore a wider range of potential transformations at each step, going beyond the linear refinement seen in chains. Tree-based schemes, with their branching paths, offer multiple avenues for solution exploration at each decision point, leading to a more comprehensive search for the optimal outcome. Graph-based schemes take this a step further by allowing for an even more diverse set of connections and interactions between different steps in the reasoning process. This flexibility enables these schemes to handle more complex and multifaceted tasks effectively, leading to higher-quality outcomes in scenarios where simple linear reasoning is insufficient.

\subsection{Performance vs.~Task \& Data Categories}

\textbf{Task and Data Categories }
We synthesize an evaluation of considered schemes, based on a broad literature overview. In general, the evaluation spans arithmetic, commonsense, symbolic, and specialized reasoning tasks. \textbf{Arithmetic reasoning} datasets like GSM8K~\cite{cobbe2021gsm8k}, SVAMP~\cite{patel2021nlp}, MAWPS~\cite{koncel2016mawps}, and DROP~\cite{dua2019drop} involve problems requiring sequential numerical reasoning, where structured decomposition is beneficial. \textbf{Commonsense reasoning} datasets include StrategyQA~\cite{geva2021strategyqa}, 2WikiMultiHopQA~\cite{ho2020constructing}, and MuSiQue~\cite{trivedi2022musique}; they emphasize multi-hop reasoning. \textbf{Symbolic reasoning}, including last-letter concatenation and spatial tasks (Brick World~\cite{hu2023chain}, SPARTUN~\cite{mirzaee2022transfer}), tests pattern recognition and out-of-distribution generalization. \textbf{Specialized reasoning} tasks, such as code generation (MBPP~\cite{austin2021program}, HumanEval~\cite{chen2021evaluating}) and causal inference (CauseEffect~\cite{srivastava2023beyond}, SocialQA~\cite{srivastava2023beyond}), benefit from structured planning and external tool use. We now describe how different reasoning topologies compare in these categories of benchmarks.

\textbf{Arithmetic Reasoning Winners: ToT, GoT}
ToT outperforms CoT by enabling structured decomposition and backtracking, making it particularly effective in datasets like GSM8K and DROP.
GoT further improves performance by integrating cumulative reasoning across steps thanks to its aggregation capabilities, reducing error propagation.
CoT struggles with multi-step arithmetic reasoning problems due to error accumulation in sequential reasoning.

\textbf{Commonsense Reasoning Winners: GoT, ToT}
GoT is highly effective in multi-hop and retrieval-based reasoning (e.g., 2WikiMultiHopQA, MuSiQue) by dynamically aggregating knowledge across multiple nodes.
ToT excels in structured multi-hop tasks (e.g., Decomposed Prompting in StrategyQA) by breaking down inference into manageable chunks.
CoT is less reliable in tasks requiring multiple layers of inference, as sequential chains cannot effectively retrieve and synthesize distant context.

\textbf{Symbolic Reasoning Winners: ToT, GoT}
ToT provides superior performance in structured symbolic tasks such as last-letter concatenation, object counting, and coin flipping by systematically exploring the solution space.
GoT enhances performance in logical and mathematical transformations by maintaining dependencies between reasoning steps.
CoT fails to generalize well in OOD cases due to its linear reasoning constraints.

\textbf{Specialized Reasoning Winners: GoT, ToT}
%
%
GoT is ideal for complex reasoning pipelines (e.g., ScienceQA~\cite{lu2022learn}) by allowing nodes to interact flexibly across multiple steps, making it particularly strong in planning-based tasks.
ToT enhances structured reasoning in mathematical and coding tasks (e.g., Reflexion~\cite{shinn2023reflexion} in HumanEval, MBPP) by iteratively refining intermediate outputs.
CoT is limited in these types of tasks due to its inherently sequential reasoning pattern.

\textbf{Sequential Reasoning Winners: CoT, ToT}
CoT performs well in structured, non-branching tasks (e.g., simple math problems, single-hop commonsense tasks) due to its interpretability and efficiency.
ToT provides additional benefits in cases where intermediate verification is needed, preventing early reasoning errors.
GoT is unnecessary for purely sequential problems and adds computational overhead without significant performance gains.

\textbf{Summary }
Chain-based methods are effective for sequential reasoning, but struggle with complex decompositions and multi-hop inference. Tree-based approaches improve on these limitations by allowing structured problem-solving via multi-step decomposition and exploration, with user-defined constraints such as branching factor, depth limits, and search strategies (BFS, DFS, beam search). ToT~\cite{yao2023tree} adaptively prunes search trees, while Thought Decomposition~\cite{xie2023decomposition} dynamically expands reasoning depth for improved accuracy at the cost of increased token usage. Though computationally costly, tree-based schemes excel in problems with hierarchical structure, outperforming chain-based ones in arithmetic and symbolic reasoning. Graph-based methods further generalize tree-based approaches by enabling non-hierarchical, interconnected reasoning, allowing nodes to influence each other dynamically. Graph schemes like GoT~\cite{besta2023graph} improve arithmetic and multi-hop reasoning efficiency, often using fewer tokens than tree-based methods.

\subsection{Evaluation Techniques}

Works on evaluating methodologies for reasoning topologies has so far focused on CoTs.
Chain-based reasoning requires methods that assess not just final answer
correctness but also the validity of intermediate reasoning steps. Nguyen et
al.~\cite{nguyen2024direct} show that LLMs frequently generate plausible
yet incorrect reasoning chains, necessitating structured verification
mechanisms. The OCEAN framework~\cite{wu2024ocean} formalizes CoT
evaluation as a Markov Decision Process (MDP), using knowledge graph-based
inverse propensity scores (KG-IPS) to assess reasoning fidelity. Meanwhile,
Chang et al.~\cite{zhang2024small} highlight the necessity of robust
verification for self-correction, showing that weak verifiers hinder accuracy
improvements even when models are fine-tuned for reasoning correction.
Additionally, discriminative evaluation methods assess LLMs' ability to identify
logical inconsistencies in reasoning paths, while generative evaluation tests
their capacity to produce structured and verifiable reasoning steps. These
studies collectively suggest that effective evaluation of chain methods must integrate
logical consistency checks, structured reasoning assessments, and robust
verification models to ensure reasoning chains genuinely support correct
answers.

\section{Design Architectures}

We also analyze design tradeoffs. We consider the following aspects: 
design architecture (what the architecture-level decomposition of different schemes into interacting modules is), 
productivity \& programmability (how well different schemes support efforts such as new extensions),
and scalability as well as parallelizability (whether considered schemes can be parallelized and whether they scale well).
%

\subsection{Design Architecture}
%
%
Here, we analyze how the overall prompting scheme, combining the LLM and some additional logic, decomposes into different modules. The module-level design architecture is explicitly discussed to a certain degree by some tree-based approaches, such as~\cite{creswell2022faithful, hu2023treeofmixedthought, long2023large, mo2023tree, yao2023tree} as well as graph-based approaches such as~\cite{besta2023graph, lei2023boosting, qi2023art, saha2023branch, yu2023thought, zhang2023cumulative}.
Detailed module-level architectures are presented by Long~\cite{long2023large}, Yao et al.~\cite{yao2023tree}, and Besta et al.~\cite{besta2023graph}.

The general architecture of considered schemes can be summarized as consisting of the following modules: a \emph{generator}, an \emph{evaluator}, a \emph{halter}, and a \emph{controller} module. The generator prompts the LLM to produce further reasoning steps given the current reasoning step and some context. The evaluator rates the current reasoning step, depending on the implementation this rating can depend on other states, the path to the initial question node, or some other context. The halter determines if a suitable solution has been found and how it should be reported or if the search should continue. The controller module coordinates the other modules as well as the construction and exploration of the tree/graph. 
%

\subsection{Productivity \& Programmability}

%




ToT by Yao et al.~\cite{yao2023tree}, Tree Prompting~\cite{morris2023tree} and SoT~\cite{ning2023skeleton} provide implementations that can be used directly for custom tasks; ToT~\cite{yao2023tree} is the only tree approach that provides an easily usable API to solve custom problems in a multi-prompt fashion. ToT by Long~\cite{long2023large}, Thought Decomposition~\cite{xie2023decomposition} and ToC~\cite{kim2023tree} provide implementations to reproduce their results, but can not easily be used for other tasks. CoT-SC~\cite{wang2022self} is a paradigm and can easily be implemented when needed, AoT~\cite{sel2023algorithm} is a single-prompt scheme and does not provide an implementation but lists in-context examples. The remaining approaches \cite{creswell2022faithful, drozdov2022compositional, hu2023treeofmixedthought, mo2023tree} have no code published at this time.

%
In graph designs, Socratic Questioning~\cite{qi2023art} and Graph of Thoughts (GoT)~\cite{besta2023graph} provide implementations that can be used directly for custom tasks. Cumulative Reasoning~\cite{zhang2023cumulative} provides an implementation to reproduce their results, but cannot easily be used for other tasks. Thought Propagation (TP)~\cite{yu2023thought}, Branch-Solve-Merge (BSM)~\cite{saha2023branch}, ControlLLM~\cite{liu2023controlllm} and ResPrompt~\cite{jiang2023resprompt} do not provide implementation but list a set of prompting examples. Among these, ResPrompt~\cite{jiang2023resprompt} is a paradigm can be easily implemented when needed. The remaining approaches~\cite{ding2023thoughts, lei2023boosting, yao2023thinking, yao2023beyond} have no code published at this time.

Overall, GoT~\cite{besta2023graph} offers the most general API for addressing custom tasks in a multi-prompt fashion.
It offers a principled way to design a prompting structure, through \emph{thought transformations}, i.e., ways to transform GoT nodes into new outcomes.
Example GoT API routines, which implement {thought transformations} include -- for example
\emph{Generate(prompt, k=4)} (prompt the LLM to produce k answers to a given prompt),
\emph{Generate(prompt, k=1)+Repeat(k=4)} (generate 4 context isolated responses of the LLM),
\emph{Aggregate(thought1, thought2)} (combine two thoughts),
\emph{KeepBest(N=1)} (query the LLM to return N best results),
\emph{Improve(thought)} (query the LLM to improve the result using information from another thought, e.g., input list and currently sorted list $\rightarrow$ try to fix incorrectly sorted elements).

\subsection{Scalability \& Parallelizability}

Only a few works address scalability and parallelizability improvements of LLMs.
Skeleton of Thought~\cite{ning2023skeleton} uses the ToT approach: the prompt query is a tree of depth one and all leaves can be processed \textbf{in parallel}.
Batch Prompt~\cite{lin2023batchprompt} improves the performance by \textbf{batching} different data points into a single prompt.
On the other hand, Lookahead Decoding~\cite{fu2023lookahead} is focused on \textbf{speculatively} guessing tokens to speed up the answer generation. We now discuss in more detail how these concepts are realized.

{Skeleton of Thought}~\cite{ning2023skeleton} tackles the problem of high inference latency in LLMs by addressing the sequential decoding problem (generating answer tokens one at a time).
They proposed a new approach where a \emph{skeleton} prompt leads the LLM to first generate a succinct list of points that form the skeleton of its future answer.
Then, each point in the skeleton is decoded in parallel, using batching whenever possible.
The method does not require changes to LLMs and can be applied to off-the-shelf models.
with reported speedups of up to 2.69$\times$.
However, this method ignores dependencies between points in the skeleton answer, and the authors propose replacing the tree with a graph model as future work.

{Lookahead Decoding}~\cite{fu2023lookahead} extends the idea of \emph{speculative decoding} where a sequence of future tokens is predicted, and the LLM later verifies each one of them in parallel.
This method represents the sequential generation of a chain of responses as a non-linear system of equations and then solves it with the Jacobi iteration method.
In each iteration step, at least one guessed token - the next one - is verified and matched successfully.
Furthermore, the method caches \emph{n-grams} generated for each token by inspecting $N$ prior Jacobi iterations, allowing it to decode multiple tokens on a positive match.
By conducting the lookahead and verification in parallel, authors achieve decoding speedup by up to 2.25$\times$, at the cost of an exponential increase in GPU FLOPs.

{BatchPrompt}~\cite{lin2023batchprompt} is a scheme that batches similar questions into a single prompt to reduce the overhead of task description and few-shot examples compared to single-question prompting.
However, naive batching decreases the overall LLM performance, and the final accuracy depends on the position of each single prompt within the batch, which may be challenging to tune.
BatchPrompt avoids this problem by performing majority voting on multiple rounds of prompting with different permutations of the question, allowing for competitive results compared to single-question prompting.
By using large batches with few voting rounds, the scheme can process queries with significantly fewer calls to the LLM.


\ifcnf
\vspace{-1em}
\fi
\section{Foundations \& Theory}

There are a few preliminary works that attempt to provide foundations for structured prompting.

{Madaan and Yazdanbakhsh~\cite{madaan2022text}} focus on \textbf{dissecting CoT into three basic components}, namely \emph{symbols} (sequences of tokens used as the basis of the LLM reasoning, for example \ul{numbers to be sorted}), \emph{patterns} (the structure within prompt that reinforces task understanding, for example the \ul{order} of numbers to be sorted), and \emph{text} (any tokens that are \ul{neither} a symbol \ul{nor} a pattern).
Here, ``patterns'' is effectively a certain single-prompt topology that enhances the LLM reasoning.
The authors discover that, for example, patterns are relevant for better task understanding, and they form a symbiotic relationship with text: the latter help to create more useful patterns while the former enable the LLM to generate text that helps in task solving.

{Tutunov et al.~\cite{tutunov2023can}} use \textbf{probabilistic graphical models} to provide a theoretical understanding of how and why LLMs are able to generate a coherent chain of thoughts.
Their model generalizes Jiang's latent space theory of language models~\cite{jiang2023latent} to chains of thoughts. In the latent space theory, thoughts convey a hidden \emph{intention}. Because the intention is not directly observed -- only the uttered thought -- there is a potential for ambiguity. This \emph{ambiguity} $\rho$ can be quantified probabilistically for the language as a whole. 
Tutonov et al. extend Jiang's model by adding an additional \emph{hidden context} variable upon which intentions are conditioned. This context ${c}$ defines a family of tasks which each define a set of coherent reasoning steps. Moreover, Tutonov et al. generalize the one-shot generation of thoughts into a chain of thoughts with an associated \emph{chain of intentions}. Every intention is conditioned on the context $c$ and the previous intentions. In turn, the $i$-th thought in the chain is conditioned on the $i$-th intention. Their main result is as follows: conditioned on the input and a sequence of $N$ example chain of thoughts generated from a context $c$, the probability that an LLM assigns to a chain of thoughts approaches the true probability of the chain given the input and the hidden context $c$ with convergence speed $\rho^N$. 
This means that, with the appropriate examples, LLMs can generate a chain of thoughts that is arbitrarily close to the true output.
%
%

{Besta et al.~\cite{besta2023graph}} discuss \textbf{tradeoffs between latency}
(number of steps to reach the final thought) \textbf{and volume}, which they define -- for a given thought $t$ -- as \textit{the number of preceding LLM thoughts that could have impacted thought $t$}.
Formally, it is the number of thoughts in the topology from which there exists
a path to thought $t$. They assume a time of $O(1)$ for each reasoning
step, fix the total cost for each topology in their analysis to $\Theta(n)$ and
also make certain assumptions for each topology. For a single chain both the
latency and the volume have a high value of $N$, which for multiple chains ($k$
independent chains) is reduced by $k$ for both metrics ($N/k$). In their tree
analysis they assume a complete $k$-ary tree, which has a low latency of $\lceil\log_k
N \rceil$, but the volume is similarly low ($O(\log_k N)$). For graphs, they assume two
complete $k$-ary trees, where the first tree, representing the division into
sub-tasks, is joined at the leaf level with another ``reversed'' $k$-ary tree
(including its edges), representing the aggregation into the final solution.
They conclude that double-tree graphs provide the best tradeoff with a latency of $\log_k N$
and a volume of $N$.

Several works, while not providing any theoretical underpinning for structured-enhanced prompting, investigate the capabilities for CoT to conduct formal analyses. This includes GSM8K~\cite{cobbe2021gsm8k}, ProofWriter~\cite{tafjord2021proofwriter}, FOLIO~\cite{han2022folio}, SimpleLogic~\cite{zhang2022paradox}, and PrOntoQA~\cite{saparov2022language}.

There have been several works that investigate the theoretical underpinning of general in-context learning. This includes Xie et al.~\cite{xie2022explanation}, Wies et al.~\cite{wies2023learnability}, Hahn and Goyal~\cite{hahn2023theory}, and Jiang~\cite{jiang2023latent}.
However, as they do not focus on the topologies of prompting, details are outside the focus of our work.

To align theoretical foundations with practical techniques, we illustrate how structured-enhanced prompting principles translate into real-world applications. Madaan and Yazdanbakhsh's decomposition of CoT into symbols, patterns, and text informs prompt engineering strategies, such as designing math prompts where number alignment (patterns) improves model understanding, or crafting logic puzzles where structured reasoning cues guide CoT outputs. In arithmetic reasoning (e.g., GSM8K), explicitly formatting step-by-step solutions with structured separators enhances model performance by leveraging predictable reasoning patterns. Tutunov et al.'s latent space theory extension provides a foundation for designing few-shot prompting methods, where increasing high-quality exemplars within the prompt minimizes ambiguity and maximizes coherence in reasoning, which is critical in commonsense tasks like StrategyQA. Their model also justifies iterative refinement techniques such as self-correction loops (e.g., Reflexion~\cite{shinn2023reflexion}), where LLMs re-evaluate their reasoning chain steps by conditioning future reasoning on past outputs. Besta et al.'s trade-off analysis directly supports task-specific topology selection, showing that tree-based prompting is ideal for hierarchical problem-solving (e.g., Least-to-Most Prompting~\cite{zhou2022least} in multi-step arithmetic) while graph-based approaches best suit compositional reasoning (e.g., multi-hop retrieval tasks like HotPotQA~\cite{yang2018hotpotqa}). Their latency vs. volume model further informs computationally efficient prompting strategies, helping select whether sequential chain-based prompting (low compute) or graph-structured prompting (better recall at higher cost) is preferable for real-world deployments, such as chatbot-based tutoring systems or automated theorem proving. Finally, benchmark-driven studies on GSM8K, ProofWriter, and FOLIO validate these structured prompting paradigms, demonstrating how carefully designed prompt hierarchies improve logical correctness in formal reasoning tasks. 

\ifcnf
\vspace{-1em}
\fi
\section{Research Opportunities}

We now review future research directions in structure-enhanced prompting.






    



\textbf{Exploring New Topology Classes}
An intriguing related approach is the study of novel \emph{topology classes}, such as hypergraphs, in prompting schemes. The existing HoT scheme~\cite{yao2023thinking} primarily uses triples, barely scratching the surface of what hypergraphs can offer in terms of their connectivity structure, where hyperdges can connect arbitrary subsets of vertices. Future research could delve into how hypergraphs can accommodate more complex relationships in data such as motifs~\cite{besta2022motif}, dense subgraphs~\cite{besta2021graphminesuite}, cliques~\cite{besta2022probgraph, gianinazzi2021parallel, strausz2022asynchronous}, and others, potentially leading to breakthroughs in LLM's understanding and reasoning capabilities. 

\textbf{Explicit Representations in Single-Prompt Settings}
The explicit representation of prompting topologies has been largely unexplored, especially in scenarios involving single prompts. Research in this area could focus on how different representations, such as Adjacency Lists, Adjacency Matrix, or numerous others~\cite{besta2018log}, can be employed to enhance the efficiency and effectiveness of LLMs. This exploration is vital, considering the impact of different data structures on the processing and interpretation of information by LLMs. The challenge lies in integrating these complex representations into a single-prompt format without compromising the simplicity and accessibility of the models.

\textbf{Automatic Derivation of Tree and Graph Topologies}
The majority of current tree and graph topologies in LLM prompting are created manually or semi-automatically. A promising research direction is the development of methodologies for the automatic derivation of these topologies. Automating this process could significantly reduce the time and effort required to configure LLMs for specific tasks, making them more accessible and efficient. This automation might involve leveraging machine learning algorithms to identify and implement the most effective topologies based on the nature of the task and the data involved.

\textbf{Advancements in Single-Prompt Schemes}
While there are a few examples of single-prompt schemes, such as the initial Chain-of-Thought (CoT), there's much room for improvement. Future research could focus on how to encapsulate more complex and detailed tree and graph structures within a single prompt, because it could reduce computational costs and simplify user interaction with LLMs (as one does not need multiple prompting interactions to build a given topology). The challenge lies in balancing the richness of the prompt with the need to maintain clarity and avoid overwhelming the model. Here, one could harness a recent line of works related to encoding graph structures within a prompt, such as GPT4Graph~\cite{guo2023gpt4graph}, GraphText~\cite{zhao2023graphtext}, GraphGPT~\cite{tang2023graphgpt}, LLMs-as-Predictors~\cite{chen2023exploring}, and others~\cite{chai2023graphllm,
fatemi2023talk,
hu2023beyond,
huang2023can,
liu2023evaluating,
qian2023can,
wang2023language,
wang2023graph,
ye2023natural,
zhang2023graph,
zhang2023making,
zhang2023llm4dyg}.

\textbf{Investigating New Scheduling Approaches}
Most current structure-enhanced prompting schemes rely on standard scheduling algorithms like Breadth-First Search (BFS), Depth-First Search (DFS), or are manually designed. Exploring new scheduling techniques could lead to more efficient and effective processing of prompts. This could include adaptive scheduling algorithms that adjust their approach based on the nature of the task or the responses of the LLM, potentially enhancing the model's performance in complex reasoning tasks.

\textbf{Investigating Novel Graph Classes}
An interesting idea is to explore \emph{graph classes} for more effective and more efficient reasoning topologies. While current approaches have harnessed, among others, general directed graphs, utilizing specialized ones could lead to better reasoning routines. One could harness, for example, structures behind low-diameter networks~\cite{besta2018slim, besta2014slim, iff2023sparse, dally07, dally08, lakhotia2022polarfly, lakhotia2023network, lakhotia2023polarstar} for potential gains in theoretical properties of LLM reasoning, such as lower latency.

\textbf{Integration with Graph Algorithms and Paradigms}
Integrating graph-related algorithms and paradigms could offer more powerful representations and schedules in LLM prompting. This integration might involve the use of advanced graph algorithms to optimize the structure and flow of the prompting process~\cite{besta2020highc, besta2020substream}, potentially leading to more accurate and efficient outcomes. Research could explore how such different graph paradigms~\cite{besta2017push} can be adapted to the unique requirements of LLM prompting.

\textbf{Diversifying Modalities in Prompting}
Different modalities in prompting, such as visual, auditory, or kinesthetic, are currently underexplored. Research in this area could involve developing multi-modal prompting systems that can understand and respond to inputs in various forms while harnessing the advantages of graphs or trees of thoughts. This diversification could lead to more interactive and inclusive LLM systems that cater to a wider range of users and use cases.

\textbf{Enhancing Retrieval in Prompting}
\ifcnf
Retrieval in prompting is another area that has received certain attention, with various recent schemes~\cite{arora2023gar, asai2023self, chen2022murag, gao2023retrieval, jiang2023active, kim2023tree, wang2023knowledgpt, weston2023system, zeitforward}. Improving retrieval mechanisms could enhance the LLM's ability to access and utilize relevant information more efficiently. This might involve developing more sophisticated algorithms for data retrieval or integrating external databases and knowledge bases~\cite{gdb_survey_paper_Angles, angles2018introduction, besta2023thegdi, davoudian2018survey} to expand the scope and depth of the LLM's responses.
\else
Retrieval in prompting is another area that has received certain attention, with various recent schemes~\cite{arora2023gar, asai2023self, chen2022murag, gao2023retrieval, jiang2023active, kim2023tree, wang2023knowledgpt, weston2023system, zeitforward}. Improving retrieval mechanisms could enhance the LLM's ability to access and utilize relevant information more efficiently. This might involve developing more sophisticated algorithms for data retrieval or integrating external databases and knowledge bases~\cite{gdb_survey_paper_Angles, angles2018introduction, besta2023gdi, besta2023thegdi, davoudian2018survey} to expand the scope and depth of the LLM's responses.
\fi

\textbf{Parallel Design in Prompting}
The aspect of parallel design in runtime prompting is an area that remains mostly unaddressed, and only a few schemes such as Skeleton-of-Thought address this challenge. Building upon these attempts could significantly enhance the speed and efficiency of LLMs. Research could focus on developing models that can simultaneously process multiple components of a prompt or handle various tasks in parallel, or the appropriate mapping to massively parallel architectures~\cite{besta2018slim, besta2019graph, gianinazzi2022spatial, iff2023hexamesh}, thereby reducing latency and improving user experience.
One could also investigate effective integration of prompting with distributed-memory infrastructure and paradigms, such as remote direct memory access (RDMA)~\cite{besta2015active, di2022building, di2019network, gerstenberger2013enabling} or serverless processing~\cite{copik2021sebs}.

\textbf{Integrating Structure-Enhanced Prompting with Graph Neural Networks}
A potential area of exploration is the integration of structure-enhanced prompting with Graph Neural Networks (GNNs)~\cite{besta2024parallel, besta2023high, chami2020machine, kipf2016semi, wu2020comprehensive, zhang2020deep, zhou2020graph} and other mechanisms for Graph Machine Learning~\cite{besta2023hot, bronstein2017geometric, hamilton2017representation}. GNNs, known for their proficiency in handling relational data and capturing dependencies in graph structures, could augment the capabilities of LLMs in processing complex, structured prompts. By embedding prompting structures into graph-based representations, GNNs can provide a more nuanced and context-aware interpretation of the prompts, potentially leading to richer and more accurate responses. Furthermore, leveraging graph-related embeddings~\cite{besta2023hot, gianinazzi2021learning} can enhance the LLMs' ability to capture the subtleties in the relationships and hierarchies present in the prompts. Finally, harnessing heterogeneous GNNs~\cite{ma2023single, sun2012mining, zhang2019heterogeneous} for integration with the concept of different semantic roles of vertices could be an interesting direction.

\textbf{Integrating Structure-Enhanced Prompting with Complex System Architectures}
An essential and emerging area of research is the integration of prompting capabilities into the environment of complex existing system architectures, such as graph representation learning systems~\cite{fey2019fast, li2020pytorch, wang2019deep}, graph databases~\cite{besta2021practice, besta2023demystifying, besta2022neural}, or relational databases. This integration aims to facilitate direct and nuanced interactions with complex data structures stored in these systems. By embedding LLMs into these environments, the prompting process can leverage the inherent organizational and relational capabilities of these databases. This approach would allow LLMs to access, interpret, and manipulate large and intricate datasets more efficiently and accurately. For instance, integrating with graph databases could enable LLMs to naturally understand and utilize the connections and relationships within data, while relational databases could provide a structured and queryable data format that complements the LLMs' linguistic capabilities.

\textbf{Hardware Acceleration} 
Understanding energy and performance bottlenecks and mitigating them with specialized techniques such as processing-in-memory~\cite{ahn2015pim, besta2021sisa, ghose2019processing_pim, mutlu2022modern, mutlu2019, seshadri2017ambit}, FPGAs~\cite{besta2019graph, de2020transformations, mittal2020survey}, or even quantum devices~\cite{bertels2021quantum} will likely be increasingly important. Such advances can also enable much more scalable models and model execution under stringent conditions.

\ifcnf
\vspace{-1em}
\fi
\section{Related Work}

We also broadly discuss related analyses, taxonomies, and surveys.

\ifcnf
\vspace{-0.5em}
\fi
\subsection{General Prompt Engineering}

There exist several detailed overviews of general prompt engineering. The main difference is that {we provide the first taxonomy and analysis of \emph{structure-enhanced} prompting methods, in which we focus on the \emph{topology of the LLM reasoning}.
Wang et al.~\cite{wang2023interactive} provide an overview of interactive NLP, in which they outline interactions of LLMs and humans, knowledge bases, models/tools, and environments.
Gu et al.~\cite{gu2023systematic} provide a systematic survey of prompt engineering on vision-language foundation models.
Liu et al.~\cite{liu2023pre} describe in great detail general prompting and describe this area through the ``pre-train, prompt, and predict'' paradigm of building a prompting scheme.
Qiao et al.~\cite{qiao2022reasoning} overview prompting schemes related to reasoning.
Chen et al.~\cite{chen2023unleashing} review the potential of prompting with LLMs.
Finally, Zhang et al.~\cite{zhang2023igniting} and Chu et al.~\cite{chu2023survey} review CoT and the associated prompting schemes.

\ifcnf
\vspace{-0.5em}
\fi
\subsection{Graph-Related Generative AI}

There have also been numerous works on graphs and LLMs. The key difference is that these works focus mostly on \emph{using graph structures as input data} (at pre-training, fine-tuning, or prompting stage).
We instead focus on graphs (and other structures) as \emph{mechanisms enabling structured LLM reasoning}~\cite{huang2022towards}.
%
%
%

\textbf{General overviews of using graphs together with LLMs} have been outlines by Li et al.~\cite{li2023survey} and Zhang et al.~\cite{zhang2023large}.
%
%
Pan et al.~\cite{pan2023integrating} very briefly mention some of the methods for integrating graphs with LLMs.
Zong et al.~briefly mention graph-related fusion in their work on self-supervised multi-modal learning~\cite{zong2023self}.
Yang et al.~\cite{yang2023data} analyze data-centric graph learning.

Next, various works propose to \textbf{enhance general generative models with knowledge graphs (KGs)}. The focus of these works is usually to use KGs in order to enhance the LLM answers, for example by grounding knowledge in general models to reduce effects such as hallucinations~\cite{hu2023survey, pan2023unifying, wei2021knowledge, yang2021survey, yang2023chatgpt}. Example schemes include Knowledge Graph Prompting (KGP)~\cite{wang2023knowledge}, Graph Neural Prompting (GNP)~\cite{tian2023graph}, Think-on-Graph (ToG)~\cite{sun2023think}, Knowledge Solver (KSL)~\cite{feng2023knowledge}, KnowledGPT~\cite{wang2023knowledgpt}, and others~\cite{brate2022improving, luo2023reasoning}.
Zhu et al.~\cite{zhu2023llms} discuss how LLMs can be used for enhancing KG construction and tasks.
Wen et al.~\cite{wen2023mindmap} present MindMap, a framework to perform reasoning on KG data. Pertinent triples from a KG are retrieved and the LLM is prompted to answer a question based on these triples and show the reasoning process by generating a ``mind map'' in the form of a textual reasoning tree.
%
%
%
%
%
%
%
%
%
%
%
%
%
 
Several works discuss \textbf{graph foundation models}~\cite{galkin2023towards, liu2023towards}. These works discuss how to -- in analogy to LLMs -- devise general models pre-trained on massive amounts of graph data, that could then be fine-tuned for more specific applications, and used together with prompting for answering various graph-related tasks.

Recents schemes illustrate how to run \textbf{analytics on graphs using prompting}~\cite{jin2023large}; examples include GPT4Graph~\cite{guo2023gpt4graph},
GraphText~\cite{zhao2023graphtext}, GraphGPT~\cite{tang2023graphgpt}, LLMs-as-Predictors~\cite{chen2023exploring}, and others~\cite{chai2023graphllm,
fatemi2023talk,
hu2023beyond,
huang2023can,
liu2023evaluating,
qian2023can,
wang2023language,
wang2023graph,
ye2023natural,
zhang2023graph,
zhang2023making,
zhang2023llm4dyg}.
%
%
%
%
%
%

Finally, several works~\cite{sun2023graph, wu2023survey} describe techniques for \textbf{graph prompt learning}, which is a class of approaches for enhanced prompting with graph pre-trained models. This class of schemes is orthogonal to our work, because it focuses on prompting for graph-oriented tasks, and it does not target the LLM structure of reasoning.
%
%



%

\ifcnf
\vspace{-1em}
\fi
\section{Conclusion}

In conclusion, the rise of Large Language Models (LLMs) has revolutionized
machine learning, extending their applications beyond Natural Language
Processing (NLP) into diverse fields like medicine, logical reasoning, and
planning. Prompt engineering has emerged as a crucial area, democratizing access
to LLMs and offering a cost-effective alternative to fine-tuning and
pre-training. However, challenges arise in optimizing LLM queries for complex
tasks due to the inherent limitations of generative Transformer models.

This paper addresses these challenges by introducing a blueprint and an accompanying 
taxonomy of prompting schemes, focusing on the underlying structure of reasoning. 
We propose to model a general prompting scheme as a graph topology, where different
classes of graphs, such as $k$--ary trees or directed graphs, can be used to 
reflect the structure and gain insights into different prompting schemes.
The taxonomy is then used to survey and analyze existing designs,
dissecting them into fundamental aspects such as the representation of the
reasoning topology, the derivation of the topology, or the encoding of the reasoning
schedule. This taxonomy forms a blueprint that can be used to facilitate designing
more effective prompting schemes.

We also conduct an analysis of structure-enhanced prompting methods in terms of their
accuracy and quality of outcomes, latency, and cost-effectiveness. Our investigation results in different insights into the tradeoffs between prompting schemes, which facilitate selecting the best method for a given budget or workload target.

Furthermore, we investigate the preliminary works into foundations of structured-enhancing prompting, parallel and scalable designs, and productivity and programmability. We also provide valuable insights into open challenges and potential research directions, navigating the path for future research avenues into more advanced prompting.

\ifnonb
\ifcnf
\vspace{-1em}
\fi
\section*{Acknowledgements}

{\small We thank Hussein Harake, Colin McMurtrie, Mark Klein, Angelo Mangili, and the whole CSCS team granting access to the Ault and Daint machines, and for their excellent technical support. 
We thank Timo Schneider for help with infrastructure at SPCL.
This project received funding from the European Research Council (Project PSAP, No.~101002047), and the European High-Performance Computing Joint Undertaking (JU) under grant agreement No.~955513 (MAELSTROM). This project was supported by the ETH Future Computing Laboratory (EFCL), financed by a donation from Huawei Technologies. This project received funding from the European Union’s HE research and innovation programme under the grant agreement No.~101070141 (Project GLACIATION).}
We gratefully acknowledge Polish high-performance computing infrastructure PLGrid (HPC Center: ACK Cyfronet AGH) for providing computer facilities and support within computational grants no.~PLG/2024/017103 and PLG/2025/018259.
\fi

\bibliographystyle{abbrv}
\bibliography{references.complete}

\begin{thebibliography}{100}

\bibitem{ahn2015pim}
J.~Ahn, S.~Yoo, O.~Mutlu, and K.~Choi.
\newblock {PIM-Enabled Instructions: A Low-Overhead, Locality-Aware
  Processing-In-Memory Architecture}.
\newblock In {\em Proceedings of the 42nd Annual International Symposium on
  Computer Architecture}, ISCA '15, pages 336--348, Portland, OR, USA, June
  2015. Association for Computing Machinery.

\bibitem{amini2019mathqa}
A.~Amini, S.~Gabriel, S.~Lin, R.~Koncel-Kedziorski, Y.~Choi, and H.~Hajishirzi.
\newblock {MathQA: Towards Interpretable Math Word Problem Solving with
  Operation-Based Formalisms}.
\newblock In J.~Burstein, C.~Doran, and T.~Solorio, editors, {\em Proceedings
  of the 2019 Conference of the North American Chapter of the Association for
  Computational Linguistics: Human Language Technologies, Volume 1 (Long and
  Short Papers)}, NAACL '19, pages 2357--2367, Minneapolis, MN, USA, June 2019.
  Association for Computational Linguistics.

\bibitem{angeli-etal-2015-leveraging}
G.~Angeli, M.~J. Johnson~Premkumar, and C.~D. Manning.
\newblock {Leveraging Linguistic Structure for Open Domain Information
  Extraction}.
\newblock In C.~Zong and M.~Strube, editors, {\em Proceedings of the 53rd
  Annual Meeting of the Association for Computational Linguistics and the 7th
  International Joint Conference on Natural Language Processing (Volume 1: Long
  Papers)}, ACL-IJCNLP '15, pages 344--354, Beijing, China, July 2015.
  Association for Computational Linguistics.

\bibitem{gdb_survey_paper_Angles}
R.~Angles and C.~Gutierrez.
\newblock {Survey of Graph Database Models}.
\newblock {\em ACM Comput. Surv.}, 40(1):1:1--1:39, Feb. 2008.

\bibitem{angles2018introduction}
R.~Angles and C.~Gutierrez.
\newblock {An Introduction to Graph Data Management}.
\newblock In G.~H.~L. Fletcher, J.~Hidders, and J.~L. Larriba{-}Pey, editors,
  {\em Graph Data Management, Fundamental Issues and Recent Developments},
  Data-Centric Systems and Applications (DCSA), pages 1--32. Springer, 2018.

\bibitem{arora2023gar}
D.~Arora, A.~Kini, S.~R. Chowdhury, N.~Natarajan, G.~Sinha, and A.~Sharma.
\newblock {GAR-meets-RAG Paradigm for Zero-Shot Information Retrieval}, Oct.
  2023.
\newblock arXiv:2310.20158.

\bibitem{asai2023self}
A.~Asai, Z.~Wu, Y.~Wang, A.~Sil, and H.~Hajishirzi.
\newblock {Self-RAG: Learning to Retrieve, Generate, and Critique Through
  Self-Reflection}.
\newblock In {\em Proceedings of the Twelfth International Conference on
  Learning Representations}, ICLR '24, Vienna, Austria, May 2024. OpenReview.

\bibitem{austin2021program}
J.~Austin, A.~Odena, M.~Nye, M.~Bosma, H.~Michalewski, D.~Dohan, E.~Jiang,
  C.~Cai, M.~Terry, Q.~Le, and C.~Sutton.
\newblock {Program Synthesis with Large Language Models}, Aug. 2021.
\newblock arXiv:2108.07732.

\bibitem{bertels2021quantum}
K.~Bertels, A.~Sarkar, A.~Krol, R.~Budhrani, J.~Samadi, E.~Geoffroy, J.~Matos,
  R.~Abreu, G.~Gielen, and I.~Ashraf.
\newblock {Quantum Accelerator Stack: A Research Roadmap}, May 2021.
\newblock arXiv:2102.02035.

\bibitem{besta2023graph}
M.~Besta, N.~Blach, A.~Kubicek, R.~Gerstenberger, L.~Gianinazzi, J.~Gajda,
  T.~Lehmann, M.~Podstawski, H.~Niewiadomski, P.~Nyczyk, and T.~Hoefler.
\newblock {Graph of Thoughts: Solving Elaborate Problems with Large Language
  Models}.
\newblock {\em Proceedings of the AAAI Conference on Artificial Intelligence},
  38(16):17682--17690, Mar. 2024.

\bibitem{besta2020highc}
M.~Besta, A.~Carigiet, K.~Janda, Z.~Vonarburg-Shmaria, L.~Gianinazzi, and
  T.~Hoefler.
\newblock {High-Performance Parallel Graph Coloring with Strong Guarantees on
  Work, Depth, and Quality}.
\newblock In {\em Proceedings of the International Conference for High
  Performance Computing, Networking, Storage and Analysis}, SC '20, pages
  99:1--99:17, Atlanta, GA, USA, Nov. 2020. IEEE Press.

\bibitem{besta2023hot}
M.~Besta, A.~C. Catarino, L.~Gianinazzi, N.~Blach, P.~Nyczyk, H.~Niewiadomski,
  and T.~Hoefler.
\newblock {HOT: Higher-Order Dynamic Graph Representation Learning with
  Efficient Transformers}.
\newblock In S.~Villar and B.~Chamberlain, editors, {\em Proceedings of the
  Second Learning on Graphs Conference (LOG '23)}, volume 231 of {\em
  Proceedings of Machine Learning Research}, pages 15:1--15:20, Virtual Event,
  Nov. 2023. PMLR.

\bibitem{besta2020substream}
M.~Besta, M.~Fischer, T.~Ben-Nun, D.~Stanojevic, J.~De~Fine~Licht, and
  T.~Hoefler.
\newblock {Substream-Centric Maximum Matchings on FPGA}.
\newblock {\em ACM Trans. Reconfigurable Technol. Syst.}, 13(2):8:1--8:33, Apr.
  2020.

\bibitem{besta2021practice}
M.~Besta, M.~Fischer, V.~Kalavri, M.~Kapralov, and T.~Hoefler.
\newblock {Practice of Streaming Processing of Dynamic Graphs: Concepts,
  Models, and Systems}.
\newblock {\em IEEE Transactions on Parallel and Distributed Systems},
  34(6):1860--1876, June 2023.

\bibitem{besta2023gdi}
M.~Besta, R.~Gerstenberger, N.~Blach, M.~Fischer, and T.~Hoefler.
\newblock {GDI: A Graph Database Interface Standard}.
\newblock {https://github.com/spcl/GDI-RMA}, Nov. 2023.
\newblock (accessed Jan. 15, 2024).

\bibitem{besta2023thegdi}
M.~Besta, R.~Gerstenberger, M.~Fischer, M.~Podstawski, N.~Blach, B.~Egeli,
  G.~Mitenkov, W.~Chlapek, M.~Michalewicz, H.~Niewiadomski, J.~M\"{u}ller, and
  T.~Hoefler.
\newblock {The Graph Database Interface: Scaling Online Transactional and
  Analytical Graph Workloads to Hundreds of Thousands of Cores}.
\newblock In {\em Proceedings of the International Conference for High
  Performance Computing, Networking, Storage and Analysis}, SC '23, pages
  22:1--22:18, Denver, CO, USA, Nov. 2023. Association for Computing Machinery.

\bibitem{besta2023demystifying}
M.~Besta, R.~Gerstenberger, E.~Peter, M.~Fischer, M.~Podstawski, C.~Barthels,
  G.~Alonso, and T.~Hoefler.
\newblock {Demystifying Graph Databases: Analysis and Taxonomy of Data
  Organization, System Designs, and Graph Queries}.
\newblock {\em ACM Comput. Surv.}, 56(2):31:1--31:40, Sept. 2023.

\bibitem{besta2022motif}
M.~Besta, R.~Grob, C.~Miglioli, N.~Bernold, G.~Kwa\'{s}niewski, G.~Gjini,
  R.~Kanakagiri, S.~Ashkboos, L.~Gianinazzi, N.~Dryden, and T.~Hoefler.
\newblock {Motif Prediction with Graph Neural Networks}.
\newblock In {\em Proceedings of the 28th ACM SIGKDD Conference on Knowledge
  Discovery and Data Mining}, KDD '22, pages 35--45, Washington DC, USA, Aug.
  2022. Association for Computing Machinery.

\bibitem{besta2018slim}
M.~Besta, S.~M. Hassan, S.~Yalamanchili, R.~Ausavarungnirun, O.~Mutlu, and
  T.~Hoefler.
\newblock {Slim NoC: A Low-Diameter On-Chip Network Topology for High Energy
  Efficiency and Scalability}.
\newblock {\em SIGPLAN Not.}, 53(2):43--55, Mar. 2018.

\bibitem{besta2014slim}
M.~Besta and T.~Hoefler.
\newblock {Slim Fly: A Cost Effective Low-Diameter Network Topology}.
\newblock In {\em Proceedings of the International Conference for High
  Performance Computing, Networking, Storage and Analysis}, SC '14, pages
  348--359, New Orleans, LA, USA, Nov. 2014. IEEE Press.

\bibitem{besta2015active}
M.~Besta and T.~Hoefler.
\newblock {Active Access: A Mechanism for High-Performance Distributed
  Data-Centric Computations}.
\newblock In {\em Proceedings of the 29th ACM on International Conference on
  Supercomputing}, ICS '15, pages 155--164, Newport Beach, CA, USA, June 2015.
  Association for Computing Machinery.

\bibitem{besta2024parallel}
M.~Besta and T.~Hoefler.
\newblock {Parallel and Distributed Graph Neural Networks: An In-Depth
  Concurrency Analysis}.
\newblock {\em IEEE Transactions on Pattern Analysis and Machine Intelligence},
  46(5):2584--2606, May 2024.

\bibitem{besta2022neural}
M.~Besta, P.~Iff, F.~Scheidl, K.~Osawa, N.~Dryden, M.~Podstawski, T.~Chen, and
  T.~Hoefler.
\newblock {Neural Graph Databases}.
\newblock In B.~Rieck and R.~Pascanu, editors, {\em Proceedings of the First
  Learning on Graphs Conference (LOG '22)}, volume 198 of {\em Proceedings of
  Machine Learning Research}, pages 31:1--31:38, Virtual Event, Dec. 2022.
  PMLR.

\bibitem{besta2021sisa}
M.~Besta, R.~Kanakagiri, G.~Kwa\'{s}niewski, R.~Ausavarungnirun,
  J.~Ber\'{a}nek, K.~Kanellopoulos, K.~Janda, Z.~Vonarburg-Shmaria,
  L.~Gianinazzi, I.~Stefan, J.~G. Luna, J.~Golinowski, M.~Copik,
  L.~Kapp-Schwoerer, S.~Di~Girolamo, N.~Blach, M.~Konieczny, O.~Mutlu, and
  T.~Hoefler.
\newblock {SISA: Set-Centric Instruction Set Architecture for Graph Mining on
  Processing-in-Memory Systems}.
\newblock In {\em Proceedings of the 54th Annual IEEE/ACM International
  Symposium on Microarchitecture}, MICRO '21, pages 282--297, Virtual Event,
  Oct. 2021. Association for Computing Machinery.

\bibitem{besta2022probgraph}
M.~Besta, C.~Miglioli, P.~Sylos~Labini, J.~T\v{e}tek, P.~Iff, R.~Kanakagiri,
  S.~Ashkboos, K.~Janda, M.~Podstawski, G.~Kwa\'{s}niewski, N.~Gleinig,
  F.~Vella, O.~Mutlu, and T.~Hoefler.
\newblock {ProbGraph: High-Performance and High-Accuracy Graph Mining with
  Probabilistic Set Representations}.
\newblock In {\em Proceedings of the International Conference on High
  Performance Computing, Networking, Storage and Analysis}, SC '22, Dallas, TX,
  USA, Nov. 2022. IEEE Press.

\bibitem{besta2017push}
M.~Besta, M.~Podstawski, L.~Groner, E.~Solomonik, and T.~Hoefler.
\newblock {To Push or To Pull: On Reducing Communication and Synchronization in
  Graph Computations}.
\newblock In {\em Proceedings of the 26th International Symposium on
  High-Performance Parallel and Distributed Computing}, HPDC '17, pages
  93--104, Washington DC, USA, June 2017. Association for Computing Machinery.

\bibitem{besta2023high}
M.~Besta, P.~Renc, R.~Gerstenberger, P.~Sylos~Labini, A.~Ziogas, T.~Chen,
  L.~Gianinazzi, F.~Scheidl, K.~Szenes, A.~Carigiet, P.~Iff,
  G.~Kwa\'{s}niewski, R.~Kanakagiri, C.~Ge, S.~Jaeger, J.~Was, F.~Vella, and
  T.~Hoefler.
\newblock {High-Performance and Programmable Attentional Graph Neural Networks
  with Global Tensor Formulations}.
\newblock In {\em Proceedings of the International Conference for High
  Performance Computing, Networking, Storage and Analysis}, SC '23, pages
  66:1--66:16, Denver, CO, USA, Nov. 2023. Association for Computing Machinery.

\bibitem{besta2019graph}
M.~Besta, D.~Stanojevic, J.~De~Fine~Licht, T.~Ben-Nun, and T.~Hoefler.
\newblock {Graph Processing on FPGAs: Taxonomy, Survey, Challenges}, Apr. 2019.
\newblock arXiv:1903.06697.

\bibitem{besta2018log}
M.~Besta, D.~Stanojevic, T.~Zivic, J.~Singh, M.~Hoerold, and T.~Hoefler.
\newblock {Log(Graph): A Near-Optimal High-Performance Graph Representation}.
\newblock In {\em Proceedings of the 27th International Conference on Parallel
  Architectures and Compilation Techniques}, PACT '18, pages 7:1--7:13,
  Limassol, Cyprus, Sept. 2018. Association for Computing Machinery.

\bibitem{besta2021graphminesuite}
M.~Besta, Z.~Vonarburg-Shmaria, Y.~Schaffner, L.~Schwarz, G.~Kwa\'{s}niewski,
  L.~Gianinazzi, J.~Beranek, K.~Janda, T.~Holenstein, S.~Leisinger,
  P.~Tatkowski, E.~Ozdemir, A.~Balla, M.~Copik, P.~Lindenberger, M.~Konieczny,
  O.~Mutlu, and T.~Hoefler.
\newblock {GraphMineSuite: Enabling High-Performance and Programmable Graph
  Mining Algorithms with Set Algebra}.
\newblock {\em Proc. VLDB Endow.}, 14(11):1922--1935, July 2021.

\bibitem{bonifati2018data}
A.~Bonifati, G.~Fletcher, H.~Voigt, and N.~Yakovets.
\newblock {Data Models}.
\newblock In {\em Querying Graphs}, Synthesis Lectures on Data Management
  (SDLM), pages 3--14. Springer International Publishing, 2018.

\bibitem{brate2022improving}
R.~Brate, M.-H. Dang, F.~Hoppe, Y.~He, A.~Mero{\~n}o-Pe{\~n}uela, and
  V.~Sadashivaiah.
\newblock {Improving Language Model Predictions via Prompts Enriched with
  Knowledge Graphs}.
\newblock In {\em Proceedings of the Workshop on Deep Learning for Knowledge
  Graphs}, DL4KG '22, Hangzhou, China, Oct. 2022.

\bibitem{bronstein2017geometric}
M.~M. Bronstein, J.~Bruna, Y.~LeCun, A.~Szlam, and P.~Vandergheynst.
\newblock {Geometric Deep Learning: Going Beyond Euclidean Data}.
\newblock {\em IEEE Signal Processing Magazine}, 34(4):18--42, July 2017.

\bibitem{brown2020language}
T.~Brown, B.~Mann, N.~Ryder, M.~Subbiah, J.~D. Kaplan, P.~Dhariwal,
  A.~Neelakantan, P.~Shyam, G.~Sastry, A.~Askell, S.~Agarwal, A.~Herbert-Voss,
  G.~Krueger, T.~Henighan, R.~Child, A.~Ramesh, D.~Ziegler, J.~Wu, C.~Winter,
  C.~Hesse, M.~Chen, E.~Sigler, M.~Litwin, S.~Gray, B.~Chess, J.~Clark,
  C.~Berner, S.~McCandlish, A.~Radford, I.~Sutskever, and D.~Amodei.
\newblock {Language Models are Few-Shot Learners}.
\newblock In H.~Larochelle, M.~Ranzato, R.~Hadsell, M.~Balcan, and H.~Lin,
  editors, {\em Proceedings of the Thirty-fourth Annual Conference on Neural
  Information Processing Systems (NeurIPS '20)}, volume~33 of {\em Advances in
  Neural Information Processing Systems}, pages 1877--1901, Virtual Event, Dec.
  2020. Curran Associates.

\bibitem{cao2023enhancing}
L.~Cao.
\newblock {GraphReason: Enhancing Reasoning Capabilities of Large Language
  Models Through a Graph-Based Verification Approach}.
\newblock In B.~Dalvi~Mishra, G.~Durrett, P.~Jansen, B.~Lipkin,
  D.~Neves~Ribeiro, L.~Wong, X.~Ye, and W.~Zhao, editors, {\em Proceedings of
  the 2nd Workshop on Natural Language Reasoning and Structured Explanations},
  NLRSE '24, pages 1--12, Bangkok, Thailand, Aug. 2024. Association for
  Computational Linguistics.

\bibitem{chai2023graphllm}
Z.~Chai, T.~Zhang, L.~Wu, K.~Han, X.~Hu, X.~Huang, and Y.~Yang.
\newblock {GraphLLM: Boosting Graph Reasoning Ability of Large Language Model},
  Oct. 2023.
\newblock arXiv:2310.05845.

\bibitem{chami2020machine}
I.~Chami, S.~Abu-El-Haija, B.~Perozzi, C.~R{\'e}, and K.~Murphy.
\newblock {Machine Learning on Graphs: A Model and Comprehensive Taxonomy}.
\newblock {\em Journal of Machine Learning Research}, 23(89):1--64, 2022.

\bibitem{chen2023unleashing}
B.~Chen, Z.~Zhang, N.~Langren{\'e}, and S.~Zhu.
\newblock {Unleashing the Potential of Prompt Engineering in Large Language
  Models: A Comprehensive Review}, Sept. 2024.
\newblock arXiv:2310.14735.

\bibitem{chen2021evaluating}
M.~Chen, J.~Tworek, H.~Jun, Q.~Yuan, H.~P. de~Oliveira~Pinto, J.~Kaplan,
  H.~Edwards, Y.~Burda, N.~Joseph, G.~Brockman, A.~Ray, R.~Puri, G.~Krueger,
  M.~Petrov, H.~Khlaaf, G.~Sastry, P.~Mishkin, B.~Chan, S.~Gray, N.~Ryder,
  M.~Pavlov, A.~Power, L.~Kaiser, M.~Bavarian, C.~Winter, P.~Tillet, F.~P.
  Such, D.~Cummings, M.~Plappert, F.~Chantzis, E.~Barnes, A.~Herbert-Voss,
  W.~H. Guss, A.~Nichol, A.~Paino, N.~Tezak, J.~Tang, I.~Babuschkin, S.~Balaji,
  S.~Jain, W.~Saunders, C.~Hesse, A.~N. Carr, J.~Leike, J.~Achiam, V.~Misra,
  E.~Morikawa, A.~Radford, M.~Knight, M.~Brundage, M.~Murati, K.~Mayer,
  P.~Welinder, B.~McGrew, D.~Amodei, S.~McCandlish, I.~Sutskever, and
  W.~Zaremba.
\newblock {Evaluating Large Language Models Trained on Code}, July 2021.
\newblock arXiv:2107.03374.

\bibitem{chen2022murag}
W.~Chen, H.~Hu, X.~Chen, P.~Verga, and W.~Cohen.
\newblock {MuRAG: Multimodal Retrieval-Augmented Generator for Open Question
  Answering over Images and Text}.
\newblock In Y.~Goldberg, Z.~Kozareva, and Y.~Zhang, editors, {\em Proceedings
  of the 2022 Conference on Empirical Methods in Natural Language Processing},
  EMNLP '22, pages 5558--5570, Abu Dhabi, United Arab Emirates, Dec. 2022.
  Association for Computational Linguistics.

\bibitem{chen2022program}
W.~Chen, X.~Ma, X.~Wang, and W.~W. Cohen.
\newblock {Program of Thoughts Prompting: Disentangling Computation from
  Reasoning for Numerical Reasoning Tasks}.
\newblock {\em Transactions on Machine Learning Research}, Nov. 2023.

\bibitem{chen2021finqa}
Z.~Chen, W.~Chen, C.~Smiley, S.~Shah, I.~Borova, D.~Langdon, R.~Moussa,
  M.~Beane, T.-H. Huang, B.~Routledge, and W.~Y. Wang.
\newblock {FinQA: A Dataset of Numerical Reasoning over Financial Data}.
\newblock In M.-F. Moens, X.~Huang, L.~Specia, and S.~W.-t. Yih, editors, {\em
  Proceedings of the 2021 Conference on Empirical Methods in Natural Language
  Processing}, EMNLP '21, pages 3697--3711, Punta Cana, Dominican Republic,
  Nov. 2021. Association for Computational Linguistics.

\bibitem{chen2022convfinqa}
Z.~Chen, S.~Li, C.~Smiley, Z.~Ma, S.~Shah, and W.~Y. Wang.
\newblock {ConvFinQA: Exploring the Chain of Numerical Reasoning in
  Conversational Finance Question Answering}.
\newblock In Y.~Goldberg, Z.~Kozareva, and Y.~Zhang, editors, {\em Proceedings
  of the 2022 Conference on Empirical Methods in Natural Language Processing},
  EMNLP '22, pages 6279--6292, Abu Dhabi, United Arab Emirates, Dec. 2022.
  Association for Computational Linguistics.

\bibitem{chen2023exploring}
Z.~Chen, H.~Mao, H.~Li, W.~Jin, H.~Wen, X.~Wei, S.~Wang, D.~Yin, W.~Fan,
  H.~Liu, and J.~Tang.
\newblock {Exploring the Potential of Large Language Models (LLMs) in Learning
  on Graphs}.
\newblock {\em SIGKDD Explor. Newsl.}, 25(2):42--61, Mar. 2024.

\bibitem{chen2023chatcot}
Z.~Chen, K.~Zhou, B.~Zhang, Z.~Gong, W.~X. Zhao, and J.-R. Wen.
\newblock {ChatCoT: Tool-Augmented Chain-of-Thought Reasoning on Chat-Based
  Large Language Models}.
\newblock In H.~Bouamor, J.~Pino, and K.~Bali, editors, {\em Findings of the
  Association for Computational Linguistics: EMNLP 2023}, pages 14777--14790,
  Singapore, Dec. 2023. Association for Computational Linguistics.

\bibitem{vicuna2023}
W.-L. Chiang, Z.~Li, Z.~Lin, Y.~Sheng, Z.~Wu, H.~Zhang, L.~Zheng, S.~Zhuang,
  Y.~Zhuang, J.~E. Gonzalez, I.~Stoica, and E.~P. Xing.
\newblock {Vicuna: An Open-Source Chatbot Impressing GPT-4 with 90\% ChatGPT
  Quality}.
\newblock https://lmsys.org/blog/2023-03-30-vicuna/, Mar. 2023.
\newblock (accessed Dec. 15, 2023).

\bibitem{chu2023survey}
Z.~Chu, J.~Chen, Q.~Chen, W.~Yu, T.~He, H.~Wang, W.~Peng, M.~Liu, B.~Qin, and
  T.~Liu.
\newblock {Navigate Through Enigmatic Labyrinth A Survey of Chain of Thought
  Reasoning: Advances, Frontiers and Future}.
\newblock In L.-W. Ku, A.~Martins, and V.~Srikumar, editors, {\em Proceedings
  of the 62nd Annual Meeting of the Association for Computational Linguistics
  (Volume 1: Long Papers)}, ACL '24, pages 1173--1203, Bangkok, Thailand, Aug.
  2024. Association for Computational Linguistics.

\bibitem{cobbe2021gsm8k}
K.~Cobbe, V.~Kosaraju, M.~Bavarian, M.~Chen, H.~Jun, L.~Kaiser, M.~Plappert,
  J.~Tworek, J.~Hilton, R.~Nakano, C.~Hesse, and J.~Schulman.
\newblock {Training Verifiers to Solve Math Word Problems}, Nov. 2021.
\newblock arXiv:2110.14168.

\bibitem{copik2021sebs}
M.~Copik, G.~Kwa\'{s}niewski, M.~Besta, M.~Podstawski, and T.~Hoefler.
\newblock {SeBS: A Serverless Benchmark Suite for Function-as-a-Service
  Computing}.
\newblock In {\em Proceedings of the 22nd International Middleware Conference},
  Middleware '21, pages 64--78, Qu\'{e}bec City, Canada, Dec. 2021. Association
  for Computing Machinery.

\bibitem{creswell2022faithful}
A.~Creswell and M.~Shanahan.
\newblock {Faithful Reasoning Using Large Language Models}, Aug. 2022.
\newblock arXiv:2208.14271.

\bibitem{creswell2022selection}
A.~Creswell, M.~Shanahan, and I.~Higgins.
\newblock {Selection-Inference: Exploiting Large Language Models for
  Interpretable Logical Reasoning}.
\newblock In {\em Proceedings of the Eleventh International Conference on
  Learning Representations}, ICLR '23, Kigali, Rwanda, May 2023. OpenReview.

\bibitem{Dalvi2021ExplainingAW}
B.~Dalvi, P.~Jansen, O.~Tafjord, Z.~Xie, H.~Smith, L.~Pipatanangkura, and
  P.~Clark.
\newblock {Explaining Answers with Entailment Trees}.
\newblock In M.-F. Moens, X.~Huang, L.~Specia, and S.~W.-t. Yih, editors, {\em
  Proceedings of the 2021 Conference on Empirical Methods in Natural Language
  Processing}, EMNLP '21, pages 7358--7370, Punta Cana, Dominican Republic,
  Nov. 2021. Association for Computational Linguistics.

\bibitem{davoudian2018survey}
A.~Davoudian, L.~Chen, and M.~Liu.
\newblock {A Survey on NoSQL Stores}.
\newblock {\em ACM Comput. Surv.}, 51(2):40:1--40:43, Apr. 2018.

\bibitem{de2020transformations}
J.~De~Fine~Licht, M.~Besta, S.~Meierhans, and T.~Hoefler.
\newblock {Transformations of High-Level Synthesis Codes for High-Performance
  Computing}.
\newblock {\em IEEE Transactions on Parallel and Distributed Systems},
  32(5):1014--1029, May 2020.

\bibitem{di2022building}
S.~Di~Girolamo, D.~De~Sensi, K.~Taranov, M.~Malesevic, M.~Besta, T.~Schneider,
  S.~Kistler, and T.~Hoefler.
\newblock {Building Blocks for Network-Accelerated Distributed File Systems}.
\newblock In {\em Proceedings of the International Conference on High
  Performance Computing, Networking, Storage and Analysis}, SC '22, pages
  10:1--10:14, Dallas, TX, USA, Nov. 2022. IEEE Press.

\bibitem{di2019network}
S.~Di~Girolamo, K.~Taranov, A.~Kurth, M.~Schaffner, T.~Schneider,
  J.~Ber\'{a}nek, M.~Besta, L.~Benini, D.~Roweth, and T.~Hoefler.
\newblock {Network-Accelerated Non-Contiguous Memory Transfers}.
\newblock In {\em Proceedings of the International Conference for High
  Performance Computing, Networking, Storage and Analysis}, SC '19, pages
  56:1--56:14, Denver, CO, USA, Nov. 2019. Association for Computing Machinery.

\bibitem{ding2023thoughts}
R.~Ding, C.~Zhang, L.~Wang, Y.~Xu, M.~Ma, W.~Zhang, S.~Qin, S.~Rajmohan,
  Q.~Lin, and D.~Zhang.
\newblock {Everything of Thoughts: Defying the Law of Penrose Triangle for
  Thought Generation}.
\newblock In L.-W. Ku, A.~Martins, and V.~Srikumar, editors, {\em Findings of
  the Association for Computational Linguistics: ACL 2024}, pages 1638--1662,
  Bangkok, Thailand, Aug. 2024. Association for Computational Linguistics.

\bibitem{dodge2016evaluating}
J.~Dodge, A.~Gane, X.~Zhang, A.~Bordes, S.~Chopra, A.~Miller, A.~Szlam, and
  J.~Weston.
\newblock {Evaluating Prerequisite Qualities for Learning End-to-End Dialog
  Systems}, Apr. 2016.
\newblock arXiv:1511.06931.

\bibitem{drozdov2022compositional}
A.~Drozdov, N.~Sch{\"a}rli, E.~Aky{\"u}rek, N.~Scales, X.~Song, X.~Chen,
  O.~Bousquet, and D.~Zhou.
\newblock {Compositional Semantic Parsing with Large Language Models}.
\newblock In {\em Proceedings of the Eleventh International Conference on
  Learning Representations}, ICLR '23, Kigali, Rwanda, May 2023. OpenReview.

\bibitem{Dua2022SuccessivePF}
D.~Dua, S.~Gupta, S.~Singh, and M.~Gardner.
\newblock {Successive Prompting for Decomposing Complex Questions}.
\newblock In Y.~Goldberg, Z.~Kozareva, and Y.~Zhang, editors, {\em Proceedings
  of the 2022 Conference on Empirical Methods in Natural Language Processing},
  EMNLP '22, pages 1251--1265, Abu Dhabi, United Arab Emirates, Dec. 2022.
  Association for Computational Linguistics.

\bibitem{dua2019drop}
D.~Dua, Y.~Wang, P.~Dasigi, G.~Stanovsky, S.~Singh, and M.~Gardner.
\newblock {DROP: A Reading Comprehension Benchmark Requiring Discrete Reasoning
  Over Paragraphs}.
\newblock In J.~Burstein, C.~Doran, and T.~Solorio, editors, {\em Proceedings
  of the 2019 Conference of the North American Chapter of the Association for
  Computational Linguistics: Human Language Technologies, Volume 1 (Long and
  Short Papers)}, NAACL '19, pages 2368--2378, Minneapolis, MN, USA, June 2019.
  Association for Computational Linguistics.

\bibitem{fatemi2023talk}
B.~Fatemi, J.~Halcrow, and B.~Perozzi.
\newblock {Talk like a Graph: Encoding Graphs for Large Language Models}.
\newblock In {\em Proceedings of the Workshop on New Frontiers in Graph
  Learning}, GLFrontiers '23, New Orleans, LA, USA, Dec. 2023. OpenReview.

\bibitem{feng2023knowledge}
C.~Feng, X.~Zhang, and Z.~Fei.
\newblock {Knowledge Solver: Teaching LLMs to Search for Domain Knowledge from
  Knowledge Graphs}, Sept. 2023.
\newblock arXiv:2309.03118.

\bibitem{fey2019fast}
M.~Fey and J.~E. Lenssen.
\newblock {Fast Graph Representation Learning with PyTorch Geometric}.
\newblock In {\em Proceedings of the Workshop on Representation Learning on
  Graphs and Manifolds}, RLGM '19, New Orleans, LA, USA, May 2019.

\bibitem{freund1996experiments}
Y.~Freund and R.~E. Schapire.
\newblock {Experiments with a New Boosting Algorithm}.
\newblock In {\em Proceedings of the Thirteenth International Conference on
  International Conference on Machine Learning}, ICML '96, pages 148--156,
  Bari, Italy, July 1996. Morgan Kaufmann.

\bibitem{fu2023lookahead}
Y.~Fu, P.~Bailis, I.~Stoica, and H.~Zhang.
\newblock {Breaking the Sequential Dependency of LLM Inference Using Lookahead
  Decoding}.
\newblock {https://lmsys.org/blog/2023-11-21-lookahead-decoding/}, Nov. 2023.
\newblock (accessed Dec. 23, 2023).

\bibitem{galkin2023towards}
M.~Galkin, X.~Yuan, H.~Mostafa, J.~Tang, and Z.~Zhu.
\newblock {Towards Foundation Models for Knowledge Graph Reasoning}.
\newblock In {\em Proceedings of the Twelfth International Conference on
  Learning Representations}, ICLR '24, Vienna, Austria, May 2024. OpenReview.

\bibitem{gao2022pal}
L.~Gao, A.~Madaan, S.~Zhou, U.~Alon, P.~Liu, Y.~Yang, J.~Callan, and G.~Neubig.
\newblock {PAL: Program-Aided Language Models}.
\newblock In A.~Krause, E.~Brunskill, K.~Cho, B.~Engelhardt, S.~Sabato, and
  J.~Scarlett, editors, {\em Proceedings of the 40th International Conference
  on Machine Learning (ICML '23)}, volume 202 of {\em Proceedings of Machine
  Learning Research}, pages 10764--10799, Honolulu, HI, USA, July 2023. PMLR.

\bibitem{gao2023retrieval}
Y.~Gao, Y.~Xiong, X.~Gao, K.~Jia, J.~Pan, Y.~Bi, Y.~Dai, J.~Sun, and H.~Wang.
\newblock {Retrieval-Augmented Generation for Large Language Models: A Survey},
  Mar. 2024.
\newblock arXiv:2312.10997.

\bibitem{garcia2020dataset}
N.~Garcia, C.~Ye, Z.~Liu, Q.~Hu, M.~Otani, C.~Chu, Y.~Nakashima, and
  T.~Mitamura.
\newblock {A Dataset and Baselines for Visual Question Answering on Art}.
\newblock In A.~Bartoli and A.~Fusiello, editors, {\em Computer Vision -- ECCV
  2020 Workshops}, volume 12536 of {\em Lecture Notes in Computer Science
  (LNCS)}, pages 92--108, Glasgow, UK, Aug. 2020. Springer.

\bibitem{gerstenberger2013enabling}
R.~Gerstenberger, M.~Besta, and T.~Hoefler.
\newblock {Enabling Highly-Scalable Remote Memory Access Programming with MPI-3
  One Sided}.
\newblock In {\em Proceedings of the International Conference on High
  Performance Computing, Networking, Storage and Analysis}, SC '13, pages
  53:1--53:12, Denver, CO, USA, Nov. 2013. Association for Computing Machinery.

\bibitem{geva2021strategyqa}
M.~Geva, D.~Khashabi, E.~Segal, T.~Khot, D.~Roth, and J.~Berant.
\newblock {Did Aristotle Use a Laptop? A Question Answering Benchmark with
  Implicit Reasoning Strategies}.
\newblock {\em Transactions of the Association for Computational Linguistics},
  9:346--361, 2021.

\bibitem{ghose2019processing_pim}
S.~Ghose, A.~Boroumand, J.~S. Kim, J.~G{\'o}mez-Luna, and O.~Mutlu.
\newblock {Processing-In-Memory: A Workload-Driven Perspective}.
\newblock {\em IBM Journal of Research and Development}, 63(6):3:1--3:19, Nov.
  2019.

\bibitem{gianinazzi2022spatial}
L.~Gianinazzi, T.~Ben-Nun, M.~Besta, S.~Ashkboos, Y.~Baumann, P.~Luczynski, and
  T.~Hoefler.
\newblock {The Spatial Computer: A Model for Energy-Efficient Parallel
  Computation}, Jan. 2023.
\newblock arXiv:2205.04934.

\bibitem{gianinazzi2021parallel}
L.~Gianinazzi, M.~Besta, Y.~Schaffner, and T.~Hoefler.
\newblock {Parallel Algorithms for Finding Large Cliques in Sparse Graphs}.
\newblock In {\em Proceedings of the 33rd ACM Symposium on Parallelism in
  Algorithms and Architectures}, SPAA '21, pages 243--253, Virtual Event, July
  2021. Association for Computing Machinery.

\bibitem{gianinazzi2021learning}
L.~Gianinazzi, M.~Fries, N.~Dryden, T.~Ben-Nun, M.~Besta, and T.~Hoefler.
\newblock {Learning Combinatorial Node Labeling Algorithms}, May 2022.
\newblock arXiv:2106.03594.

\bibitem{goyal2017making}
Y.~Goyal, T.~Khot, A.~Agrawal, D.~Summers-Stay, D.~Batra, and D.~Parikh.
\newblock {Making the V in VQA Matter: Elevating the Role of Image
  Understanding in Visual Question Answering}.
\newblock {\em International Journal of Computer Vision}, 127(4):398--414, Apr.
  2019.

\bibitem{gu2023systematic}
J.~Gu, Z.~Han, S.~Chen, A.~Beirami, B.~He, G.~Zhang, R.~Liao, Y.~Qin, V.~Tresp,
  and P.~Torr.
\newblock {A Systematic Survey of Prompt Engineering on Vision-Language
  Foundation Models}, July 2023.
\newblock arXiv:2307.12980.

\bibitem{gu2022vision}
J.~Gu, E.~Stefani, Q.~Wu, J.~Thomason, and X.~Wang.
\newblock {Vision-and-Language Navigation: A Survey of Tasks, Methods, and
  Future Directions}.
\newblock In S.~Muresan, P.~Nakov, and A.~Villavicencio, editors, {\em
  Proceedings of the 60th Annual Meeting of the Association for Computational
  Linguistics (Volume 1: Long Papers)}, ACL '22, pages 7606--7623, Dublin,
  Ireland, May 2022. Association for Computational Linguistics.

\bibitem{guo2023gpt4graph}
J.~Guo, L.~Du, and H.~Liu.
\newblock {GPT4Graph: Can Large Language Models Understand Graph Structured
  Data? An Empirical Evaluation and Benchmarking}, July 2023.
\newblock arXiv:2305.15066.

\bibitem{yang2023data}
Y.~Guo, D.~Bo, C.~Yang, Z.~Lu, Z.~Zhang, J.~Liu, Y.~Peng, and C.~Shi.
\newblock {Data-Centric Graph Learning: A Survey}.
\newblock {\em IEEE Transactions on Big Data}, 11(1):1--20, Feb. 2025.

\bibitem{hahn2023theory}
M.~Hahn and N.~Goyal.
\newblock {A Theory of Emergent In-Context Learning as Implicit Structure
  Induction}, Mar. 2023.
\newblock arXiv:2303.07971.

\bibitem{hamilton2017representation}
W.~L. Hamilton, R.~Ying, and J.~Leskovec.
\newblock {Representation Learning on Graphs: Methods and Applications}.
\newblock {\em Bulletin of the Technical Committee on Data Engineering},
  40(3):52--74, Sept. 2017.

\bibitem{han2022folio}
S.~Han, H.~Schoelkopf, Y.~Zhao, Z.~Qi, M.~Riddell, W.~Zhou, J.~Coady, D.~Peng,
  Y.~Qiao, L.~Benson, L.~Sun, A.~Wardle-Solano, H.~Szab{\'o}, E.~Zubova,
  M.~Burtell, J.~Fan, Y.~Liu, B.~Wong, M.~Sailor, A.~Ni, L.~Nan, J.~Kasai,
  T.~Yu, R.~Zhang, A.~Fabbri, W.~M. Kryscinski, S.~Yavuz, Y.~Liu, X.~V. Lin,
  S.~Joty, Y.~Zhou, C.~Xiong, R.~Ying, A.~Cohan, and D.~Radev.
\newblock {FOLIO: Natural Language Reasoning with First-Order Logic}.
\newblock In Y.~Al-Onaizan, M.~Bansal, and Y.-N. Chen, editors, {\em
  Proceedings of the 2024 Conference on Empirical Methods in Natural Language
  Processing}, EMNLP '24, pages 22017--22031, Miami, FL, USA, Nov. 2024.
  Association for Computational Linguistics.

\bibitem{hartmann2022survey}
M.~Hartmann and D.~Sonntag.
\newblock {A Survey on Improving NLP Models with Human Explanations}.
\newblock In J.~Andreas, K.~Narasimhan, and A.~Nematzadeh, editors, {\em
  Proceedings of the First Workshop on Learning with Natural Language
  Supervision}, LNLS '22, pages 40--47, Dublin, Ireland, May 2022. Association
  for Computational Linguistics.

\bibitem{heyueya2023solving}
J.~He-Yueya, G.~Poesia, R.~E. Wang, and N.~D. Goodman.
\newblock {Solving Math Word Problems by Combining Language Models With
  Symbolic Solvers}, Apr. 2023.
\newblock arXiv:2304.09102.

\bibitem{hendryckstest2021}
D.~Hendrycks, C.~Burns, S.~Basart, A.~Zou, M.~Mazeika, D.~Song, and
  J.~Steinhardt.
\newblock {Measuring Massive Multitask Language Understanding}.
\newblock In {\em Proceedings of the Ninth International Conference on Learning
  Representations}, ICLR '21, Virtual Event, May 2021. OpenReview.

\bibitem{hendrycksmath2021}
D.~Hendrycks, C.~Burns, S.~Kadavath, A.~Arora, S.~Basart, E.~Tang, D.~Song, and
  J.~Steinhardt.
\newblock {Measuring Mathematical Problem Solving with the MATH Dataset}.
\newblock In J.~Vanschoren and S.~Yeung, editors, {\em Proceedings of the
  Thirty-fifth Conference on Neural Information Processing Systems: Datasets
  and Benchmarks Track}, NeurIPS '21, Virtual Event, Dec. 2021.

\bibitem{ho2020constructing}
X.~Ho, A.-K. Duong~Nguyen, S.~Sugawara, and A.~Aizawa.
\newblock {Constructing A Multi-Hop QA Dataset for Comprehensive Evaluation of
  Reasoning Steps}.
\newblock In D.~Scott, N.~Bel, and C.~Zong, editors, {\em Proceedings of the
  28th International Conference on Computational Linguistics}, COLING '20,
  pages 6609--6625, Barcelona, Spain, Dec. 2020. International Committee on
  Computational Linguistics.

\bibitem{hosseinii2014learning}
M.~J. Hosseini, H.~Hajishirzi, O.~Etzioni, and N.~Kushman.
\newblock {Learning to Solve Arithmetic Word Problems with Verb
  Categorization}.
\newblock In A.~Moschitti, B.~Pang, and W.~Daelemans, editors, {\em Proceedings
  of the 2014 Conference on Empirical Methods in Natural Language Processing},
  EMNLP '14, pages 523--533, Doha, Qatar, Oct. 2014. Association for
  Computational Linguistics.

\bibitem{hu2023chain}
H.~Hu, H.~Lu, H.~Zhang, W.~Lam, and Y.~Zhang.
\newblock {Chain-of-Symbol Prompting Elicits Planning in Large Language
  Models}, Aug. 2024.
\newblock arXiv:2305.10276.

\bibitem{hu2023survey}
L.~Hu, Z.~Liu, Z.~Zhao, L.~Hou, L.~Nie, and J.~Li.
\newblock {A Survey of Knowledge Enhanced Pre-Trained Language Models}.
\newblock {\em IEEE Transactions on Knowledge and Data Engineering},
  36(4):1413--1430, Apr. 2024.

\bibitem{hu2023treeofmixedthought}
P.~Hu, J.~Qi, X.~Li, H.~Li, X.~Wang, B.~Quan, R.~Wang, and Y.~Zhou.
\newblock {Tree-of-Mixed-Thought: Combining Fast and Slow Thinking for
  Multi-Hop Visual Reasoning}, Aug. 2023.
\newblock arXiv:2308.09658.

\bibitem{hu2023beyond}
Y.~Hu, Z.~Zhang, and L.~Zhao.
\newblock {Beyond Text: A Deep Dive into Large Language Models' Ability on
  Understanding Graph Data}.
\newblock In {\em Proceedings of the Workshop on New Frontiers in Graph
  Learning}, GLFrontiers '23, New Orleans, LA, USA, Dec. 2023. OpenReview.

\bibitem{huang2022towards}
J.~Huang and K.~C.-C. Chang.
\newblock {Towards Reasoning in Large Language Models: A Survey}.
\newblock In A.~Rogers, J.~Boyd-Graber, and N.~Okazaki, editors, {\em Findings
  of the Association for Computational Linguistics: ACL 2023}, pages
  1049--1065, Toronto, Canada, July 2023. Association for Computational
  Linguistics.

\bibitem{huang2023can}
J.~Huang, X.~Zhang, Q.~Mei, and J.~Ma.
\newblock {Can LLMs Effectively Leverage Graph Structural Information through
  Prompts, and Why}.
\newblock {\em Transactions on Machine Learning Research}, June 2024.

\bibitem{iff2023hexamesh}
P.~Iff, M.~Besta, M.~Cavalcante, T.~Fischer, L.~Benini, and T.~Hoefler.
\newblock {HexaMesh: Scaling to Hundreds of Chiplets with an Optimized Chiplet
  Arrangement}.
\newblock In {\em Proceedings of the 60th ACM/IEEE Design Automation
  Conference}, DAC '23, pages 1--6, San Francisco, CA, USA, July 2023. IEEE
  Press.

\bibitem{iff2023sparse}
P.~Iff, M.~Besta, M.~Cavalcante, T.~Fischer, L.~Benini, and T.~Hoefler.
\newblock {Sparse Hamming Graph: A Customizable Network-on-Chip Topology}.
\newblock In {\em Proceedings of the 2023 60th ACM/IEEE Design Automation
  Conference}, DAC '23, pages 1--6, San Francisco, CA, USA, July 2023.

\bibitem{jiang2023latent}
H.~Jiang.
\newblock {A Latent Space Theory for Emergent Abilities in Large Language
  Models}, Sept. 2023.
\newblock arXiv:2304.09960.

\bibitem{jiang2023resprompt}
S.~Jiang, Z.~Shakeri, A.~Chan, M.~Sanjabi, H.~Firooz, Y.~Xia, B.~Akyildiz,
  Y.~Sun, J.~Li, Q.~Wang, and A.~Celikyilmaz.
\newblock {RESPROMPT: Residual Connection Prompting Advances Multi-Step
  Reasoning in Large Language Models}.
\newblock In K.~Duh, H.~Gomez, and S.~Bethard, editors, {\em Proceedings of the
  2024 Conference of the North American Chapter of the Association for
  Computational Linguistics: Human Language Technologies (Volume 1: Long
  Papers)}, ACL '24, pages 5784--5809, Mexico City, Mexico, June 2024.
  Association for Computational Linguistics.

\bibitem{jiang2023active}
Z.~Jiang, F.~Xu, L.~Gao, Z.~Sun, Q.~Liu, J.~Dwivedi-Yu, Y.~Yang, J.~Callan, and
  G.~Neubig.
\newblock {Active Retrieval Augmented Generation}.
\newblock In H.~Bouamor, J.~Pino, and K.~Bali, editors, {\em Proceedings of the
  2023 Conference on Empirical Methods in Natural Language Processing}, EMNLP
  '23, pages 7969--7992, Singapore, Dec. 2023. Association for Computational
  Linguistics.

\bibitem{jin2023large}
B.~Jin, G.~Liu, C.~Han, M.~Jiang, H.~Ji, and J.~Han.
\newblock {Large Language Models on Graphs: A Comprehensive Survey}.
\newblock {\em IEEE Transactions on Knowledge and Data Engineering},
  36(12):8622--8642, Dec. 2024.

\bibitem{jung2022maieutic}
J.~Jung, L.~Qin, S.~Welleck, F.~Brahman, C.~Bhagavatula, R.~Le~Bras, and
  Y.~Choi.
\newblock {Maieutic Prompting: Logically Consistent Reasoning with Recursive
  Explanations}.
\newblock In Y.~Goldberg, Z.~Kozareva, and Y.~Zhang, editors, {\em Proceedings
  of the 2022 Conference on Empirical Methods in Natural Language Processing},
  EMNLP '22, pages 1266--1279, Abu Dhabi, United Arab Emirates, Dec. 2022.
  Association for Computational Linguistics.

\bibitem{Keysers2020}
D.~Keysers, N.~Sch\"{a}rli, N.~Scales, H.~Buisman, D.~Furrer, S.~Kashubin,
  N.~Momchev, D.~Sinopalnikov, L.~Stafiniak, T.~Tihon, D.~Tsarkov, X.~Wang,
  M.~van Zee, and O.~Bousquet.
\newblock {Measuring Compositional Generalization: A Comprehensive Method on
  Realistic Data}.
\newblock In {\em Proceedings of the Eighth International Conference on
  Learning Representations}, ICLR '20, Virtual Event, Apr. 2020. OpenReview.

\bibitem{khot2022hey}
T.~Khot, K.~Richardson, D.~Khashabi, and A.~Sabharwal.
\newblock {Hey AI, Can You Solve Complex Tasks by Talking to Agents?}
\newblock In S.~Muresan, P.~Nakov, and A.~Villavicencio, editors, {\em Findings
  of the Association for Computational Linguistics: ACL 2022}, pages
  1808--1823, Dublin, Ireland, May 2022. Association for Computational
  Linguistics.

\bibitem{khot2022decomposed}
T.~Khot, H.~Trivedi, M.~Finlayson, Y.~Fu, K.~Richardson, P.~Clark, and
  A.~Sabharwal.
\newblock {Decomposed Prompting: A Modular Approach for Solving Complex Tasks}.
\newblock In {\em Proceedings of the Eleventh International Conference on
  Learning Representations}, ICLR '23, Kigali, Rwanda, May 2023. OpenReview.

\bibitem{kim2023tree}
G.~Kim, S.~Kim, B.~Jeon, J.~Park, and J.~Kang.
\newblock {Tree of Clarifications: Answering Ambiguous Questions with
  Retrieval-Augmented Large Language Models}.
\newblock In H.~Bouamor, J.~Pino, and K.~Bali, editors, {\em Proceedings of the
  2023 Conference on Empirical Methods in Natural Language Processing}, EMNLP
  '23, pages 996--1009, Singapore, Dec. 2023. Association for Computational
  Linguistics.

\bibitem{dally07}
J.~Kim, W.~J. Dally, and D.~Abts.
\newblock {Flattened Butterfly: A Cost-Efficient Topology for High-Radix
  Networks}.
\newblock In {\em Proceedings of the 34th Annual International Symposium on
  Computer Architecture}, ISCA '07, pages 126--137, San Diego, CA, USA, June
  2007. Association for Computing Machinery.

\bibitem{dally08}
J.~Kim, W.~J. Dally, S.~Scott, and D.~Abts.
\newblock {Technology-Driven, Highly-Scalable Dragonfly Topology}.
\newblock In {\em Proceedings of the 35th Annual International Symposium on
  Computer Architecture}, ISCA '08, pages 77--88, Beijing, China, June 2008.
  IEEE Press.

\bibitem{kim2020cogs}
N.~Kim and T.~Linzen.
\newblock {COGS: A Compositional Generalization Challenge Based on Semantic
  Interpretation}.
\newblock In B.~Webber, T.~Cohn, Y.~He, and Y.~Liu, editors, {\em Proceedings
  of the 2020 Conference on Empirical Methods in Natural Language Processing},
  EMNLP '20, pages 9087--9105, Virtual Event, Nov. 2020. Association for
  Computational Linguistics.

\bibitem{kim2023llm}
S.~Kim, S.~Moon, R.~Tabrizi, N.~Lee, M.~W. Mahoney, K.~Keutzer, and A.~Gholami.
\newblock {An LLM Compiler for Parallel Function Calling}.
\newblock In R.~Salakhutdinov, Z.~Kolter, K.~Heller, A.~Weller, N.~Oliver,
  J.~Scarlett, and F.~Berkenkamp, editors, {\em Proceedings of the 41st
  International Conference on Machine Learning (ICML '24)}, volume 235 of {\em
  Proceedings of Machine Learning Research}, pages 24370--24391, Vienna,
  Austria, July 2024. PMLR.

\bibitem{kipf2016semi}
T.~N. Kipf and M.~Welling.
\newblock {Semi-Supervised Classification with Graph Convolutional Network}.
\newblock In {\em Proceedings of the 5th International Conference on Learning
  Representations}, ICLR '17, Toulon, France, Apr. 2017. OpenReview.

\bibitem{kojima2022large}
T.~Kojima, S.~S. Gu, M.~Reid, Y.~Matsuo, and Y.~Iwasawa.
\newblock {Large Language Models Are Zero-Shot Reasoners}.
\newblock In S.~Koyejo, S.~Mohamed, A.~Agarwal, D.~Belgrave, K.~Cho, and A.~Oh,
  editors, {\em Proceedings of the Thirty-sixth Annual Conference on Neural
  Information Processing Systems (NeurIPS '22)}, volume~35 of {\em Advances in
  Neural Information Processing Systems}, pages 22199--22213. Curran
  Associates, New Orleans, LA, USA, Dec. 2022.

\bibitem{koncel2016mawps}
R.~Koncel-Kedziorski, S.~Roy, A.~Amini, N.~Kushman, and H.~Hajishirzi.
\newblock {MAWPS: A Math Word Problem Repository}.
\newblock In K.~Knight, A.~Nenkova, and O.~Rambow, editors, {\em Proceedings of
  the 2016 Conference of the North American Chapter of the Association for
  Computational Linguistics: Human Language Technologies}, NAACL-HLT '16, pages
  1152--1157, San Diego, CA, USA, June 2016. Association for Computational
  Linguistics.

\bibitem{kumar-etal-2022-autotnli}
D.~Kumar, V.~Gupta, S.~Sharma, and S.~Zhang.
\newblock {Realistic Data Augmentation Framework for Enhancing Tabular
  Reasoning}.
\newblock In Y.~Goldberg, Z.~Kozareva, and Y.~Zhang, editors, {\em Findings of
  the Association for Computational Linguistics: EMNLP 2022}, pages 4411--4429,
  Abu Dhabi, United Arab Emirates, Dec. 2022. Association for Computational
  Linguistics.

\bibitem{lake2018generalization}
B.~Lake and M.~Baroni.
\newblock {Generalization without Systematicity: On the Compositional Skills of
  Sequence-to-Sequence Recurrent Networks}.
\newblock In J.~Dy and A.~Krause, editors, {\em Proceedings of the 35th
  International Conference on International Conference on Machine Learning
  (ICML '18)}, volume~80 of {\em Proceedings of Machine Learning Research},
  pages 2873--2882, Stockholm, Sweden, July 2018. PMLR.

\bibitem{lakhotia2022polarfly}
K.~Lakhotia, M.~Besta, L.~Monroe, K.~Isham, P.~Iff, T.~Hoefler, and F.~Petrini.
\newblock {PolarFly: A Cost-Effective and Flexible Low-Diameter Topology}.
\newblock In {\em Proceedings of the International Conference on High
  Performance Computing, Networking, Storage and Analysis}, SC '22, pages
  12:1--12:15, Dallas, TX, USA, Nov. 2022. IEEE Press.

\bibitem{lakhotia2023network}
K.~Lakhotia, K.~Isham, L.~Monroe, M.~Besta, T.~Hoefler, and F.~Petrini.
\newblock {In-Network Allreduce with Multiple Spanning Trees on PolarFly}.
\newblock In {\em Proceedings of the 35th ACM Symposium on Parallelism in
  Algorithms and Architectures}, SPAA '23, pages 165--176, Orlando, FL, USA,
  June 2023. Association for Computing Machinery.

\bibitem{lakhotia2023polarstar}
K.~Lakhotia, L.~Monroe, K.~Isham, M.~Besta, N.~Blach, T.~Hoefler, and
  F.~Petrini.
\newblock {PolarStar: Expanding the Horizon of Diameter-3 Networks}.
\newblock In {\em Proceedings of the 36th ACM Symposium on Parallelism in
  Algorithms and Architectures}, SPAA '24, page 345–357, Nantes, France, June
  2024. Association for Computing Machinery.

\bibitem{lei2023boosting}
B.~Lei, p.-H. Lin, C.~Liao, and C.~Ding.
\newblock {Boosting Logical Reasoning in Large Language Models Through a New
  Framework: The Graph of Thought}, Aug. 2023.
\newblock arXiv:2308.08614.

\bibitem{breiman1984classification}
{Leo Breiman, Jerome Friedman, Charles J. Stone, R.A. Olshen}.
\newblock {\em {Classification and Regression Trees}}.
\newblock Chapman and Hall, 1984.

\bibitem{li2020pytorch}
S.~Li, Y.~Zhao, R.~Varma, O.~Salpekar, P.~Noordhuis, T.~Li, A.~Paszke,
  J.~Smith, B.~Vaughan, P.~Damania, and S.~Chintala.
\newblock {PyTorch Distributed: Experiences on Accelerating Data Parallel
  Training}.
\newblock {\em Proc. VLDB Endow.}, 13(12):3005--3018, Aug. 2020.

\bibitem{li2023survey}
Y.~Li, Z.~Li, P.~Wang, J.~Li, X.~Sun, H.~Cheng, and J.~X. Yu.
\newblock {A Survey of Graph Meets Large Language Model: Progress and Future
  Directions}.
\newblock In K.~Larson, editor, {\em Proceedings of the Thirty-Third
  International Joint Conference on Artificial Intelligence}, IJCAI '24, pages
  8123--8131, Jeju, South Korea, Aug. 2024. International Joint Conferences on
  Artificial Intelligence Organization.
\newblock Survey Track.

\bibitem{li2023makingstep}
Y.~Li, Z.~Lin, S.~Zhang, Q.~Fu, B.~Chen, J.-G. Lou, and W.~Chen.
\newblock {Making Language Models Better Reasoners with Step-Aware Verifier}.
\newblock In A.~Rogers, J.~Boyd-Graber, and N.~Okazaki, editors, {\em
  Proceedings of the 61st Annual Meeting of the Association for Computational
  Linguistics (Volume 1: Long Papers)}, ACL '23, pages 5315--5333, Toronto,
  Canada, July 2023. Association for Computational Linguistics.

\bibitem{lin2023batchprompt}
J.~Lin, M.~Diesendruck, L.~Du, and R.~Abraham.
\newblock {BatchPrompt: Accomplish More with Less}.
\newblock In {\em Proceedings of the Twelfth International Conference on
  Learning Representations}, ICLR '24, Vienna, Austria, May 2024. OpenReview.

\bibitem{ling2017program}
W.~Ling, D.~Yogatama, C.~Dyer, and P.~Blunsom.
\newblock {Program Induction by Rationale Generation: Learning to Solve and
  Explain Algebraic Word Problems}.
\newblock In R.~Barzilay and M.-Y. Kan, editors, {\em Proceedings of the 55th
  Annual Meeting of the Association for Computational Linguistics (Volume 1:
  Long Papers)}, ACL '17, pages 158--167, Vancouver, Canada, July 2017.
  Association for Computational Linguistics.

\bibitem{ling2023deductive}
Z.~Ling, Y.~Fang, X.~Li, Z.~Huang, M.~Lee, R.~Memisevic, and H.~Su.
\newblock {Deductive Verification of Chain-of-Thought Reasoning}.
\newblock In A.~Oh, T.~Neumann, A.~Globerson, K.~Saenko, M.~Hardt, and
  S.~Levine, editors, {\em Proceedings of the Thirty-seventh Annual Conference
  on Neural Information Processing Systems (NeurIPS '23)}, volume~36 of {\em
  Advances in Neural Information Processing Systems}, pages 36407--36433, New
  Orleans, LA, USA, Dec. 2023. Curran Associates.

\bibitem{liu2023evaluating}
C.~Liu and B.~Wu.
\newblock {Evaluating Large Language Models on Graphs: Performance Insights and
  Comparative Analysis}, Sept. 2023.
\newblock arXiv:2308.11224.

\bibitem{liu2020logiqa}
J.~Liu, L.~Cui, H.~Liu, D.~Huang, Y.~Wang, and Y.~Zhang.
\newblock {LogiQA: A Challenge Dataset for Machine Reading Comprehension with
  Logical Reasoning}.
\newblock In C.~Bessiere, editor, {\em Proceedings of the Twenty-Ninth
  International Joint Conference on Artificial Intelligence}, IJCAI '20, pages
  3622--3628, Virtual Event, Jan. 2021. International Joint Conferences on
  Artificial Intelligence Organization.
\newblock Main Track.

\bibitem{liu2023towards}
J.~Liu, C.~Yang, Z.~Lu, J.~Chen, Y.~Li, M.~Zhang, T.~Bai, Y.~Fang, L.~Sun,
  P.~S. Yu, and C.~Shi.
\newblock {Towards Graph Foundation Models: A Survey and Beyond}, July 2024.
\newblock arXiv:2310.11829.

\bibitem{liu2023pre}
P.~Liu, W.~Yuan, J.~Fu, Z.~Jiang, H.~Hayashi, and G.~Neubig.
\newblock {Pre-Train, Prompt, and Predict: A Systematic Survey of Prompting
  Methods in Natural Language Processing}.
\newblock {\em ACM Comput. Surv.}, 55(9):195:1--195:35, Jan. 2023.

\bibitem{liu2023plan}
T.~Liu, Q.~Guo, Y.~Yang, X.~Hu, Y.~Zhang, X.~Qiu, and Z.~Zhang.
\newblock {Plan, Verify and Switch: Integrated Reasoning with Diverse
  X-of-Thoughts}.
\newblock In H.~Bouamor, J.~Pino, and K.~Bali, editors, {\em Proceedings of the
  2023 Conference on Empirical Methods in Natural Language Processing}, EMNLP
  '23, pages 2807--2822, Singapore, Dec. 2023. Association for Computational
  Linguistics.

\bibitem{liu2023controlllm}
Z.~Liu, Z.~Lai, Z.~Gao, E.~Cui, Z.~Li, X.~Zhu, L.~Lu, Q.~Chen, Y.~Qiao, J.~Dai,
  and W.~Wang.
\newblock {ControlLLM: Augment Language Models with Tools by Searching
  on Graphs}.
\newblock In A.~Leonardis, E.~Ricci, S.~Roth, O.~Russakovsky, T.~Sattler, and
  G.~Varol, editors, {\em Proceedings of the 18th Conference on Computer Vision
  (ECCV '24)}, volume 15070 of {\em Lecture Notes in Computer Science (LNCS)},
  pages 89--105, Milan, Italy, Oct. 2024. Springer Nature Switzerland.

\bibitem{long2023large}
J.~Long.
\newblock {Large Language Model Guided Tree-of-Thought}, May 2023.
\newblock arXiv:2305.08291.

\bibitem{long2016simpler}
R.~Long, P.~Pasupat, and P.~Liang.
\newblock {Simpler Context-Dependent Logical Forms via Model Projections}.
\newblock In K.~Erk and N.~A. Smith, editors, {\em Proceedings of the 54th
  Annual Meeting of the Association for Computational Linguistics (Volume 1:
  Long Papers)}, ACL '16, pages 1456--1465, Berlin, Germany, Aug. 2016.
  Association for Computational Linguistics.

\bibitem{lu2022learn}
P.~Lu, S.~Mishra, T.~Xia, L.~Qiu, K.-W. Chang, S.-C. Zhu, O.~Tafjord, P.~Clark,
  and A.~Kalyan.
\newblock {Learn to Explain: Multimodal Reasoning via Thought Chains for
  Science Question Answering}.
\newblock In S.~Koyejo, S.~Mohamed, A.~Agarwal, D.~Belgrave, K.~Cho, and A.~Oh,
  editors, {\em Proceedings of the Thirty-sixth Annual Conference on Neural
  Information Processing Systems (NeurIPS '22)}, volume~35 of {\em Advances in
  Neural Information Processing Systems}, pages 2507--2521, New Orleans, LA,
  USA, Dec. 2022. Curran Associates.

\bibitem{lu2023chameleon}
P.~Lu, B.~Peng, H.~Cheng, M.~Galley, K.-W. Chang, Y.~N. Wu, S.-C. Zhu, and
  J.~Gao.
\newblock {Chameleon: Plug-and-Play Compositional Reasoning with Large Language
  Models}.
\newblock In A.~Oh, T.~Neumann, A.~Globerson, K.~Saenko, M.~Hardt, and
  S.~Levine, editors, {\em Proceedings of the Thirty-seventh Annual Conference
  on Neural Information Processing Systems (NeurIPS '23)}, volume~36 of {\em
  Advances in Neural Information Processing Systems}, pages 43447--43478, New
  Orleans, LA, USA, Dec. 2023. Curran Associates.

\bibitem{lu2023dynamic}
P.~Lu, L.~Qiu, K.-W. Chang, Y.~N. Wu, S.-C. Zhu, T.~Rajpurohit, P.~Clark, and
  A.~Kalyan.
\newblock {Dynamic Prompt Learning via Policy Gradient for Semi-Structured
  Mathematical Reasoning}.
\newblock In {\em Proceedings of the Eleventh International Conference on
  Learning Representations}, ICLR '23, Kigali, Rwanda, May 2023. OpenReview.

\bibitem{luo2023reasoning}
L.~Luo, Y.-F. Li, G.~Haffari, and S.~Pan.
\newblock {Reasoning on Graphs: Faithful and Interpretable Large Language Model
  Reasoning}.
\newblock In {\em Proceedings of the Twelfth International Conference on
  Learning Representations}, ICLR '24, Vienna, Austria, May 2024. OpenReview.

\bibitem{ma2023single}
A.~Ma, X.~Wang, J.~Li, C.~Wang, T.~Xiao, Y.~Liu, H.~Cheng, J.~Wang, Y.~Li,
  Y.~Chang, J.~Li, D.~Wang, Y.~Jiang, L.~Su, G.~Xin, S.~Gu, Z.~Li, B.~Liu,
  D.~Xu, and Q.~Ma.
\newblock {Single-Cell Biological Network Inference Using a Heterogeneous Graph
  Transformer}.
\newblock {\em Nature Communications}, 14(1):964:1--964:18, 2023.

\bibitem{madaan2023self}
A.~Madaan, N.~Tandon, P.~Gupta, S.~Hallinan, L.~Gao, S.~Wiegreffe, U.~Alon,
  N.~Dziri, S.~Prabhumoye, Y.~Yang, S.~Gupta, B.~P. Majumder, K.~Hermann,
  S.~Welleck, A.~Yazdanbakhsh, and P.~Clark.
\newblock {Self-Refine: Iterative Refinement with Self-Feedback}.
\newblock In A.~Oh, T.~Neumann, A.~Globerson, K.~Saenko, M.~Hardt, and
  S.~Levine, editors, {\em Proceedings of the Thirty-seventh Annual Conference
  on Neural Information Processing Systems (NeurIPS '23)}, volume~36 of {\em
  Advances in Neural Information Processing Systems}, pages 46534--46594, New
  Orleans, LA, USA, Dec. 2023. Curran Associates.

\bibitem{madaan2022text}
A.~Madaan and A.~Yazdanbakhsh.
\newblock {Text and Patterns: For Effective Chain of Thought, It Takes Two to
  Tango}, Oct. 2022.
\newblock arXiv:2209.07686.

\bibitem{manning-etal-2014-stanford}
C.~Manning, M.~Surdeanu, J.~Bauer, J.~Finkel, S.~Bethard, and D.~McClosky.
\newblock {The Stanford CoreNLP Natural Language Processing Toolkit}.
\newblock In K.~Bontcheva and J.~Zhu, editors, {\em Proceedings of 52nd Annual
  Meeting of the Association for Computational Linguistics: System
  Demonstrations}, ACL '14, pages 55--60, Baltimore, MD, USA, June 2014.
  Association for Computational Linguistics.

\bibitem{mirzaee2022transfer}
R.~Mirzaee and P.~Kordjamshidi.
\newblock {Transfer Learning with Synthetic Corpora for Spatial Role Labeling
  and Reasoning}.
\newblock In Y.~Goldberg, Z.~Kozareva, and Y.~Zhang, editors, {\em Proceedings
  of the 2022 Conference on Empirical Methods in Natural Language Processing},
  EMNLP '22, pages 6148--6165, Abu Dhabi, United Arab Emirates, Dec. 2022.
  Association for Computational Linguistics.

\bibitem{mittal2020survey}
S.~Mittal.
\newblock {A Survey of FPGA-based Accelerators for Convolutional Neural
  Networks}.
\newblock {\em Neural Computing and Applications}, 32(4):1109--1139, Feb. 2020.

\bibitem{mo2023tree}
S.~Mo and M.~Xin.
\newblock {Tree of Uncertain Thoughts Reasoning for Large Language Models},
  Sept. 2023.
\newblock arXiv:2309.07694.

\bibitem{mutlu2022modern}
O.~Mutlu, S.~Ghose, J.~G{\'o}mez-Luna, and R.~Ausavarungnirun.
\newblock {A Modern Primer on Processing in Memory}.
\newblock In M.~M.~S. Aly and A.~Chattopadhyay, editors, {\em Emerging
  Computing: From Devices to Systems - Looking Beyond Moore and Von Neumann},
  Computer Architecture and Design Methodologies (CADM), pages 171--243.
  Springer Nature Singapore, 2023.

\bibitem{mutlu2019}
O.~Mutlu, S.~Ghose, J.~Gómez-Luna, and R.~Ausavarungnirun.
\newblock {Processing Data Where It Makes Sense: Enabling In-Memory
  Computation}.
\newblock {\em Microprocessors and Microsystems}, 67:28--41, June 2019.

\bibitem{nguyen2024direct}
M.-V. Nguyen, L.~Luo, F.~Shiri, D.~Phung, Y.-F. Li, T.-T. Vu, and G.~Haffari.
\newblock {Direct Evaluation of Chain-of-Thought in Multi-Hop Reasoning with
  Knowledge Graphs}.
\newblock In L.-W. Ku, A.~Martins, and V.~Srikumar, editors, {\em Findings of
  the Association for Computational Linguistics: ACL 2024}, pages 2862--2883,
  Bangkok, Thailand, Aug. 2024. Association for Computational Linguistics.

\bibitem{ning2023skeleton}
X.~Ning, Z.~Lin, Z.~Zhou, Z.~Wang, H.~Yang, and Y.~Wang.
\newblock {Skeleton-of-Thought: Prompting LLMs for Efficient Parallel
  Generation}.
\newblock In {\em Proceedings of the Twelfth International Conference on
  Learning Representations}, ICLR '24, Vienna, Austria, May 2024. OpenReview.

\bibitem{pan2023unifying}
S.~Pan, L.~Luo, Y.~Wang, C.~Chen, J.~Wang, and X.~Wu.
\newblock {Unifying Large Language Models and Knowledge Graphs: A Roadmap}.
\newblock {\em IEEE Transactions on Knowledge and Data Engineering},
  36(7):3580--3599, July 2024.

\bibitem{pan2023integrating}
S.~Pan, Y.~Zheng, and Y.~Liu.
\newblock {Integrating Graphs with Large Language Models: Methods and
  Prospects}.
\newblock {\em IEEE Intelligent Systems}, 39(1):64--68, Jan. 2024.

\bibitem{patel2021nlp}
A.~Patel, S.~Bhattamishra, and N.~Goyal.
\newblock {Are NLP Models Really Able to Solve Simple Math Word Problems?}
\newblock In K.~Toutanova, A.~Rumshisky, L.~Zettlemoyer, D.~Hakkani-Tur,
  I.~Beltagy, S.~Bethard, R.~Cotterell, T.~Chakraborty, and Y.~Zhou, editors,
  {\em Proceedings of the 2021 Conference of the North American Chapter of the
  Association for Computational Linguistics: Human Language Technologies},
  NAACL '21, pages 2080--2094, Virtual Event, June 2021. Association for
  Computational Linguistics.

\bibitem{press2210measuring}
O.~Press, M.~Zhang, S.~Min, L.~Schmidt, N.~Smith, and M.~Lewis.
\newblock {Measuring and Narrowing the Compositionality Gap in Language
  Models}.
\newblock In H.~Bouamor, J.~Pino, and K.~Bali, editors, {\em Findings of the
  Association for Computational Linguistics: EMNLP 2023}, pages 5687--5711,
  Singapore, Dec. 2023. Association for Computational Linguistics.

\bibitem{puri2021codenet}
R.~Puri, D.~S. Kung, G.~Janssen, W.~Zhang, G.~Domeniconi, V.~Zolotov, J.~Dolby,
  J.~Chen, M.~Choudhury, L.~Decker, V.~Thost, L.~Buratti, S.~Pujar, S.~Ramji,
  U.~Finkler, S.~Malaika, and F.~Reiss.
\newblock {CodeNet: A Large-Scale AI for Code Dataset for Learning a Diversity
  of Coding Tasks}, Aug. 2021.
\newblock arXiv:2105.12655.

\bibitem{qi2023art}
J.~Qi, Z.~Xu, Y.~Shen, M.~Liu, D.~Jin, Q.~Wang, and L.~Huang.
\newblock {The Art of SOCRATIC QUESTIONING: Recursive Thinking with Large
  Language Models}.
\newblock In H.~Bouamor, J.~Pino, and K.~Bali, editors, {\em Proceedings of the
  2023 Conference on Empirical Methods in Natural Language Processing}, EMNLP
  '23, pages 4177--4199, Singapore, Dec. 2023. Association for Computational
  Linguistics.

\bibitem{qian2023can}
C.~Qian, H.~Tang, Z.~Yang, H.~Liang, and Y.~Liu.
\newblock {Can Large Language Models Empower Molecular Property Prediction?},
  July 2023.
\newblock arXiv:2307.07443.

\bibitem{qiao2022reasoning}
S.~Qiao, Y.~Ou, N.~Zhang, X.~Chen, Y.~Yao, S.~Deng, C.~Tan, F.~Huang, and
  H.~Chen.
\newblock {Reasoning with Language Model Prompting: A Survey}.
\newblock In A.~Rogers, J.~Boyd-Graber, and N.~Okazaki, editors, {\em
  Proceedings of the 61st Annual Meeting of the Association for Computational
  Linguistics (Volume 1: Long Papers)}, ACL '23, pages 5368--5393, Toronto,
  Canada, July 2023. Association for Computational Linguistics.

\bibitem{radford2019language}
A.~Radford, J.~Wu, R.~Child, D.~Luan, D.~Amodei, and I.~Sutskever.
\newblock {Language Models are Unsupervised Multitask Learners}.
\newblock \url{https://openai.com/research/better-language-models}, 2019.
\newblock (accessed: Dec. 4, 2023).

\bibitem{raffel2023exploring}
C.~Raffel, N.~Shazeer, A.~Roberts, K.~Lee, S.~Narang, M.~Matena, Y.~Zhou,
  W.~Li, and P.~J. Liu.
\newblock {Exploring the Limits of Transfer Learning with a Unified
  Text-to-Text Transformer}.
\newblock {\em Journal of Machine Learning Research}, 21(140):1--67, 2020.

\bibitem{rasouli2019pie}
A.~Rasouli, I.~Kotseruba, T.~Kunic, and J.~Tsotsos.
\newblock {PIE: A Large-Scale Dataset and Models for Pedestrian Intention
  Estimation and Trajectory Prediction}.
\newblock In {\em Proceedings of the 2019 IEEE/CVF International Conference on
  Computer Vision}, ICCV '19, pages 6261--6270, Seoul, Korea, Nov. 2019. IEEE
  Press.

\bibitem{richards2023autogpt}
T.~B. Richards.
\newblock {AutoGPT: build \& use AI agents - Github}.
\newblock https://github.com/Significant-Gravitas/AutoGPT, Mar. 2023.
\newblock (accessed Jan. 23, 2024).

\bibitem{roy2015solving}
S.~Roy and D.~Roth.
\newblock {Solving General Arithmetic Word Problems}.
\newblock In L.~M{\`a}rquez, C.~Callison-Burch, and J.~Su, editors, {\em
  Proceedings of the 2015 Conference on Empirical Methods in Natural Language
  Processing}, EMNLP '15, pages 1743--1752, Lisbon, Portugal, Sept. 2015.
  Association for Computational Linguistics.

\bibitem{saha2023branch}
S.~Saha, O.~Levy, A.~Celikyilmaz, M.~Bansal, J.~Weston, and X.~Li.
\newblock {Branch-Solve-Merge Improves Large Language Model Evaluation and
  Generation}.
\newblock In K.~Duh, H.~Gomez, and S.~Bethard, editors, {\em Proceedings of the
  2024 Conference of the North American Chapter of the Association for
  Computational Linguistics: Human Language Technologies (Volume 1: Long
  Papers)}, NAACL '24, pages 8352--8370, Mexico City, Mexico, June 2024.
  Association for Computational Linguistics.

\bibitem{sakarvadia2023memory}
M.~Sakarvadia, A.~Ajith, A.~Khan, D.~Grzenda, N.~Hudson, A.~Bauer, K.~Chard,
  and I.~Foster.
\newblock {Memory Injections: Correcting Multi-Hop Reasoning Failures During
  Inference in Transformer-Based Language Models}.
\newblock In Y.~Belinkov, S.~Hao, J.~Jumelet, N.~Kim, A.~McCarthy, and
  H.~Mohebbi, editors, {\em Proceedings of the 6th BlackboxNLP Workshop:
  Analyzing and Interpreting Neural Networks for NLP}, BlackboxNLP '23, pages
  342--356, Singapore, Dec. 2023. Association for Computational Linguistics.

\bibitem{saparov2022language}
A.~Saparov and H.~He.
\newblock {Language Models Are Greedy Reasoners: A Systematic Formal Analysis
  of Chain-of-Thought}.
\newblock In {\em Proceedings of the Eleventh International Conference on
  Learning Representations}, ICLR '23, Kigali, Rwanda, May 2023. OpenReview.

\bibitem{schick2023toolformer}
T.~Schick, J.~Dwivedi-Yu, R.~Dess{\`\i}, R.~Raileanu, M.~Lomeli, E.~Hambro,
  L.~Zettlemoyer, N.~Cancedda, and T.~Scialom.
\newblock {Toolformer: Language Models Can Teach Themselves to Use Tools}.
\newblock In A.~Oh, T.~Naumann, A.~Globerson, K.~Saenko, M.~Hardt, and
  S.~Levine, editors, {\em Proceedings of the Thirty-seventh Annual Conference
  on Neural Information Processing Systems (NeurIPS '23)}, volume~36 of {\em
  Advances in Neural Information Processing Systems}, pages 68539--68551, New
  Orleans, LA, USA, Dec. 2023. Curran Associates.

\bibitem{sel2023algorithm}
B.~Sel, A.~Tawaha, V.~Khattar, R.~Jia, and M.~Jin.
\newblock {Algorithm of Thoughts: Enhancing Exploration of Ideas in Large
  Language Models}.
\newblock In R.~Salakhutdinov, Z.~Kolter, K.~Heller, A.~Weller, N.~Oliver,
  J.~Scarlett, and F.~Berkenkamp, editors, {\em Proceedings of the 41st
  International Conference on Machine Learning (ICML '24)}, volume 235 of {\em
  Proceedings of Machine Learning Research}, pages 44136--44189, Vienna,
  Austria, July 2024. PMLR.

\bibitem{seshadri2017ambit}
V.~Seshadri, D.~Lee, T.~Mullins, H.~Hassan, A.~Boroumand, J.~Kim, M.~A. Kozuch,
  O.~Mutlu, P.~B. Gibbons, and T.~C. Mowry.
\newblock {Ambit: In-Memory Accelerator for Bulk Bitwise Operations Using
  Commodity DRAM Technology}.
\newblock In {\em Proceedings of the 50th Annual IEEE/ACM International
  Symposium on Microarchitecture}, MICRO '17, pages 273--287, Cambridge, MA,
  USA, Oct. 2017. Association for Computing Machinery.

\bibitem{shadrikov2016psychological}
V.~Shadrikov, S.~Kurginyan, and O.~Martynova.
\newblock {Psychological Studies of Thought: Thoughts about a Concept of
  Thought}.
\newblock {\em Psychology. Journal of Higher School of Economics},
  13(3):558--575, 2016.

\bibitem{shinn2023reflexion}
N.~Shinn, F.~Cassano, A.~Gopinath, K.~Narasimhan, and S.~Yao.
\newblock {Reflexion: Language Agents with Verbal Reinforcement Learning}.
\newblock In A.~Oh, T.~Neumann, A.~Globerson, K.~Saenko, M.~Hardt, and
  S.~Levine, editors, {\em Proceedings of the Thirty-seventh Annual Conference
  on Neural Information Processing Systems (NeurIPS '23)}, volume~36 of {\em
  Advances in Neural Information Processing Systems}, pages 8634--8652, New
  Orleans, LA, USA, Dec. 2023. Curran Associates.

\bibitem{ALFWorld20}
M.~Shridhar, X.~Yuan, M.-A. C\^ot\'e, Y.~Bisk, A.~Trischler, and M.~Hausknecht.
\newblock {ALFWorld: Aligning Text and Embodied Environments for Interactive
  Learning}.
\newblock In {\em Proceedings of the Ninth International Conference on Learning
  Representations}, ICLR '21, Virtual Event, May 2021. OpenReview.

\bibitem{morris2023tree}
C.~Singh, J.~Morris, A.~Rush, J.~Gao, and Y.~Deng.
\newblock {Tree Prompting: Efficient Task Adaptation without Fine-Tuning}.
\newblock In H.~Bouamor, J.~Pino, and K.~Bali, editors, {\em Proceedings of the
  2023 Conference on Empirical Methods in Natural Language Processing}, EMNLP
  '23, pages 6253--6267, Singapore, Dec. 2023. Association for Computational
  Linguistics.

\bibitem{singh2022explaining}
C.~Singh, J.~X. Morris, J.~Aneja, A.~Rush, and J.~Gao.
\newblock {Explaining Data Patterns in Natural Language with Language Models}.
\newblock In Y.~Belinkov, S.~Hao, J.~Jumelet, N.~Kim, A.~McCarthy, and
  H.~Mohebbi, editors, {\em Proceedings of the 6th BlackboxNLP Workshop:
  Analyzing and Interpreting Neural Networks for NLP}, BlackboxNLP '23, pages
  31--55, Singapore, Dec. 2023. Association for Computational Linguistics.

\bibitem{smallwood2021neural}
J.~Smallwood, A.~Turnbull, H.~Wang, N.~S. Ho, G.~L. Poerio,
  T.~Karapanagiotidis, D.~Konu, B.~Mckeown, M.~Zhang, C.~Murphy, D.~Vatansever,
  D.~Bzdok, M.~Konishi, R.~Leech, P.~Seli, J.~W. Schooler, B.~Bernhardt, D.~S.
  Margulies, and E.~Jefferies.
\newblock {The Neural Correlates of Ongoing Conscious Thought}.
\newblock {\em iScience}, 24(3):102132, Mar. 2021.

\bibitem{srivastava2023beyond}
A.~Srivastava et~al.
\newblock {Beyond the Imitation Game: Quantifying and Extrapolating the
  Capabilities of Language Models}.
\newblock {\em Transactions on Machine Learning Research}, May 2023.
\newblock Featured Certification.

\bibitem{stelmakh2023asqa}
I.~Stelmakh, Y.~Luan, B.~Dhingra, and M.-W. Chang.
\newblock {ASQA: Factoid Questions Meet Long-Form Answers}.
\newblock In Y.~Goldberg, Z.~Kozareva, and Y.~Zhang, editors, {\em Proceedings
  of the 2022 Conference on Empirical Methods in Natural Language Processing},
  EMNLP '22, pages 8273--8288, Abu Dhabi, United Arab Emirates, Dec. 2022.
  Association for Computational Linguistics.

\bibitem{strausz2022asynchronous}
A.~Strausz, F.~Vella, S.~Di~Girolamo, M.~Besta, and T.~Hoefler.
\newblock {Asynchronous Distributed-Memory Triangle Counting and LCC with RMA
  Caching}.
\newblock In {\em Proceedings of the IEEE International Parallel and
  Distributed Processing Symposium}, IPDPS '22, pages 291--301, Lyon, France,
  June 2022. IEEE Press.

\bibitem{suhr2017corpus}
A.~Suhr, M.~Lewis, J.~Yeh, and Y.~Artzi.
\newblock {A Corpus of Natural Language for Visual Reasoning}.
\newblock In R.~Barzilay and M.-Y. Kan, editors, {\em Proceedings of the 55th
  Annual Meeting of the Association for Computational Linguistics (Volume 2:
  Short Papers)}, ACL '17, pages 217--223, Vancouver, Canada, July 2017.
  Association for Computational Linguistics.

\bibitem{sun2023think}
J.~Sun, C.~Xu, L.~Tang, S.~Wang, C.~Lin, Y.~Gong, L.~Ni, H.-Y. Shum, and
  J.~Guo.
\newblock {Think-on-Graph: Deep and Responsible Reasoning of Large Language
  Model on Knowledge Graph}.
\newblock In {\em Proceedings of the Twelfth International Conference on
  Learning Representations}, ICLR '24, Vienna, Austria, May 2024. OpenReview.

\bibitem{sun2023graph}
X.~Sun, J.~Zhang, X.~Wu, H.~Cheng, Y.~Xiong, and J.~Li.
\newblock {Graph Prompt Learning: A Comprehensive Survey and Beyond}, Nov.
  2023.
\newblock arXiv:2311.16534.

\bibitem{sun2012mining}
Y.~Sun and J.~Han.
\newblock {\em {Mining Heterogeneous Information Networks: Principles and
  Methodologies}}.
\newblock Synthesis Lectures on Data Mining and Knowledge Discovery (SLDMKD).
  Springer International Publishing, 2012.

\bibitem{tafjord2021proofwriter}
O.~Tafjord, B.~Dalvi, and P.~Clark.
\newblock {ProofWriter: Generating Implications, Proofs, and Abductive
  Statements over Natural Language}.
\newblock In C.~Zong, F.~Xia, W.~Li, and R.~Navigli, editors, {\em Findings of
  the Association for Computational Linguistics: ACL-IJCNLP 2021}, pages
  3621--3634, Virtual Event, Aug. 2021. Association for Computational
  Linguistics.

\bibitem{talmor-etal-2019-commonsenseqa}
A.~Talmor, J.~Herzig, N.~Lourie, and J.~Berant.
\newblock {CommonsenseQA: A Question Answering Challenge Targeting Commonsense
  Knowledge}.
\newblock In J.~Burstein, C.~Doran, and T.~Solorio, editors, {\em Proceedings
  of the 2019 Conference of the North American Chapter of the Association for
  Computational Linguistics: Human Language Technologies, Volume 1 (Long and
  Short Papers)}, NAACL '19, pages 4149--4158, Minneapolis, MN, USA, June 2019.
  Association for Computational Linguistics.

\bibitem{tang2023graphgpt}
J.~Tang, Y.~Yang, W.~Wei, L.~Shi, L.~Su, S.~Cheng, D.~Yin, and C.~Huang.
\newblock {GraphGPT: Graph Instruction Tuning for Large Language Models}.
\newblock In {\em Proceedings of the 47th International ACM SIGIR Conference on
  Research and Development in Information Retrieval}, SIGIR '24, pages
  491--500, Washington DC, USA, July 2024. Association for Computing Machinery.

\bibitem{thirunavukarasu2023large}
A.~J. Thirunavukarasu, D.~S.~J. Ting, K.~Elangovan, L.~Gutierrez, T.~F. Tan,
  and D.~S.~W. Ting.
\newblock {Large Language Models in Medicine}.
\newblock {\em Nature Medicine}, 29(8):1930--1940, July 2023.

\bibitem{tian2023graph}
Y.~Tian, H.~Song, Z.~Wang, H.~Wang, Z.~Hu, F.~Wang, N.~V. Chawla, and P.~Xu.
\newblock {Graph Neural Prompting with Large Language Models}.
\newblock {\em Proceedings of the AAAI Conference on Artificial Intelligence},
  38(17):19080--19088, Mar. 2024.

\bibitem{trivedi2022musique}
H.~Trivedi, N.~Balasubramanian, T.~Khot, and A.~Sabharwal.
\newblock {MuSiQue: Multihop Questions via Single-Hop Question Composition}.
\newblock {\em Transactions of the Association for Computational Linguistics},
  10:539--554, 2022.

\bibitem{tutunov2023can}
R.~Tutunov, A.~Grosnit, J.~Ziomek, J.~Wang, and H.~Bou-Ammar.
\newblock {Why Can Large Language Models Generate Correct Chain-of-Thoughts?},
  June 2024.
\newblock arXiv:2310.13571.

\bibitem{wang2023language}
H.~Wang, S.~Feng, T.~He, Z.~Tan, X.~Han, and Y.~Tsvetkov.
\newblock {Can Language Models Solve Graph Problems in Natural Language?}
\newblock In A.~Oh, T.~Neumann, A.~Globerson, K.~Saenko, M.~Hardt, and
  S.~Levine, editors, {\em Proceedings of the Thirty-seventh Annual Conference
  on Neural Information Processing Systems (NeurIPS '23)}, volume~36 of {\em
  Advances in Neural Information Processing Systems}, pages 30840--30861.
  Curran Associates, New Orleans, LA, USA, Dec. 2023.

\bibitem{wang2023graph}
H.~Wang, Y.~Gao, X.~Zheng, P.~Zhang, H.~Chen, and J.~Bu.
\newblock {Graph Neural Architecture Search with GPT-4}, Mar. 2024.
\newblock arXiv:2310.01436.

\bibitem{wang2023plan}
L.~Wang, W.~Xu, Y.~Lan, Z.~Hu, Y.~Lan, R.~K.-W. Lee, and E.-P. Lim.
\newblock {Plan-and-Solve Prompting: Improving Zero-Shot Chain-of-Thought
  Reasoning by Large Language Models}.
\newblock In A.~Rogers, J.~Boyd-Graber, and N.~Okazaki, editors, {\em
  Proceedings of the 61st Annual Meeting of the Association for Computational
  Linguistics (Volume 1: Long Papers)}, ACL '23, pages 2609--2634, Toronto,
  Canada, July 2023. Association for Computational Linguistics.

\bibitem{wang2019deep}
M.~Wang, D.~Zheng, Z.~Ye, Q.~Gan, M.~Li, X.~Song, J.~Zhou, C.~Ma, L.~Yu,
  Y.~Gai, T.~Xiao, T.~He, G.~Karypis, J.~Li, and Z.~Zhang.
\newblock {Deep Graph Library: A Graph-Centric, Highly-Performant Package for
  Graph Neural Networks}, Aug. 2020.
\newblock arXiv:1909.01315.

\bibitem{wang2022self}
X.~Wang, J.~Wei, D.~Schuurmans, Q.~Le, E.~Chi, and D.~Zhou.
\newblock {Self-Consistency Improves Chain of Thought Reasoning in Language
  Models}.
\newblock In {\em Proceedings of the Eleventh International Conference on
  Learning Representations}, ICLR '23, Kigali, Rwanda, May 2023. OpenReview.

\bibitem{wang2023knowledgpt}
X.~Wang, Q.~Yang, Y.~Qiu, J.~Liang, Q.~He, Z.~Gu, Y.~Xiao, and W.~Wang.
\newblock {KnowledGPT: Enhancing Large Language Models with Retrieval and
  Storage Access on Knowledge Bases}, Aug. 2023.
\newblock arXiv:2308.11761.

\bibitem{wang2023knowledge}
Y.~Wang, N.~Lipka, R.~A. Rossi, A.~Siu, R.~Zhang, and T.~Derr.
\newblock {Knowledge Graph Prompting for Multi-Document Question Answering}.
\newblock {\em Proceedings of the AAAI Conference on Artificial Intelligence},
  38(17):19206--19214, Mar. 2024.

\bibitem{wang2023describe}
Z.~Wang, S.~Cai, G.~Chen, A.~Liu, X.~S. Ma, and Y.~Liang.
\newblock {Describe, Explain, Plan and Select: Interactive Planning with LLMs
  Enables Open-World Multi-Task Agents}.
\newblock In A.~Oh, T.~Neumann, A.~Globerson, K.~Saenko, M.~Hardt, and
  S.~Levine, editors, {\em Proceedings of the Thirty-seventh Annual Conference
  on Neural Information Processing Systems (NeurIPS '23)}, volume~36 of {\em
  Advances in Neural Information Processing Systems}, pages 34153--34189, New
  Orleans, LA, USA, Dec. 2023. Curran Associates.

\bibitem{wang2023interactive}
Z.~Wang, G.~Zhang, K.~Yang, N.~Shi, W.~Zhou, S.~Hao, G.~Xiong, Y.~Li, M.~Y.
  Sim, X.~Chen, Q.~Zhu, Z.~Yang, A.~Nik, Q.~Liu, C.~Lin, S.~Wang, R.~Liu,
  W.~Chen, K.~Xu, D.~Liu, Y.~Guo, and J.~Fu.
\newblock {Interactive Natural Language Processing}, May 2023.
\newblock arXiv:2305.13246.

\bibitem{wei2022chain}
J.~Wei, X.~Wang, D.~Schuurmans, M.~Bosma, B.~Ichter, F.~Xia, E.~Chi, Q.~V. Le,
  and D.~Zhou.
\newblock {Chain-of-Thought Prompting Elicits Reasoning in Large Language
  Models}.
\newblock In S.~Koyejo, S.~Mohamed, A.~Agarwal, D.~Belgrave, K.~Cho, and A.~Oh,
  editors, {\em Proceedings of the Thirty-sixth Annual Conference on Neural
  Information Processing Systems (NeurIPS '22)}, volume~35 of {\em Advances in
  Neural Information Processing Systems}, pages 24824--24837, New Orleans, LA,
  USA, Dec. 2022. Curran Associates.

\bibitem{wei2021knowledge}
X.~Wei, S.~Wang, D.~Zhang, P.~Bhatia, and A.~Arnold.
\newblock {Knowledge Enhanced Pretrained Language Models: A Comprehensive
  Survey}, Oct. 2021.
\newblock arXiv:2110.08455.

\bibitem{wen2023mindmap}
Y.~Wen, Z.~Wang, and J.~Sun.
\newblock {MindMap: Knowledge Graph Prompting Sparks Graph of Thoughts in Large
  Language Models}.
\newblock In L.-W. Ku, A.~Martins, and V.~Srikumar, editors, {\em Proceedings
  of the 62nd Annual Meeting of the Association for Computational Linguistics
  (Volume 1: Long Papers)}, ACL '24, pages 10370--10388, Bangkok, Thailand,
  Aug. 2024. Association for Computational Linguistics.

\bibitem{weng2022large}
Y.~Weng, M.~Zhu, F.~Xia, B.~Li, S.~He, S.~Liu, B.~Sun, K.~Liu, and J.~Zhao.
\newblock {Large Language Models Are Better Reasoners with Self-Verification}.
\newblock In H.~Bouamor, J.~Pino, and K.~Bali, editors, {\em Findings of the
  Association for Computational Linguistics: EMNLP 2023}, pages 2550--2575,
  Singapore, Dec. 2023. Association for Computational Linguistics.

\bibitem{weston2023system}
J.~Weston and S.~Sukhbaatar.
\newblock {System 2 Attention (Is Something You Might Need Too)}, Nov. 2023.
\newblock arXiv:2311.11829.

\bibitem{wies2023learnability}
N.~Wies, Y.~Levine, and A.~Shashua.
\newblock {The Learnability of In-Context Learning}.
\newblock In A.~Oh, T.~Neumann, A.~Globerson, K.~Saenko, M.~Hardt, and
  S.~Levine, editors, {\em Proceedings of the Thirty-seventh Annual Conference
  on Neural Information Processing Systems (NeurIPS '23)}, volume~36 of {\em
  Advances in Neural Information Processing Systems}, pages 41618--41650, New
  Orleans, LA, USA, Dec. 2023. Curran Associates.

\bibitem{wu2024ocean}
J.~Wu, X.~Li, R.~Wang, Y.~Xia, Y.~Xiong, J.~Wang, T.~Yu, X.~Chen, B.~Kveton,
  L.~Yao, J.~Shang, and J.~McAuley.
\newblock {OCEAN: Offline Chain-of-Thought Evaluation and Alignment in Large
  Language Models}.
\newblock In {\em Proceedings of the Thirteenth International Conference on
  Learning Representations}, ICLR '25, Singapore, Apr. 2025. OpenReview.

\bibitem{wu2022promptchainer}
T.~Wu, E.~Jiang, A.~Donsbach, J.~Gray, A.~Molina, M.~Terry, and C.~J. Cai.
\newblock {PromptChainer: Chaining Large Language Model Prompts Through Visual
  Programming}.
\newblock In {\em Extended Abstracts of the Conference on Human Factors in
  Computing Systems}, CHI EA '22, pages 359:1--359:10, New Orleans, LA, USA,
  May 2022. Association for Computing Machinery.

\bibitem{wu2023survey}
X.~Wu, K.~Zhou, M.~Sun, X.~Wang, and N.~Liu.
\newblock {A Survey of Graph Prompting Methods: Techniques, Applications, and
  Challenges}, May 2023.
\newblock arXiv:2303.07275.

\bibitem{wu2020comprehensive}
Z.~Wu, S.~Pan, F.~Chen, G.~Long, C.~Zhang, and P.~S. Yu.
\newblock {A Comprehensive Survey on Graph Neural Networks}.
\newblock {\em IEEE Transactions on Neural Networks and Learning Systems},
  32(1):4--24, Jan. 2021.

\bibitem{xie2022explanation}
S.~M. Xie, A.~Raghunathan, P.~Liang, and T.~Ma.
\newblock {An Explanation of In-Context Learning as Implicit Bayesian
  Inference}.
\newblock In {\em Proceedings of the Tenth International Conference on Learning
  Representations}, ICLR '22, Kigali, Rwanda, Apr. 2022. OpenReview.

\bibitem{xie2023decomposition}
Y.~Xie, K.~Kawaguchi, Y.~Zhao, J.~X. Zhao, M.-Y. Kan, J.~He, and M.~Xie.
\newblock {Self-Evaluation Guided Beam Search for Reasoning}.
\newblock In A.~Oh, T.~Neumann, A.~Globerson, K.~Saenko, M.~Hardt, and
  S.~Levine, editors, {\em Proceedings of the Thirty-seventh Annual Conference
  on Neural Information Processing Systems (NeurIPS '23)}, volume~36 of {\em
  Advances in Neural Information Processing Systems}, pages 41618--41650, New
  Orleans, LA, USA, Dec. 2023. Curran Associates.

\bibitem{xu2023knn}
B.~Xu, Q.~Wang, Z.~Mao, Y.~Lyu, Q.~She, and Y.~Zhang.
\newblock {$k$NN Prompting: Beyond-Context Learning with Calibration-Free
  Nearest Neighbor Inference}.
\newblock In {\em Proceedings of the Eleventh International Conference on
  Learning Representations}, ICLR '23, Kigali, Rwanda, May 2023. OpenReview.

\bibitem{xu2023wizardlm}
C.~Xu, Q.~Sun, K.~Zheng, X.~Geng, P.~Zhao, J.~Feng, C.~Tao, Q.~Lin, and
  D.~Jiang.
\newblock {WizardLM: Empowering Large Pre-Trained Language Models to Follow
  Complex Instructions}.
\newblock In {\em Proceedings of the Twelfth International Conference on
  Learning Representations}, ICLR '24, Vienna, Austria, May 2024. OpenReview.

\bibitem{yang2021survey}
J.~Yang, X.~Hu, G.~Xiao, and Y.~Shen.
\newblock {A Survey of Knowledge Enhanced Pre-Trained Language Models}.
\newblock {\em ACM Trans. Asian Low-Resour. Lang. Inf. Process.}, Mar. 2024.

\bibitem{yang2023chatgpt}
L.~Yang, H.~Chen, Z.~Li, X.~Ding, and X.~Wu.
\newblock {Give us the Facts: Enhancing Large Language Models With Knowledge
  Graphs for Fact-Aware Language Modeling}.
\newblock {\em IEEE Transactions on Knowledge and Data Engineering},
  36(7):3091--3110, July 2024.

\bibitem{yang2018hotpotqa}
Z.~Yang, P.~Qi, S.~Zhang, Y.~Bengio, W.~Cohen, R.~Salakhutdinov, and C.~D.
  Manning.
\newblock {HotpotQA: A Dataset for Diverse, Explainable Multi-Hop Question
  Answering}.
\newblock In E.~Riloff, D.~Chiang, J.~Hockenmaier, and J.~Tsujii, editors, {\em
  Proceedings of the 2018 Conference on Empirical Methods in Natural Language
  Processing}, EMNLP '18, pages 2369--2380, Brussels, Belgium, Nov. 2018.
  Association for Computational Linguistics.

\bibitem{yao2023thinking}
F.~Yao, C.~Tian, J.~Liu, Z.~Zhang, Q.~Liu, L.~Jin, S.~Li, X.~Li, and X.~Sun.
\newblock {Thinking Like an Expert: Multimodal Hypergraph-of-Thought (HoT)
  Reasoning to Boost Foundation Modals}, Aug. 2023.
\newblock arXiv:2308.06207.

\bibitem{yao2023tree}
S.~Yao, D.~Yu, J.~Zhao, I.~Shafran, T.~Griffiths, Y.~Cao, and K.~Narasimhan.
\newblock {Tree of Thoughts: Deliberate Problem Solving with Large Language
  Models}.
\newblock In A.~Oh, T.~Neumann, A.~Globerson, K.~Saenko, M.~Hardt, and
  S.~Levine, editors, {\em Proceedings of the Thirty-seventh Annual Conference
  on Neural Information Processing Systems (NeurIPS '23)}, volume~36 of {\em
  Advances in Neural Information Processing Systems}, pages 11809--11822, New
  Orleans, LA, USA, Dec. 2023. Curran Associates.

\bibitem{yao2022react}
S.~Yao, J.~Zhao, D.~Yu, N.~Du, I.~Shafran, K.~R. Narasimhan, and Y.~Cao.
\newblock {ReAct: Synergizing Reasoning and Acting in Language Models}.
\newblock In {\em Proceedings of the Eleventh International Conference on
  Learning Representations}, ICLR '23, Kigali, Rwanda, May 2023. OpenReview.

\bibitem{yao2023beyond}
Y.~Yao, Z.~Li, and H.~Zhao.
\newblock {GoT: Effective Graph-of-Thought Reasoning in Language Models}.
\newblock In K.~Duh, H.~Gomez, and S.~Bethard, editors, {\em Findings of the
  Association for Computational Linguistics: NAACL 2024}, pages 2901--2921,
  Mexico City, Mexico, June 2024. Association for Computational Linguistics.

\bibitem{ye2023natural}
R.~Ye, C.~Zhang, R.~Wang, S.~Xu, and Y.~Zhang.
\newblock {Language Is All a Graph Needs}.
\newblock In Y.~Graham and M.~Purver, editors, {\em Findings of the Association
  for Computational Linguistics: EACL 2024}, pages 1955--1973, St. Julian's,
  Malta, Mar. 2024. Association for Computational Linguistics.

\bibitem{ye2023large}
Y.~Ye, B.~Hui, M.~Yang, B.~Li, F.~Huang, and Y.~Li.
\newblock {Large Language Models Are Versatile Decomposers: Decomposing
  Evidence and Questions for Table-Based Reasoning}.
\newblock In {\em Proceedings of the 46th International ACM SIGIR Conference on
  Research and Development in Information Retrieval}, SIGIR '23, pages
  174--184, Taipei, Taiwan, 2023. Association for Computing Machinery.

\bibitem{yu2023thought}
J.~Yu, R.~He, and R.~Ying.
\newblock {Thought Propagation: An Analogical Approach to Complex Reasoning
  with Large Language Models}.
\newblock In {\em Proceedings of the Twelfth International Conference on
  Learning Representations}, ICLR '24, Vienna, Austria, May 2024. OpenReview.

\bibitem{zeitforward}
L.~Zeit-Altpeter.
\newblock {Forward Looking Active Retrieval Augmented Generation}.
\newblock In {\em Proceedings of the AI Summer School at Uni Jena (Poster)},
  2023.

\bibitem{zhang2019heterogeneous}
C.~Zhang, D.~Song, C.~Huang, A.~Swami, and N.~V. Chawla.
\newblock {Heterogeneous Graph Neural Network}.
\newblock In {\em Proceedings of the 25th ACM SIGKDD International Conference
  on Knowledge Discovery \& Data Mining}, KDD '19, pages 793--803, Anchorage,
  AK, USA, Aug. 2019. Association for Computing Machinery.

\bibitem{zhang2022paradox}
H.~Zhang, L.~H. Li, T.~Meng, K.-W. Chang, and G.~V. den Broeck.
\newblock {On the Paradox of Learning to Reason from Data}.
\newblock In E.~Elkind, editor, {\em Proceedings of the Thirty-Second
  International Joint Conference on Artificial Intelligence}, IJCAI '23, pages
  3365--3373, Macao, SAR, Aug. 2023. International Joint Conferences on
  Artificial Intelligence Organization.
\newblock Main Track.

\bibitem{zhang2023graph}
J.~Zhang.
\newblock {Graph-ToolFormer: To Empower LLMs with Graph Reasoning Ability via
  Prompt Augmented by ChatGPT}, May 2023.
\newblock arXiv:2304.11116.

\bibitem{zhang2023making}
Y.~Zhang, Z.~Chen, L.~Guo, Y.~Xu, W.~Zhang, and H.~Chen.
\newblock {Making Large Language Models Perform Better in Knowledge Graph
  Completion}.
\newblock In {\em Proceedings of the 32nd ACM International Conference on
  Multimedia}, MM '24, pages 233--242, Melbourne, Australia, Oct. 2024.
  Association for Computing Machinery.

\bibitem{zhang2024small}
Y.~Zhang, M.~Khalifa, L.~Logeswaran, J.~Kim, M.~Lee, H.~Lee, and L.~Wang.
\newblock {Small Language Models Need Strong Verifiers to Self-Correct
  Reasoning}.
\newblock In L.-W. Ku, A.~Martins, and V.~Srikumar, editors, {\em Findings of
  the Association for Computational Linguistics: ACL 2024}, pages 15637--15653,
  Bangkok, Thailand, Aug. 2024. Association for Computational Linguistics.

\bibitem{zhang2023cumulative}
Y.~Zhang, J.~Yang, Y.~Yuan, and A.~C. Yao.
\newblock {Cumulative Reasoning with Large Language Models}.
\newblock {\em Transactions on Machine Learning Research}, July 2025.

\bibitem{zhang2020deep}
Z.~Zhang, P.~Cui, and W.~Zhu.
\newblock {Deep Learning on Graphs: A Survey}.
\newblock {\em IEEE Transactions on Knowledge and Data Engineering},
  34(1):249--270, Jan. 2022.

\bibitem{zhang2023large}
Z.~Zhang, H.~Li, Z.~Zhang, Y.~Qin, X.~Wang, and W.~Zhu.
\newblock {Graph Meets LLMs: Towards Large Graph Models}.
\newblock In {\em Proceedings of the Workshop on New Frontiers in Graph
  Learning}, GLFrontiers '23, New Orleans, LA, USA, Dec. 2023. OpenReview.

\bibitem{zhang2023llm4dyg}
Z.~Zhang, X.~Wang, Z.~Zhang, H.~Li, Y.~Qin, and W.~Zhu.
\newblock {LLM4DyG: Can Large Language Models Solve Spatial-Temporal Problems
  on Dynamic Graphs?}
\newblock In {\em Proceedings of the 30th ACM SIGKDD Conference on Knowledge
  Discovery and Data Mining}, KDD '24, pages 4350--4361, Barcelona, Spain, Aug.
  2024. Association for Computing Machinery.

\bibitem{zhang2023igniting}
Z.~Zhang, Y.~Yao, A.~Zhang, X.~Tang, X.~Ma, Z.~He, Y.~Wang, M.~Gerstein,
  R.~Wang, G.~Liu, and H.~Zhao.
\newblock {Igniting Language Intelligence: The Hitchhiker's Guide From
  Chain-of-Thought Reasoning to Language Agents}, Nov. 2023.
\newblock arXiv:2311.11797.

\bibitem{zhao2023graphtext}
J.~Zhao, L.~Zhuo, Y.~Shen, M.~Qu, K.~Liu, M.~Bronstein, Z.~Zhu, and J.~Tang.
\newblock {GraphText: Graph Reasoning in Text Space}.
\newblock In {\em Proceedings of the Workshop on Adaptive Foundation Models:
  Evolving AI for Personalized and Efficient Learning}, AFM '24, Vancouver,
  Canada, Dec. 2024. OpenReview.

\bibitem{zhao2023survey}
W.~X. Zhao, K.~Zhou, J.~Li, T.~Tang, X.~Wang, Y.~Hou, Y.~Min, B.~Zhang,
  J.~Zhang, Z.~Dong, Y.~Du, C.~Yang, Y.~Chen, Z.~Chen, J.~Jiang, R.~Ren, Y.~Li,
  X.~Tang, Z.~Liu, P.~Liu, J.-Y. Nie, and J.-R. Wen.
\newblock {A Survey of Large Language Models}, Mar. 2025.
\newblock arXiv:2303.18223.

\bibitem{zhao2023enhancing}
X.~Zhao, M.~Li, W.~Lu, C.~Weber, J.~H. Lee, K.~Chu, and S.~Wermter.
\newblock {Enhancing Zero-Shot Chain-of-Thought Reasoning in Large Language
  Models Through Logic}.
\newblock In N.~Calzolari, M.-Y. Kan, V.~Hoste, A.~Lenci, S.~Sakti, and N.~Xue,
  editors, {\em Proceedings of the 2024 Joint International Conference on
  Computational Linguistics, Language Resources and Evaluation}, LREC-COLING
  '24, pages 6144--6166, Torino, Italia, May 2024. ELRA and ICCL.

\bibitem{zheng2023judging}
L.~Zheng, W.-L. Chiang, Y.~Sheng, S.~Zhuang, Z.~Wu, Y.~Zhuang, Z.~Lin, Z.~Li,
  D.~Li, E.~Xing, H.~Zhang, J.~E. Gonzalez, and I.~Stoica.
\newblock {Judging LLM-as-a-Judge with MT-Bench and Chatbot Arena}.
\newblock In A.~Oh, T.~Neumann, A.~Globerson, K.~Saenko, M.~Hardt, and
  S.~Levine, editors, {\em Proceedings of the Thirty-seventh Annual Conference
  on Neural Information Processing Systems (NeurIPS '23)}, volume~36 of {\em
  Advances in Neural Information Processing Systems}, pages 46595--46623, New
  Orleans, LA, USA, Dec. 2023. Curran Associates.

\bibitem{zhou2022least}
D.~Zhou, N.~Sch{\"a}rli, L.~Hou, J.~Wei, N.~Scales, X.~Wang, D.~Schuurmans,
  C.~Cui, O.~Bousquet, Q.~V. Le, and E.~H. Chi.
\newblock {Least-to-Most Prompting Enables Complex Reasoning in Large Language
  Models}.
\newblock In {\em Proceedings of the Eleventh International Conference on
  Learning Representations}, ICLR '23, Kigali, Rwanda, May 2023. OpenReview.

\bibitem{zhou2020graph}
J.~Zhou, G.~Cui, S.~Hu, Z.~Zhang, C.~Yang, Z.~Liu, L.~Wang, C.~Li, and M.~Sun.
\newblock {Graph Neural Networks: A Review of Methods and Applications}.
\newblock {\em AI Open}, 1:57--81, 2020.

\bibitem{zhu2023llms}
Y.~Zhu, X.~Wang, J.~Chen, S.~Qiao, Y.~Ou, Y.~Yao, S.~Deng, H.~Chen, and
  N.~Zhang.
\newblock {LLMs for Knowledge Graph Construction and Reasoning: Recent
  Capabilities and Future Opportunities}.
\newblock {\em World Wide Web}, 27(5), Aug. 2024.

\bibitem{zong2023self}
Y.~Zong, O.~M. Aodha, and T.~Hospedales.
\newblock {Self-Supervised Multimodal Learning: A Survey}.
\newblock {\em IEEE Transactions on Pattern Analysis and Machine Intelligence},
  pages 1--20, 2024.

\end{thebibliography}

\ifcnf

\vspace{-3.5em}
\begin{IEEEbiography}
[{\includegraphics[width=1in,height=1.25in,clip,keepaspectratio]{mbesta}}]
{Maciej Besta}
is a researcher at ETH Zurich. He works on understanding and accelerating
large-scale irregular computations, such as graph streaming, graph neural
networks, or graph databases, at all levels of the computing stack.
\end{IEEEbiography}
\vspace{-4em}
\begin{IEEEbiography}
[{\includegraphics[width=1in,height=1.25in,clip,keepaspectratio]{htor}}]
{Torsten Hoefler}
is a Professor at ETH Zurich, where he leads the Scalable Parallel Computing
Lab. His research aims at understanding performance of parallel computing
systems ranging from parallel computer architecture through parallel programming
to parallel algorithms.
\end{IEEEbiography}

\fi

\iftr
\appendices
\section{Detailed Descriptions of Chain Schemes}





We study prompting schemes that use chain topologies. We list them and determine their relationship to our blueprint in Table~\ref{tab:schemes}.
%
%

\subsection{Single-Prompt Schemes}

Chain-based reasoning can be conducted within a single prompt, potentially containing multiple in-context examples. It is used in both textual and visual reasoning tasks. 
%


In the seminal \textbf{Chain-of-Thought (CoT)}~\cite{wei2022chain} scheme, in-context examples consisting of intermediate reasoning steps for a given complex question are introduced to prompt the LLM to generate a similar multi-step reasoning chain. This chain is used to improve the quality of the LLM answer compared to basic IO prompting.

\textbf{Zero-shot-CoT} was proposed by Kojima et al.~\cite{kojima2022large}. It comes with multi-step reasoning \emph{without} in-context examples, by simply prompting the LLM with one sentence, \emph{``Let's think step by step''}, or using other similar statements. This zero-shot prompting improves upon other zero-shot methods in terms of the accuracy of reasoning outcomes. 

%

\textbf{SelfAsk}~\cite{press2210measuring} is a single-prompt scheme similar to CoT which augments the in-context examples with intermediate questions. Instead of only providing a step-by-step reasoning chain in the examples, each step in this chain is expanded to pose a follow-up question and then answer it, e.g.,  \emph{``Superconductivity was discovered in 1911."} is augmented to  \emph{``Follow up: When was superconductivity discovered? Intermediate answer: Superconductivity was discovered in 1911"}. 

\textbf{Plan-and-Solve (PS) Prompting}~\cite{wang2023plan} builds a chain based on a devised plan, harnessing Zero-shot CoT. It first divides the complex task into a list of sub-tasks and then executes the stepwise plan for the solution.
%
PS can be extended to \textbf{PS+} by adding instructions in the prompt to request the LLM to extract variables and explicitly calculate the intermediate values. PS+ has no impact on the length of the chain.

\textbf{Program of Thoughts (PoT)}~\cite{chen2022program} is a single-prompt scheme similar to CoT which generates code to solve a question. Rather than the natural-language-based examples used in CoT, the LLM is prompted by code exemplars, where each statement refers to one intermediate step in the reasoning chain. Zero-shot prompting can also be applied, e.g., \emph{``Let's write a Python program step by step and return the result. Firstly we need to define the variables."}. The final result is obtained by executing the generated code.

\subsection{Multi-Prompt Schemes}
\label{subsec:multi-prompt}


%
%
Elicited by multiple rounds of prompting, the LLM reasoning process can be a chain consisting of several messages.
\textbf{Selection-Inference (SI)}~\cite{creswell2022selection} aims to address multi-step logical reasoning problems where all necessary information is provided within the input context.
Each reasoning step in SI involves two distinct sub-tasks: selection and inference. 
Firstly, the selection sub-task is responsible for identifying and selecting relevant information needed for the subsequent reasoning step, effectively pruning the context. 
Following this, the inference sub-task generates a new intermediate piece of information, thereby performing the actual reasoning step. This newly generated information then becomes available for use in subsequent reasoning steps for the selection sub-task. The length of the chain is bounded to a  fixed number of steps, and the prompts for both sub-tasks include relevant few-shot examples.

\textbf{Chain-of-symbol (CoS)}~\cite{hu2023chain} is a multi-prompt scheme utilizing two prompts for solving spatial planning tasks in natural language. 
%
Prompted in a zero-shot manner, the LLM initially generates CoT-like in-context exemplars and modifies them by replacing the spatial relationships with symbols. For example, the navigational task \emph{``Start at bank A. There are two stores on the map, store B and store G. The road from bank A to store B goes through bank C, house H, and cinema F to store B, totaling 600 meters.''} is succinctly converted into the symbol sequence \emph{``bank A / bank C / house H / cinema F / store B, (200 + 100 + 100 + 200 = 600).''}. Exemplars simplified in this way are then provided as prompts for LLMs to conduct CoT-based reasoning.



\subsubsection{Multi-Prompt Chains with Decomposition}
The introduction of a node for decomposition at the start of a reasoning chain can not only determine the chain's depth, but also facilitate more detailed reasoning approaches in the subsequent sub-steps. This allows for a finer-grained resolution of sub-questions, enriching the overall reasoning process. Apart from the schemes described below, similar strategies are applied in works~\cite{Dua2022SuccessivePF, jung2022maieutic, ye2023large}.

\textbf{Least-to-Most Prompting}~\cite{zhou2022least} grows a reasoning chain, where the decomposition of complex tasks is conducted in the first node and the sub-tasks are solved in the subsequent nodes. Prompted with few-shot examples showcasing the decomposition of questions into lists of sub-questions, the LLM firstly generates a sequence of sub-questions given an original question. Then the solving process starts by providing the LLM with example solutions for simple questions. Subsequently during each sub-step, the LLM generates the sub-solution, given the question for this step and the list of previously solved questions with the generated answers. The chain terminates when the final answer is returned after solving all sub-problems in the list.



\textbf{Decomposed Prompting}~\cite{khot2022decomposed} is a modular framework for a detailed decomposition of complex tasks. To generate a reasoning chain, the LLM is prompted by demonstrations comprised of sequential question-operation triplets, which form ``sub-questions''.
The growth of the chain follows the autoregressive pattern of CoT, while each sub-step captures the input and passes the output through the contextual environment. Under the limited scope of the tasks, the types of operation functions are fixed, enabling the LLM to conduct in-context learning for operations before inference.







\subsubsection{Multi-Prompt Chains with Verification}

The introduction of verification enables the reasoning frameworks to iteratively refine the generated context. With this strategy, the execution of chain-based reasoning is effectively extended with loops, with conditions on how many times one can loop over a node (based on the number of iterations or some terminal conditions). Apart from the schemes described below, similar strategies are applied in works~\cite{li2023makingstep, ling2023deductive, weng2022large}.


\textbf{LogiCoT}~\cite{zhao2023enhancing} is a zero-shot, multi-prompt framework that leverages iterative verification to minimize cumulative errors in the reasoning chain. After generating a reasoning chain through few-shot CoT, this framework systematically navigates the reasoning chain with zero-shot prompts to pinpoint the first instance where the original reasoning deviated. It preserves the accurate part of the reasoning chain up to the erroneous node and then prompts the LLM to diagnose and correct this specific misstep, thereby forming an alternative reasoning chain. This process repeats until a fully verified reasoning chain is established, with each node passing the verification.

\textbf{SELF-REFINE}~\cite{madaan2023self} adopts a distinct approach to iterative refinement compared to {LogiCoT}. Instead of focusing on pinpointing and correcting errors within a reasoning chain, it initially generates a complete output and then enters a cyclical process, alternating between generating specific feedback with concrete actions for enhancement and applying this feedback to refine the output. 

Similar to {SELF-REFINE}, \textbf{Reflexion}~\cite{shinn2023reflexion} employs the concept of iterative refinement through generating and applying feedback, but introduces a modular approach to the process, segmenting the refinement into specialized stages. It starts with an initial output by the LLM, followed by an evaluation through task-specific grading functions, and then generates informative feedback, which is subsequently applied. This reasoning chain continues looping until the evaluation stage confirms that the output is correct.

\textbf{Reasoning Graph Verifier (RGV)}~\cite{cao2023enhancing} is a framework that addresses arithmetic questions. Their approach is a variant of CoT-SC, in which they employ a trainable verifier module to decide on the best CoT solution.

\subsubsection{Multi-Prompt Chains with External Tools}


To better integrate multiple execution methods, some schemes opt to devise a plan that specifies tools for handling each sub-task, before executing the reasoning chain.

%
\textbf{Plan, Verify and Switch (PVS)}~\cite{liu2023plan} iteratively engages in planning, verification, and switching among different methods until the acquisition of the final answer. Initially, the LLM is prompted with definitions of three chain-based reasoning methods, which include CoT, PoT, and a method using the solution of linear equations to represent the reasoning process (named EoT). Then, at the first stage of each iteration, the LLM selects a reasoning method according to the input question and subsequently follows in-context examples on the application of the method to complete a reasoning chain accordingly.
Then, the verification module acquires conditional information and the intermediate variables from the sub-steps of the chain and executes calculations, validating the correctness of the final answer. If the verification fails, the next iteration is activated, in which the LLM selects a new method from the non-selected ones and generates a new chain without memories of the previous reasoning process.
The planning module facilitates the dynamic switch of reasoning methods, thereby integrating the strengths of different paradigms.

%
\textbf{Chameleon}~\cite{lu2023chameleon} is a reasoning framework that provides the LLM with access to various external tools, such as programming interpreters and table readers. Given the question, tool descriptions, and tool usage examples, the LLM generates a plan for the reasoning chain, in which each node corresponds to a tool used in the sub-step. During the execution of the chain, the output of previous steps together with the cached tool execution results are taken as the input of each current step. The nodes of the chain are executed sequentially until the final result is returned.

There are also tool-based chain schemes with no planning module; they dynamically select the reasoning method.

\textbf{ChatCoT}~\cite{chen2023chatcot} realizes tool-based reasoning chains through multi-round conversations with the LLM. Initially, the LLM is shown how to decompose CoT reasoning as a multi-turn conversation, in order to learn problem-solving in a step-by-step, autoregressive conversational way. Additionally, the knowledge of tools and relevant examples are included in the context to elicit their selection and correct usage. During inference, the chain proceeds and unfolds without a premature plan until the problem is solved, while the LLM simultaneously selects and executes appropriate tools during the multi-turn conversation. The conversation is constrained to a maximum number of turns, thereby constraining the maximum reasoning depth, while ensuring all previous turns are kept within the context window.

\section{Detailed Descriptions of Tree Schemes}

We next analyze works that harness trees as the prompting topology. We structure the discussion based on the harnessed topology variants, see Figure~\ref{fig:topologies-dets}.

\subsection{Trees of Chains}

While harnessing trees as prompting topologies has been explicitly established in the works by Long~\cite{long2023large} and Yao~\cite{yao2023tree}, this idea has been present earlier.
%
%
%
%
Specifically, \textbf{Chain-of-Thought with Self-Consistency (CoT-SC)}~\cite{wang2022self} is an early scheme that harnesses the tree structure to a certain degree. Here, multiple CoTs originate from the same initial (root) prompt, forming a ``tree of chains''. The chain providing the best outcome to the initial question, is selected as the final answer.

\subsection{Single-Level Trees}

\textbf{Skeleton-of-Thought (SoT)}~\cite{ning2023skeleton} is a prompting scheme aiming to reduce the end-to-end generation latency of LLMs, caused by their inherent sequential decoding. Instead of generating one long continuous answer, this scheme uses a divide-and-conquer approach. In a first prompt, the LLM is instructed to generate a skeleton of the answer, i.e., a list of points that are independently answerable. Then, for each of these points, a new prompt is issued in parallel to answer just this specific part of the question. As these points are processed in parallel, the overall latency is reduced.  One can additionally include a prompt at the beginning which lets the LLM decide on solving the question via independent points, or -- whenever reasonable -- by using a single continuous answer.

\subsection{$k$--Ary Trees}

First, the \textbf{Tree-of-Thought (ToT) design by Long}~\cite{long2023large} utilizes a tree structure to decompose a problem into sub-problems and solve them using separate LLM prompts. After the LLM suggests possible next steps and corresponding partial solutions, a \emph{checker module} decides if any of these solutions is valid, whether it can be selected as the final one, or whether it should backtrack to the previous step. All issued prompts and answers are explicitly stored as a tree structure and navigated through using a \emph{controller module}. The LLM prompting is only used to generate the next individual steps (i.e., hops) in this tree, whereas the overall problem solving process is coordinated by the controller. 

\textbf{Tree of Thoughts (ToT) by Yao et al.}~\cite{yao2023tree} differs from the above ToT approach in using the LLM itself as a solution \emph{evaluator} with access to all generated solutions, instead of using a programmed or learned evaluator module. This allows to rate states individually or vote across intermediate solutions to select the most promising one to continue with the search. Both mentioned ToT approaches are a generalization of the IO, CoT, and CoT-SC prompting schemes.

\subsubsection{Pre-ToT Schemes}

\textbf{Thought Decomposition}~\cite{xie2023decomposition} is a multi-prompt
scheme based on stochastic beam search and self-evaluation. At each level
(reasoning step), $n$ new intermediate nodes are generated for each of the
(usually) $k$ input nodes. Each node of this set of $nk$ nodes is
evaluated and then the set is pruned down to $k$ output nodes via sampling
with stochastic beam search.
For both the generation and the evaluation prompts, few-shot examples are used
with a focus on textual or program-aided reasoning, depending on the dataset
used for evaluation.

\textbf{Creswell and Shanahan}~\cite{creswell2022faithful} describe an extension of the chain-based Selection-Inference~\cite{creswell2022selection}, which is used to answer multiple-choice questions by chaining together correct reasoning steps over a predefined context, i.e., a set of statements (the provided statements are sufficient to derive the answer to the question). The space of all possible reasoning chains forms a tree rooted at the initial question. Each node represents a reasoning step that is derived by (1) selecting a subset of statements from the context and then (2) inferring a new statement, which is then added to the context. For both steps, the authors use separately fine-tuned LMs. This tree is explored using beam search, where another fine-tuned LM assesses the value of the current node, and the search ends as soon as a halter-LM decides that the question can be answered with the current context and then answers it.
\textbf{Dynamic Least-to-Most Prompting}~\cite{drozdov2022compositional} extends
least-to-most prompting with a tree-based problem decomposition and a dynamic
external tree-based few-shot example selection. The goal is to turn natural
language questions into formal representations such as SPARQL queries. Initially
the input question is decomposed into sub-problems using a series of prompts.
This process yields an arbitrary decision tree for the input in contrast to a chain in
traditional least-to-most prompting. This decision tree is used to
dynamically select few-shot examples by externally matching that tree against a
precomputed decision tree for examples. In the final step, the decision
tree is linearized into a sequence of sub-problems with increasing complexity
which are then translated by using prompts enhanced with the respective examples
selected in the second step.
%


\subsubsection{Post-ToT Schemes}

Different schemes have extended the initial ToT designs.
\textbf{Algorithm of Thoughts (AoT)}~\cite{sel2023algorithm} is a single-prompt approach that utilizes in-context examples formulated in an algorithmic fashion. Instead of providing step-by-step examples as in CoT prompting, AoT harnesses algorithmic reasoning steps in the examples, that explore the solution space (tree) with either DFS or BFS. 
As opposed to ToT, AoT uses only a single prompt.
%

\textbf{Tree of Uncertain Thought (TouT)}~\cite{mo2023tree} extends ToT with local ``uncertainty scores'' by incorporating the variance of multiple LLM responses into the state evaluation function.

\textbf{Tree-of-Mixed-Thought (TomT)}~\cite{hu2023treeofmixedthought} uses ToT-based prompting to answer questions on visual scene graphs, e.g., \emph{``Does the red chair have the same number of legs as the table in front of it?''}. The LLM is tasked to generate Python code that solves the question using a set of provided functions to extract and reason on the data of the scene graph. This prompting scheme uses a DFS variant of ToT where each individual node generates up to $s$ consecutive reasoning steps, i.e. lines of code, opposed to the original ToT where $s=1$. Evaluating the correctness of the reasoning steps is done as in ToT for each node, where one checks if the generated code so far can (1) run correctly, and (2) the interactions with the scene graph are consistent with the existing elements in the scene.

\textbf{Tree of Clarifications (ToC)}~\cite{kim2023tree} enables to answer ambiguous questions by first retrieving pertinent external information and then recursively prompting an LLM to construct a tree of disambiguations for the initial question. This tree is explored using BFS and after no more ambiguations exist, a long form answer is generated, by combining all the previously explored nodes.

\textbf{Tree Prompting}~\cite{morris2023tree} is a high-level approach that
proposes the training of a binary decision tree to classify text during
inference. Several mechanisms to construct the prompts that make up the nodes of
the decision tree during learning are proposed: prompts based on few-shot
examples, human curated instruction prompts, dynamic prompts constructed with
discrete prompt search methods like iPrompt~\cite{singh2022explaining}, and
$k$NN prompting~\cite{xu2023knn}, which constructs new nodes based on the
nearest neighbors. $k$NN prompting can result in multiple prompts per decision
tree node with the other mechanisms using only a single prompt. Additional
improvements can be made by employing tree ensembles such as random
forests~\cite{breiman1984classification} or gradient-boosted
trees~\cite{freund1996experiments}. During inference the number of prompts is
bound by the depth of the decision tree, where as the training can be resource
intensive.

\subsection{Analysis \& Comparison of Designs}

We now broadly discuss and analyze tree designs with respect to different aspects of our blueprint. A detailed analysis can be found in Appendix \ref{sec: tree_extensive}.



\subsubsection{Topology \& Its Construction}

The key novel architectural feature of tree schemes is \textbf{exploration} of a step, i.e., the ability to generate multiple new steps based on a given single one.
The vast majority of tree schemes are multi-prompt.
Most multi-prompt schemes use a dynamic approach to building the tree topology. The details of how the topology is exactly shaped depend on the specific question.
For most multi-prompt approaches, the user can adapt the tree topology to a certain degree, i.e., by varying the branching factor and limiting the depth of the tree.

\subsubsection{Performance}

Increasing the \textbf{branching factor} often leads to more \textbf{diversity} of outcomes, which can be beneficial for accuracy, but also increases the number of prompts, i.e., computational cost.
The most advantageous branching factor is hard to find and often depends on the specific problem to solve.
Easily decomposable problems may benefit less from more branching than complex problems. Specifically, more complicated problems profit more from decomposing them into \mbox{many/diverse} sub-problems (e.g., this ensures enough diversity for self-consistency to work better). In contrast, a question that has clearly only two sub-parts does not benefit from many more subdivisions, as the additional branches can either be redundant or wrong.
Single-prompt approaches can perform better on some problems than multi-prompt approaches, while using only a single prompt compared to possibly hundreds. 

%

\section{Detailed Descriptions of Graph Schemes}

We finally describe schemes with topologies beyond trees or chains. We observe that they can be further grouped into subclasses, based on the harnessed class of graphs. We picture different types of graph-based schemes in Figure~\ref{fig:topologies-dets}.

\subsection{Special Classes of Graphs}

Different schemes harness certain classes of graphs.

%
\textbf{Branch-Solve-Merge (BSM)}~\cite{saha2023branch} employs a 1--level double tree structure to first divide a problem into independently solvable sub-problems, and then combine them into a final solution. The first prompt instructs the LLM to propose sub-problems, which are then solved independently. The final prompt instructs the LLM to merge the results of the sub-problems into a single output.

%
\textbf{Thought Propagation (TP)}~\cite{yu2023thought} employs a multi--level double tree structure for prompting the LLM. This approach follows the idea that multi-step problems are easier to solve with access to example solutions of similar problems. This is done in three steps. In the first step, given an input problem the LLM is prompted to propose a set of related problems. The main idea here is that solving these related problems can then be used as reference examples for solving the input problem instead of reasoning from scratch. Additionally, solving similar problems can lead to high-level plans for the input problem and allows TP to rectify errors during planning. The LLM is prompted using in-context examples to generate related problems for both situations. In the second step, the LLM is prompted to solve the input problem as well as the related problems. For this prompt, existing prompting techniques, e.g., CoT, ToT, etc., can be used. Even though the solutions to the related problems are not expert-level, they can be refined in the next step. In the last step, the LLM is prompted to come up with new solutions for the initial problem based on the solutions from the related problems. Additionally, the LLM is prompted to derive high-level plans to solve the input problem using the solutions of the related problems. This process can also be extended to recursively generate further related problems, which yields in the general form a double tree structure with depth $k$.

%
%
\textbf{Socratic Questioning}~\cite{qi2023art} is a prompting scheme that models recursive exploration of the thought space using a tree structure. Hereby, the original question is recursively decomposed into sub-tasks until all tasks can be solved with high confidence. These results are then aggregated and propagated back up the tree to answer the original question. This results in an overall double tree reasoning topology. Additionally, the approach also provides a model to generate an image caption related to the text prompt and thus allows multi-modal reasoning.

\subsection{Directed Graphs}

Some schemes embrace a general directed graph model.

\textbf{Graph of Thoughts (GoT)}~\cite{besta2023graph} uses a multi-prompt approach to improve the LLM problem solving performance by decomposing a given task into sub-tasks that form a graph. This decomposition is specified as a \emph{Graph of Operations (GoO)}. The GoO coordinates how the LLM is prompted and how the results, which form a separate graph called \emph{Graph Reasoning State (GRS)}, are further used in the reasoning process. 

%
\textbf{Graph of Thought}~\cite{lei2023boosting} presents a multi-prompt approach where a graph is constructed recursively in a DFS manner by starting at the \emph{question node} that represents the question to be answered by the LLM. From this node, possible reasoning paths are generated by the LLM. For each path, new nodes, i.e., intermediate reasoning steps, are generated by the LLM and are then used to grow the graph.
To limit the size of the graph, the scheme uses a depth limit and requires a set of \emph{condition nodes} to be provided at the start. These nodes represent axioms for the reasoning process that form initial nodes in the reasoning paths. After the graph construction, a path from the condition nodes to the question node is searched, and a checker module validates each reasoning step along this path. This checking is done by a series of calls to the LLM and only passes with a positive score, if all calls agree on the validity of the step. If no valid path is found, the graph is updated to enable valid reasoning paths.
%
%
In the first step of the update, all nodes only depending on condition nodes (via a valid step) are added to the condition set. Then, new nodes and edges are added to the existing graph in the same DFS manner as when creating the graph from scratch.
These graph updates are repeated until a valid path to the question node is found and therefore a solution to the initial question.

%
\textbf{Graph-of-Thought}~\cite{yao2023beyond} describes a two-stage framework to answer multi-modal questions, i.e., textual questions accompanied by images.
In the first stage, the model generates natural language rationales based on the input text, which provide additional context and knowledge to support answering the given question. This rationale generation is learned as part of the overall model pipeline. In the second stage, these rationales are then appended to the initial question and passed again through the model to predict an answer.
The prediction consists of multiple steps. First, subject-verb-object triples are extracted from the input question using the Stanfords OpenIE system \cite{angeli-etal-2015-leveraging}. Using these triples, a Graph-of-Thought is constructed by first converting each triple to a 3--node path and then merging similar nodes using the Stanford CoreNLP system~\cite{manning-etal-2014-stanford}. Next, the input text and the accompanying image are encoded using a Transformer encoder and a vision encoder, respectively. The Graph-of-Thought is encoded using a GAN. The resulting features from text and image are then combined using cross-attention and passed through a gated fusion layer before passing them to the final Transformer decoder that predicts the rationales in the first stage and the answers in the second stage. The scheme uses the pre-trained T5~\cite{raffel2023exploring} model and fine-tunes it for the rationale generation.

%
\textbf{ControlLLM}~\cite{liu2023controlllm} is a framework that allows answering multi-modal queries, e.g. understanding and generating images, videos, and audio. First, the LLM is prompted to decompose the query into predefined sub-tasks, e.g., "image-processing" or "video-generation". Hereby, the LLM also infers appropriate input and output types, e.g., text or image, from a predefined set for each sub-task. In a next step, a so called \emph{Tool Graph} is constructed by using the generated sub-tasks and the input resources, e.g. video or text, as nodes. Connections between the nodes are drawn for all compatible input/output types. Nodes corresponding to sub-tasks or resources are called \emph{tool nodes} or \emph{resource nodes} respectively. In a next step, the graph is searched in a DFS manner until a solution, i.e., path from input to output resource node, is found. The authors evaluated different heuristics for the DFS, i.e., greedy search, beam search, adaptive beam search and exhaustive search. The graph traversal is guided by the LLM and in each step the LLM is prompted to rate the relevance of the connected tool nodes for solving the overall task, this rating is then used as the heuristic. In a last step, the found solution, i.e., path from input to output in the tool graph, is executed to retrieve the requested output. Finally, the LLM is prompted to summarize the result for the user.
The overall approach is similar to GoT~\cite{besta2023graph}, with the main difference being that each node can access and use a tool and the graph traversal uses a LLM-based heuristic.
%

%
%
\textbf{Cumulative Reasoning}~\cite{zhang2023cumulative} describes a paradigm to solve multi-step problems by iteratively constructing a directed acyclic graph (DAG). A proposer module suggest a next deduction step based on any previous steps, a verifier module evaluates the proposed step and finally, a reporter module checks if a valid solution has been reached and otherwise a next step is proposed.
In this approach, one manually writes problem specific prompts that follow this paradigm. One also manually provides the implementation on how the individual modules interact, i.e., how the DAG is formed. Conceptually, this approach is an extension of ToT as it allows each step to use all previously derived results, meaning different sub trees can be connected to form a DAG.

%
%
\textbf{Everything of Thoughts (XoT)}~\cite{ding2023thoughts} is a two-stage framework utilizing a reinforcement learning model trained using Monte Carlo Tree Search on specific problem sets to generate graphs of thoughts as well as iteratively prompting the LLM to revise and infer solutions based on these graphs. First, the model has to be trained in advance and can be applied to solve problems with clear intermediate steps and solution states, e.g., Game of 24 or 2x2 pocket cube. This model is then used to infer a graph of thought where paths leading to a solved state of the problem are converted to text and fed to an LLM as assisting knowledge to solve the initial question. The LLM is instructed to review the steps and refine them if necessary, hereby iteratively refining the solution graph by applying the first stage to replace incorrect nodes.
%

\textbf{ResPrompt}~\cite{jiang2023resprompt} is a single-prompt approach extending CoT by augmenting the few-shot examples with residual links, i.e., previously derived results. If a current reasoning step depends on previous results, these results are referenced verbatim in parentheses, effectively forming an arbitrary reasoning graph. Later stages can explicitly rely on multiple previous results as opposed to a single previous result.


\subsection{Hypergraphs}

Finally, we also consider a \textbf{hypergraph}, which generalizes a graph by enabling edges to connect arbitrary subsets of nodes instead of being links between just two nodes. We include hypergraphs in the taxonomy, because preliminary works already harness them for multi-modal prompting~\cite{yao2023thinking}.
%

%
%
\textbf{Hypergraph-of-Thought (HoT)}~\cite{yao2023thinking} is a multi-modal reasoning paradigm modeling the thought process as a hypergraph. First, a graph-of-thoughts as in~\cite{yao2023beyond} is constructed. Then a textual hypergraph is constructed, sharing the same nodes. The hyperedges are then defined as node triples, e.g., "(Lionel Messi, place of birth, Rosario)".
%
%
Additionally, a visual hypergraph-of-thought is constructed by performing $k$-means clustering on image patches, where a cluster corresponds to a hyperedge. Both hypergraphs are then encoded and combined to perform graph learning.

\subsection{Analysis \& Comparison of Designs}

We now broadly discuss and analyze graph designs with
respect to different aspects of our blueprint.
A detailed analysis can be found in Appendix \ref{sec: graph_extensive}.

\subsubsection{Topology \& Its Construction}

Firstly, the considered schemes exhibit a blend of single- and multi-prompt aspects, allowing for a high degree of flexibility and control over the prompting process. This is evident in the diverse approaches taken by different schemes such as GoT, ControlLLM, and Cumulative Reasoning, each offering unique ways of constructing and utilizing graphs for problem-solving. Secondly, the user's control over the topology of the graph is significant in most schemes, enabling customization of the reasoning process based on specific needs, such as setting branching factors or defining the depth of the graph. Thirdly, the role of the LLM in these graph-based schemes is multifaceted, involving the generation, evaluation, and modification of nodes within the graph, as well as determining the conclusion of the reasoning process. Lastly, there is a notable variation in the degree of user and LLM influence on the topology across different schemes, with some allowing direct user control, while others rely on predefined heuristics or the LLM's decision-making capabilities.

\subsubsection{Performance}

The considered works universally show improvements in effectiveness of graph-based prompting schemes over chains and trees across various tasks, suggesting a promising direction for future research and application in the field of AI and machine learning.

\section{Benchmarks}

In this section, we introduce benchmarks designed to test the reasoning capabilities of LLMs. These benchmarks will be instrumental in the performance analysis discussion in the following section.

\subsection{Arithmetic Reasoning}
Arithmetic reasoning tasks, also called math word problems, focus on narrative-based mathematical questions where the model must extract and solve numerical equations from a given text-based scenario. In the datasets, each item is presented as question-answering (QA) pairs, in which the answers are formatted as direct answers or multiple choice. Frequently used datasets include GSM8K (Grade School Math)~\cite{cobbe2021gsm8k}, SVAMP (Simple Variations on Arithmetic Math word Problems)~\cite{patel2021nlp}, MAWPS (MAth Word ProblemS)~\cite{koncel2016mawps}, MultiArith~\cite{roy2015solving}, AddSub~\cite{hosseinii2014learning}, DROP (Discrete
Reasoning Over the content of Paragraphs)~\cite{dua2019drop} and AQUA~\cite{ling2017program}.  

Based on these benchmarks, datasets enhanced for more comprehensive textual understanding and more complex calculations were developed. GSM-hard~\cite{gao2022pal} extends GSM8K by introducing calculations of much larger numbers. AQUA-RAT~\cite{ling2017program} requires more detailed rationales in answering than AQUA. MathQA~\cite{amini2019mathqa} selects the QAs from AQUA and provides operation programs. MATH~\cite{hendrycksmath2021} is built on the problem sets from math competitions and
includes \LaTeX~representations. ALGEBRA~\cite{heyueya2023solving} focuses on the algebraic domain.

Researchers take specific domains into consideration. FinQA (Financial Question Answering)~\cite{chen2021finqa} involves the analysis and interpretation of financial data, reports, and tables. ConvFinQA~\cite{chen2022convfinqa} presents the financial questions in a conversational manner. To test the comprehension of structured data, TabMWP~\cite{lu2023dynamic} collects math word problems that involve tabular data.

\subsection{Commonsense and Logical Reasoning}
In commonsense reasoning, benchmarks test the abilities to understand text and to generate responses that align with human-like commonsense knowledge. Following that principal, group of datasets composed of multi-hop questions are constructed, including HotPotQA~\cite{yang2018hotpotqa}, StrategyQA~\cite{geva2021strategyqa}, MuSiQue~\cite{trivedi2022musique}, Bamboogle~\cite{press2210measuring}, and CommaQA-E (Communicating with Agents for QA)~\cite{khot2022hey}.
Apart from the complexity of the questions, more QA datasets incorporating stepwise explanations for the answers are built, such as 2WikiMultiHopQA (2Wiki stands for Wikipedia and Wikidata)~\cite{ho2020constructing}, ProofWriter~\cite{tafjord2021proofwriter}, EntailmentBank~\cite{Dalvi2021ExplainingAW}. SocialQA~\cite{srivastava2023beyond} focuses on daily social events.

Logical reasoning tasks aim to test the ability to understand contextually cause-and-effect relationships. CauseEffect~\cite{srivastava2023beyond} tasks the model with determining the causal event given descriptions of two events.

\subsection{Symbolic Reasoning and Other Domains}
There are several tasks involving manipulations of symbols in different domains.
For Last Letter Concatenation~\cite{wei2022chain}, the model concatenates the last the letters of two given words into an abbreviation.
Coin Clipping~\cite{wei2022chain} tasks the model with determining, whether the head side of a coin faces up after a group of "flip/not flip" commands.

In the domain of spatial planning tasks, Brick World~\cite{hu2023chain} asks the model to acquire a specific brick after sequentially grabbing group of bricks in 2D/3D scenarios. SCAN (Simplified version of the CommAI Navigation tasks)~\cite{lake2018generalization} consists of sets of compositional navigation commands paired with the corresponding action sequences.
Other text-based spatial tasks are also popular, such as NLVR-based Manipulation~\cite{suhr2017corpus} and SPARTUN~\cite{mirzaee2022transfer}

Semantic parsing datasets focus on the ability of the models to
generalize from knowledge about components. Examples include CFQ (Compositional
Freebase Questions)~\cite{Keysers2020}, COGS (COmpositional Generalization
Challenge based on Semantic Interpretation)~\cite{kim2020cogs}, and
Alchemy~\cite{long2016simpler}.

Other benchmarks are intended for special domains, such as coding
(CodeNet~\cite{puri2021codenet}) or pedestrian traffic behavior (PIE (Pedestrian
Intention Estimation)~\cite{rasouli2019pie}) for autonomous driving.

Benchmarks that target multiple domains like MT-Bench~\cite{zheng2023judging}
and MMLU (Massive Multitask Language Understanding)~\cite{hendryckstest2021}
measure how broad the knowledge of a model is. Sometimes just multiple areas of
a single domain are targeted, for example with ScienceQA~\cite{lu2022learn}.
ASQA (Answer Summaries for Questions which are
Ambiguous)~\cite{stelmakh2023asqa} focuses on ambiquous questions in the context
of long-form question answering. Several benchmarks such as
ALFWorld~\cite{ALFWorld20} and VQA-V2 (Visual Question
Answering)~\cite{goyal2017making} look into visual tasks, while others target
logical reasoning such as LogiQA~\cite{liu2020logiqa} and
FOLIO~\cite{han2022folio}. AutoTNLI ~\cite{kumar-etal-2022-autotnli} consists of
counterfactual entity-based tables.

Other evaluation methods also include puzzles, like the Game of 24 (given four
numbers, find the combination of those numbers and the four basic mathematical
operations, so that the end result is 24), the $8$-Puzzle (3x3 board with a
missing piece, goal is to slide the pieces to their target location) or 2x2
pocket cube (smaller version of the Rubik's cube).

\section{Detailed Analyses}

\subsection{Chain}
\label{sec: chain_extensive}
\subsubsection{Performance Analysis}

Chain-structured prompting methods have been applied to solve questions in diverse fields. We conduct a qualitative comparison of these methods in different domains.

\paragraph{\textbf{Arithmetic Reasoning}}

IO prompting exhibits very low performance on mathematical tasks. 
Compared with IO prompting, CoT attains an accuracy of around 60\% (40\% higher) on GSM8K~\cite{cobbe2021gsm8k}, 80\% (10\% higher) on SVAMP~\cite{patel2021nlp} and 90\% (14\% higher) on MAWPS~\cite{koncel2016mawps}. The accuracy of the rationales significantly increases with the scale of LLMs. 
Zero-shot-CoT does not outperform CoT on arithmetic tasks, but obtains a 60\% gain in accuracy on MultiArith~\cite{roy2015solving} over IO prompting, simultaneously exceeding IO prompting on GSM8K, AQUA~\cite{garcia2020dataset} and SVAMP tasks.
Compared with CoT, PoT with few-shot examples improves on GSM8K and AQUA by more than 8\% and 4\% on SVAMP. For the financial datasets FinQA~\cite{chen2021finqa} and ConvFinQA~\cite{chen2022convfinqa}, few-shot PoT reaches a roughly 20\% improvement over CoT.
In comparison with CoT, schemes that employ decomposition demonstrate a modest advantage on math reasoning tasks. Least-to-Most Prompting increases the accuracy on DROP~\cite{dua2019drop} by 10\% because of the strong decomposability of the problems in the dataset. Although Least-to-Most Prompting only shows a slight average improvement on GSM8K, it significantly enhances accuracy (by 5\%) for problems within GSM8K that require more than 5 inference steps to solve. PS+ prompting yields at least 5\% improvement on MultiArith, AQUA, AddSub~\cite{hosseinii2014learning}, and SVAMP, but only a slight gain on GSM8K due to the high complexity and low decomposability of the problems in GSM8K.
Schemes incorporating reflection nodes show limited advancement in math-related tasks, yet exhibit promising capabilities in contextual and textual comprehension.
Tested on all math-related benchmarks mentioned above, PVS exhibits a 5.5\% improvement on average. Moreover, PVS shows an improvement of accuracy larger than 10\% on ALGEBRA~\cite{heyueya2023solving} and GSM-hard~\cite{gao2022pal} that contain algebraic calculations with large numbers, due to the integration of PoT and EoT methods.
With the plug-in of the table reader tools, Chameleon increases the accuracy of answering by 8\% over CoT in the math questions represented in tabular context on the TabMWP~\cite{lu2023dynamic} dataset.

\paragraph{\textbf{Commonsense Reasoning}}

CoT outperforms IO prompting in terms of accuracy (75.6\% vs 69.4\%) on the StrategyQA~\cite{geva2021strategyqa} dataset, and performs 20\% better on sports understanding tasks than experienced humans. Zero-shot-CoT does not provide performance gains on commonsense reasoning benchmarks over IO prompting.
The evaluation of SelfAsk focuses on multi-hop questions that require composing multiple facts to solve and improves over CoT by smaller margins on 2WikiMultiHopQA~\cite{ho2020constructing} and MuSiQue~\cite{trivedi2022musique}, but by 11\% on Bamboogle~\cite{press2210measuring} without heavy token usage.
Tested on bAbI QA~\cite{dodge2016evaluating} and ProofWriter OWA~\cite{tafjord2021proofwriter} datasets, the SI framework enabled, on average, a 7B LLM (58.75\% accuracy) to outperform a 280B LLM (44.03\%) within the scope of CoT frameworks.
Taking CoT as the baseline, schemes with decomposition show varying levels of performance on commonsense reasoning tasks.
The problems in SCAN~\cite{lake2018generalization} require the conversion of single textual commands into action sequences, which suits the schemes that incorporate decomposition nodes. Here, Least-to-Most achieves 99.7\% accuracy compared to just 16.2\% for CoT.
While PS+ shows limited improvement on StrategyQA due to the simple guidance from zero-shot prompting, Decomposed Prompting attains a 20\% improvement on the CommaQA-E~\cite{khot2022hey} dataset and a 30\% improvement on 2WikiMultiHopQA and Musique, which stem from the designed retrieval function.
With the retrieval and calculation tools, ChatCoT obtains a 20\% gain on the HotPotQA~\cite{yang2018hotpotqa} dataset compared with CoT.

Refinement improves incorrect thoughts, leading to higher reasoning abilities of LLMs. Reflexion's success rate on HotPotQA progressively improves with the accumulation of refinement iterations and surpasses 70\% after six attempts, while CoT maintains a steady success rate of around 30\%. LogiCoT also exhibits enhancement in reasoning tasks related with contextual understanding and causality inference, achieving an improvement of around 10\% on the SocialQA~\cite{srivastava2023beyond} and CauseEffect~\cite{srivastava2023beyond} datasets.

\paragraph{\textbf{Symbolic Reasoning}}

In the task of concatenating the last letters of words, CoT achieves almost 100\% accuracy on cases that were part of the in-context exemplars and around 70\% accuracy on out-of-domain (OOD) cases, while IO prompting shows nearly no accuracy on both cases. For coin flipping, IO prompting with a 540B LLM achieves nearly 100\% on in-domain cases, but is inferior to CoT (50\% vs 90\%) on OOD cases. The problem-solving rate of Zero-shot-CoT is lower than CoT, but higher than IO on coin flip and last letter concatenation tasks.
In the domain of spatial reasoning, CoS exhibites an increase of accuracy ranging from 2\% to 40\% on Brick World, NLVR-based Manipulation~\cite{suhr2017corpus}, Natural Language Navigation~\cite{gu2022vision} and the spatial QA dataset SPARTUN~\cite{mirzaee2022transfer}, which completely exceeds the performance of CoT while using a lower number of tokens.
The use of decomposition enables multi-prompt chain schemes to outperform CoT on OOD cases of the last letter concatenation task, where the length of the tested words exceeds those in the exemplars. 
The accuracy of the LLM gradually decreases from 94\% to 74\% as word length increases from 4 to 12. In contrast, CoT's accuracy drops from 70\% to 32\%. The accuracy of the zero-shot method PS+ is still 10\% higher than few-shot CoT's. Moreover, in the more complex task of concatenating the \( k^{\text{th}} \) letter, Decomposed Prompting shows higher accuracy (more than 96\%) than CoT (around 70\%) and Least-to-Most (around 12\%).

\paragraph{\textbf{Reasoning in Special Domains}}

The performance evaluation of the schemes in special domains is only selectively and qualitatively detailed here, due to the non-uniform, artificially designed metrics utilized in the assessment. 
Refinement schemes perform well in the code-related tasks. SELF-REFINE successfully increases readability of code on the CodeNet~\cite{puri2021codenet} dataset, and obtains higher speedup than the IO prompting methods on code optimization with the PIE~\cite{rasouli2019pie} dataset. Meanwhile Reflexion achieves high accuracy in code generation on MBPP~\cite{austin2021program} and HumanEval~\cite{chen2021evaluating}.

\subsubsection{Representations of Topology \& Schedule}
Listing~\ref{lst:CoT-prompt} shows an example chain topology from \cite{hu2023chain}.

\lstalias{plantuml}{}
\lstset{escapeinside={<@}{@>}}
\begin{lstlisting}[
    float,
    language=plantuml,
    numbers=none, 
    % basicstyle=\ttfamily\small, 
    showspaces=false, showtabs=false,
    breaklines=true,
    showstringspaces=false,
    breakatwhitespace=true,
    breakautoindent=false,
    breakindent=8pt,
    columns=flexible,
    label={lst:CoT-prompt}, 
    caption={\textbf{An example \ul{implicit} \ul{single-prompt} chain topology, encoded with \ul{text}}. It shows the CoT example of a Brick World problem from~\cite{hu2023chain} with a linear sequence of four connected nodes followed by the final solution node.}
]
<@\textbf{User:}@>
Question: There are a set of bricks. Brick E is on top of brick A. Now we have to get a specific brick. The bricks must be grabbed from top to bottom, and if the lower brick is to be grabbed, the upper brick must be removed first. How to get brick A?
Answer: Let us think step by step:

<@\textbf{LLM:}@>
1. To get brick A, we find E is on top of A.
2. We find E is on the top.
3. We need to remove brick E first, as it is on top of brick A.
4. Brick A is now accessible and can be grabbed.
So we get the result as E, A.
\end{lstlisting}

\subsection{Tree}
\label{sec: tree_extensive}
We now broadly discuss and analyze tree designs with respect to different aspects of our blueprint.



\subsubsection{Topology \& Its Construction}

\paragraph{\textbf{Multi-Prompt Schemes}}
The vast majority of schemes are {multi-prompt}~\cite{creswell2022faithful, drozdov2022compositional, hu2023treeofmixedthought, kim2023tree, long2023large, mo2023tree, ning2023skeleton, morris2023tree, wang2022self, xie2023decomposition, yao2023tree}.
%
%
%
%
%
In all these schemes, the topology is {to a certain degree dynamically constructed by the LLM (within the set boundaries of the user and the defined approach)}. 
\emph{The role of the LLMs} can be summarised as (1) generating new child nodes, (2) evaluating given nodes and (3) deciding when we reached a final solution and reporting the results. 
Then, \emph{the user to a large degree also controls the construction process}.
In ToT by Long~\cite{long2023large}, the user provides a step limit as well as a checker module (rule-based or as a DNN) that decides if a reasoning step is valid or if backtracking to a previous node is necessary.
ToT by Yao et al.~\cite{yao2023tree} allows the user to choose the number of samples to generate at each node from which the most promising $b$ candidates are kept for BFS, here $b$ defines the branching factor of the tree. When using DFS, the user provides a value threshold, so when a node evaluates to a lower score than the threshold, one backtracks to the parent and continues from there. For both presented exploration schemes (BFS and DFS) the user provides an upper bound on the total number of nodes.
Thought Decomposition~\cite{xie2023decomposition} explores the tree using stochastic beam search. Hereby, the tree is constructed level by level and, in each level, $k$ candidates are kept (the beam size) and $n$ new nodes are generated for each candidate. The randomness in the stochastic beam search is controlled by a user-defined parameter, similarly, the user sets the temperature for generating new samples from the LLM. The total number of steps in the search, i.e., the tree depth, is capped at $16$.
In CoT-SC~\cite{wang2022self}, the user simply provides the number of CoT samples that should be generated, resulting in a depth--1 tree.
Creswell and Shanahan~\cite{creswell2022faithful} use beam search to explore the tree, letting the user choose the number of candidate nodes to generate on each level, the beam size as well as the maximum depth of the tree.
In Dynamic Least-to-Most Prompting~\cite{drozdov2022compositional}, the user has no direct influence on the structure of the tree topology. The topology is entirely based on the results of the LLMs decomposition of the input.
TouT~\cite{mo2023tree} has the same user parameters as ToT by Yao et al.~\cite{yao2023tree}, but additionally uses an uncertainty threshold for DFS to backtrack from nodes where the uncertainty gets too large.
In TomT~\cite{hu2023treeofmixedthought}, the user can influence the tree topology by limiting the total number of nodes, setting a branching factor and selecting a block size $k$ which indirectly limits the depth of the tree by generating $k$ consecutive reasoning steps for each node instead of using one node per reasoning step as in the ordinary ToT~\cite{yao2023tree}.
ToC~\cite{kim2023tree} limits the tree size by an upper node limit as well as setting a maximum depth. The user can choose a branching factor $k$ which is used to dynamically select $k$-shot in-context examples for the generation prompt such that the appropriate number of child nodes are generated.
Finally, in SoT~\cite{ning2023skeleton} the depth--1 tree topology is only influenced by the response of the LLM (i.e., how the LLM determines the number of child nodes). The user can only indirectly influence this topology by possibly altering the node generation prompt itself.
%


\paragraph{\textbf{Single-Prompt Schemes}}

Only AoT~\cite{sel2023algorithm} exclusively uses a single-prompt tree topology. Here, in-context examples are designed such that the reasoning follows a tree structure, i.e., the examples show algorithmic reasoning steps such as trying out different solutions with DFS on the solution space, to make the LLM search for the solution in a similar tree-structured fashion. Thus, while in-context examples are provided by the user as guidance, the topology for solving the input question is constructed dynamically on-the-fly solely by the LLM.

\paragraph{\textbf{Beyond Single- and Multi-Prompt}}

In all the above schemes, the root node holds the initial question and remaining nodes represent intermediate or final solutions. 
Contrarily, Tree Prompting~\cite{morris2023tree} \emph{builds its topology during an initial learning phase}, by learning the structure of a binary decision tree through training samples, with the goal of text classification. During inference, the tree structure is fixed and independent of the specific input.
%
%
The user can indirectly influence the tree topology by manually providing the prompt-candidates used for learning the binary decision tree at the training stage.

%
%
%

\subsubsection{Performance}

We perform a qualitative performance evaluation of different tree schemes. We consider what problems or datasets the approach is evaluated on, how the quality of answers scales with respect to the number of prompts (or tokens), and what the trade-offs in quality are with respect to different tree topologies or reasoning schedules. 


\paragraph{\textbf{Considered Problems \& Datasets}}

ToT by Long~\cite{long2023large}, ToT by Yao et al.~\cite{yao2023tree}, AoT~\cite{sel2023algorithm}, and TouT~\cite{mo2023tree} evaluate their performance on problems that can be simply decomposed in a tree-structured way, such as 5x5 Sudoku puzzles in ToT by Long, and Game of 24 or 5x5 Crossword puzzle for the other three approaches.
Thought Decomposition~\cite{xie2023decomposition} and CoT-SC~\cite{wang2022self} perform their evaluation on arithmetic reasoning, e.g, GSM8K~\cite{cobbe2021gsm8k}, symbolic reasoning, e.g., object counting or last letter concatenation, as well as on commonsense reasoning tasks such as CommonsenseQA~\cite{talmor-etal-2019-commonsenseqa} and StrategyQA~\cite{geva2021strategyqa}.
The scheme of Creswell and Shanahan~\cite{creswell2022faithful} assumes access to an existing context to reason on and evaluate their approach on ProofWriter~\cite{tafjord2021proofwriter} and EntailmentBankQA~\cite{Dalvi2021ExplainingAW}.
Dynamic Least-to-Most Prompting~\cite{drozdov2022compositional} is evaluated on semantic parsing datasets (CFQ~\cite{Keysers2020} and COGS~\cite{kim2020cogs}), TomT~\cite{hu2023treeofmixedthought} create a new synthetic visual question answering dataset for evaluation by combining two existing datasets, ToC~\cite{kim2023tree} is evaluated on a long-form QA dataset (ASQA~\cite{stelmakh2023asqa}), Tree Prompting~\cite{morris2023tree} uses $13$ text classification datasets and SoT~\cite{ning2023skeleton} is evaluated on diverse questions from \cite{vicuna2023, xu2023wizardlm}.

\paragraph{\textbf{Accuracy vs.~\#prompts, topology variant, \& reasoning schedule}}

Long~\cite{long2023large} does not discuss performance evaluation and the scheme is executed until a solution is found or a maximum of $100$ prompts are issued. Yao et al.~\cite{yao2023tree} evaluate their approach on Game of 24 using BFS on a tree of depth $3$ with a branch factor of $5$ and scoring each node $3$ times, on Creative Writing using a tree of depth $2$, branch factor of $5$ and selecting the best node in each level by scoring $5$ times, and finally on solving $5 \times 5$ Crosswords with a ToT of depth $5$ to $10$ using DFS, a branch factor of $5$ and at most $100$ DFS steps.
Thought Decomposition~\cite{xie2023decomposition} performs beam search on a ToT with branching factor $16$ and beam size $5$, essentially generating $5\cdot 16=80$ samples at each tree level, while the depth is limited to $16$. The authors find out that increasing the branching factor leads to more diversity and thus improves performance when using majority voting on the last level. This increase comes at the cost of using more tokens.
In CoT-SC~\cite{wang2022self}, the authors evaluate the number of reasoning chains to sample, i.e., using a tree of depth one with a branching factor of $1$ up to $40$ while reporting the average over $10$ runs. Sampling more outputs improves accuracy, but also increases computational cost; most gains are realized after sampling $5$ to $10$ CoT-paths in practice.
The approach of Creswell an Shanahan~\cite{creswell2022faithful} assumes access to a context, i.e., a set of statements, to reason on. The individual LMs (selection, inference, two different halter modules, and correctness prediction of a step) need to be fine-tuned separately in advance, which makes this approach very specific and incurs a setup cost.
Dynamic Least-to-Most Prompting~\cite{drozdov2022compositional} comes with no explicit discussion about the number of tokens used, but the number of prompts varies with the decomposition of the specific input phrase and so does the number of in-context exemplars. The authors claim a speedup of over a factor $2$ compared to CoT-SC prompting as only a single result is generated and no majority voting is necessary as in CoT-SC.
AoT~\cite{sel2023algorithm} uses a single prompt for the Game of 24 (reporting $9\%$ out of token errors) and two prompts for the 5x5 Crossword puzzle. The authors claim that ToT uses many more prompts, i.e., over $100$ rsp.~$200$, for the two problem setups. Hereby, AoT is reported to perform better than ToT for the Game of 24. The authors show that the impact of more exploration steps inside the examples leads to longer prompt generations, i.e., slower inference for the same number of games.
TouT~\cite{mo2023tree} sample $20$ LLM responses per ToT node to compute local variances and claim improved performance to ToT for both Game of 24 and 5x5 Crossword puzzles and further increasing or decreasing the number of samples leads to degrading performance.
In TomT~\cite{hu2023treeofmixedthought}, the DFS is limited to $30$ steps with their ToT-One-Stop and the reference ToT implementation using a branching factor of $3$. The authors report improved accuracy for ToT-One-Stop compared to ToT while using approximately half of the number of steps.
ToC~\cite{kim2023tree} claims comparable performance to Cot-SC with using less than $20$ LLM calls.
According to the published results, Tree Prompting~\cite{morris2023tree} enables smaller models to outperform larger models and the accuracy can be improved at the cost of more LLM calls; the number of LLM calls for the experiments is limited to $40$. 
Finally, SoT~\cite{ning2023skeleton} reports end-to-end latency speedups across different task categories and LLMs. The authors compare the two step decoding process of SoT, i.e., creating short bullet point style answers which are expanded in the second step, against ``{normal} decoding'', where a continuous answer is generated in one go. There are a varying number of bullet points generated with the average being $9$ bullet points. The authors assess the answer quality with the help of two LLM judges (FastChat and LLMZoo) using different metrics. SoT reports a high token overhead of roughly $60$x to $90$x, which can be optimized down to roughly $30$x.

\subsection{Graph}
\label{sec: graph_extensive}

We now broadly discuss and analyze graph designs with
respect to different aspects of our blueprint.

\subsubsection{Topology \& Its Construction}



Most graph schemes combine to a certain degree single- and multi-prompt aspects.
GoT by Besta et al.~\cite{besta2023graph} requires a problem specific graph of operations for solving a problem in a multi-prompting fashion. One has complete freedom over the individual prompts to the LLM and can incorporate arbitrary single-prompting schemes. GoT by Lei et al.~\cite{lei2023boosting} dynamically constructs the graph, at the time of this writing no code or prompt templates have been published, so it is not clear to what extend single-prompting is used. GoT by Yao et al.~\cite{yao2023beyond} uses a NLP approach to construct a graph of thought and uses an LM as part of the AI-pipeline to learn how to use these graphs of thought for predicting an answer. ControlLLM~\cite{liu2023controlllm} is similar to GoT~\cite{besta2023graph} in the sense that all nodes in the \emph{Tool Graph} correspond to nodes in the GoT with access to tools. Cumulative Reasoning~\cite{zhang2023cumulative} is a conceptual extension of ToT with access to all previously derived results, i.e., nodes, resulting in a DAG. XoT~\cite{ding2023thoughts} uses a problem-specific pre-trained RL-model to create a graph of thought and uses the LLM to review and summarize the results.

In graph schemes, \emph{the user has significant control over the topology and its creation}.
In BSM~\cite{saha2023branch}, the topology is a double tree with one intermediate level. The user can set either the number of branches, or its upper bound by specifying the number of sub-problems to generate inside a single prompt. 
In Thought Propagation~\cite{yu2023thought}, the user specifies the branching factor in a single prompt to generate analogous problems, and can predefine the number of levels of the double tree to grow the topology.
The user in Socratic Questioning~\cite{qi2023art} defines the branching factor of the topology by specifying its value, lower limit, or upper limit within a single prompt. Furthermore, the user can limit the growth of the topology by defining the maximum tree depth.
%
In Graph of Thoughts (GoT)~\cite{besta2023graph}, the user defines the complete topology of reasoning across multiple prompts for a given use case, including, but not limited to, the branching factor, the maximum depth of tree, and the number of child nodes to keep. 
The topology in Graph of Thought (GoT)~\cite{lei2023boosting} is influenced by a user-defined depth limit.
In Cumulative Reasoning~\cite{zhang2023cumulative}, the user can influence the total number of nodes as the proposer (LLM) iteratively generates a new node until a user-specified limit is reached.
%
In ControlLLM~\cite{liu2023controlllm}, the user does not define the number of nodes during the task decomposition stage for solving a given task. However, a user may have implicit influence on the number of nodes via prompting, e.g., by adding the phrase ``Parse out as few tasks as possible''.
The user in Everything of Thoughts (XoT)~\cite{ding2023thoughts} can indirectly influence the topology by defining the number of times the LLM is prompted to review and refine the solution graph. 
In ResPrompt~\cite{jiang2023resprompt}, the user simply provides in-context examples which directly dictate the basis of the topology. 
Differing from previous schemes, users in two-stage frameworks such as Graph-of-Thought (GoT)~\cite{yao2023beyond} and Hypergraph-of-Thought (HoT)~\cite{yao2023thinking} do not influence the topology as the (hyper)graph-of-thought is constructed by predefined heuristics. 


Similar to prompting with trees, \emph{the LLM may also influence the topology of prompting with graphs} within user-defined boundaries. The role of the LLM can be summarized as (1) generating new child nodes, (2) evaluating given nodes, (3) deleting or backtracking from a new node (e.g., based on evaluation results), and (4) determining when the reasoning process should be concluded.

\subsubsection{Performance}

\paragraph{\textbf{Considered Problems \& Datasets}}

BSM~\cite{saha2023branch} evaluates its performance on answering questions from multiple areas by using MT-Bench~\cite{zheng2023judging}. 
TP~\cite{yu2023thought} performs shortest-path reasoning, creative writing and LLM-Agent Planning on ALFWorld~\cite{ALFWorld20}.
Socratic Questioning~\cite{qi2023art} evaluates their method on both, text-only datasets (e.g. MMLU~\cite{hendryckstest2021} and MATH~\cite{hendrycksmath2021}), logical reasoning questions (LogiQA~\cite{liu2020logiqa}), as well as multi-modal visual QA datasets such as VQA-V2~\cite{goyal2017making} and two others.
GoT by Besta et al.~\cite{besta2023graph} performs integer sorting with duplicates, set intersections, keyword counting and document merging.
GoT by Lei et al.~\cite{lei2023boosting} measures its performance on the Game of $24$, solving higher-degree polynomial equations as well as deriving formulas for recursive sequences.
GoT by Yao et al.~\cite{yao2023beyond} evaluates on the arithmetic reasoning benchmark GSM8K~\cite{cobbe2021gsm8k} and on ScienceQA~\cite{lu2022learn}, which contains science questions with images.
ControlLLM~\cite{liu2023controlllm} provides its own benchmark containing tasks for image editing and perception, visual QA and the usage of over $20$ different common tools.
Cumulative Reasoning~\cite{zhang2023cumulative} evaluates its approach on FOLIO~\cite{han2022folio}, AutoTNLI~\cite{kumar-etal-2022-autotnli}, Game of $24$, as well as mathematical reasoning on the MATH~\cite{hendrycksmath2021} dataset.
XoT~\cite{ding2023thoughts} evaluates on Game of $24$, $8$-Puzzle and on a $2$x$2$ pocket cube.
ResPrompt~\cite{jiang2023resprompt} uses multiple math reasoning benchmarks for evaluation, i.e., GSM8K~\cite{cobbe2021gsm8k}, AQUA-RAT~\cite{ling2017program}, MathQA~\cite{amini2019mathqa} and SVAMP~\cite{patel2021nlp}, as well as the sequential reasoning benchmark SCONE-Alchemy~\cite{long2016simpler} and commonsense reasoning questions from StrategyQA~\cite{geva2021strategyqa}.
HoT~\cite{yao2023thinking} evaluates their approach on ScienceQA~\cite{lu2022learn}.

In summary, there is no clear set of benchmarks that is used consistently with all approaches, but most include math reasoning problems or some common reasoning tasks such as Game of $24$.

\paragraph{\textbf{Accuracy vs.~\#prompts, topology variant, \& reasoning schedule}}

BSM~\cite{saha2023branch} uses a branching factor of $5$ for its experiments and outperforms zero-shot prompting and CoT-SC (sampling $5$ times). BSM makes LLaMA-2-70B-chat competitive with zero-shot GPT-4 for the turn-1 questions. BSM has similar computational requirements as CoT-SC, but yields higher scores. Additionally, BSM benefits smaller models (LLaMA-2-7B), where CoT-SC is ineffective. The benefits of increasing the branching factor seems to be saturated at $4$, but depends on the specific problem.
TP~\cite{yu2023thought} for shortest-path reasoning makes significant performance gains over all the baselines (IO, CoT, ToT) for all evaluated models, i.e., PaLM-2, GPT-3.5 and GPT-4. The authors report that there is only a marginal difference between $1$-shot and $5$-shot prompting without further exploration. The performance can directly be influenced by the number of layers used for TP. Hereby, $1$-layer TP has competitive performance to ToT with a similar amount of tokens used.
%
Socratic Questioning~\cite{qi2023art} is compared against IO, CoT, CoT-SC and ToT prompting. The authors report ~3-4\% absolute gain over other methods. The reported ToT accuracy is relatively low compared to the other baselines. 
GoT by Besta et al.~\cite{besta2023graph} significantly improves upon IO, CoT, and ToT on all four tasks using GPT-3.5. The accuracy enhancement of GoT over the best-performing baseline (ToT) is achieved with lower or comparable number of tokens. The performance advantages of GoT over the baselines increase with the problem size.

\fi

\end{document}